\documentclass{article}
\PassOptionsToPackage{numbers, compress}{natbib}

\usepackage[main, final]{neurips_2026}

\usepackage[utf8]{inputenc} 
\usepackage[T1]{fontenc}    

\usepackage[hypertexnames=false, colorlinks=true, urlcolor=blue, linkcolor=red, hyperfootnotes=false]{hyperref}
\usepackage{bookmark}

\usepackage{url}            
\usepackage{booktabs}       
\usepackage{amsfonts}       
\usepackage{nicefrac}       
\usepackage{microtype}      
\usepackage{xcolor}         

\usepackage{xcolor}         
\usepackage[table]{xcolor} 

\usepackage{subcaption}
\usepackage{microtype}
\usepackage{url}
\usepackage{booktabs}
\usepackage{lineno}
\usepackage{pifont}

\definecolor{darkblue}{rgb}{0, 0, 0.5}
\hypersetup{colorlinks=true, citecolor=darkblue, linkcolor=darkblue, urlcolor=darkblue}

\usepackage{caption}
\usepackage{graphicx}
\usepackage[utf8]{inputenc}
\usepackage[T1]{fontenc}
\usepackage[english]{babel}
\usepackage{url}
\usepackage{times}
\usepackage{csquotes}
\usepackage{xcolor}
\usepackage{breakurl}
\usepackage{ragged2e}
\usepackage[export]{adjustbox}
\newrobustcmd{\B}{\boldmath}
\usepackage{multirow}
\usepackage{dsfont}
\usepackage{amsmath,amssymb,amsthm}
\usepackage{amsfonts}
\usepackage{cleveref}
\usepackage{scalefnt}
\usepackage{enumitem}
\usepackage{upgreek}

\usepackage{subcaption}
\usepackage{booktabs}  
\usepackage{amsfonts}  
\usepackage{nicefrac}  
\usepackage{microtype} 
\usepackage{xcolor}    
\usepackage[linesnumbered,ruled,vlined,noend]{algorithm2e}
\usepackage{color}
\usepackage{float}
\usepackage{bm}
\usepackage{dsfont}
\usepackage{xspace,lipsum}
\usepackage{thm-restate}
\usepackage[disable]{todonotes}
\usepackage[outdir=./]{epstopdf}
\usepackage{tikz}
\usepackage{pgfplots}
\pgfplotsset{compat=1.11}
\usepgfplotslibrary{fillbetween}
\usetikzlibrary{intersections}
\usetikzlibrary{shapes.geometric}
\usepackage{cleveref}
\usepackage{tabularx}
\usepackage{wrapfig}

\usepackage{thmtools}
\theoremstyle{plain}
\newtheorem{theorem}{Theorem}[section]

\newtheorem{lemma}[theorem]{Lemma}

\theoremstyle{definition}

\newtheorem{assumption}[theorem]{Assumption}
\theoremstyle{remark}

\newcommand{\algname}[1]{{\sf\color{darkgray}\scalefont{0.90}{#1}}\xspace}

\newcommand{\W}{\mathbf{W}}
\newcommand{\F}{\mathrm{F}}
\newcommand{\normF}[1]{\left\|#1\right\|_\F}
\newcommand{\ip}[2]{\langle #1,\,#2\rangle_\F}

\title{Stochastic Projections and Rank-Aware Learning Rates for Federated LoRA}

\author{%
  Saber Malekmohammadi \\
  Stanford University\\
  \texttt{sabermm@stanford.edu} \\
  \And
  Kevin Kuo \\
  Carnegie Mellon University \\
  \texttt{kkuo2@andrew.cmu.edu}
  \AND
  Virginia Smith \\
  Carnegie Mellon University \\
  \texttt{smithv@cmu.edu}
  \And
  Golnoosh Farnadi \\
  McGill University, MILA \\
  \texttt{farnadig@mila.quebec}
}

\begin{document}

\maketitle

\begin{abstract}
Fine-tuning large-scale foundation models across resource-constrained, distributed clients presents significant computational and communication challenges. While merging Low-Rank Adaptation (\algname{LoRA}) with Federated Learning (\texttt{FL}) enables rank-based flexibility, the interdependence of the adaptation matrices ($\mathbf{A}$ and $\mathbf{B}$) often complicates optimization. In this work, we theoretically demonstrate that \algname{LoRA} fine-tuning with Gaussian initialization and a frozen $\mathbf{A}$ matrix acts as a stochastic approximation of full fine-tuning, where batch gradients are subject to a random, rank-dependent perturbation. Motivated by this insight, we propose freezing and periodically resampling $\mathbf{A}$ matrices at the client level. This strategy decouples the $\mathbf{A}$ and $\mathbf{B}$ gradients, diminishes \algname{LoRA} low-rank constraints and reduces both computational and uplink communication overheads by 50\%, allowing for the reallocation of resources toward higher adaptation ranks. Furthermore, we show that in this frozen-$\mathbf{A}$ regime, the adaptation rank $r$ directly governs the signal-to-noise ratio (\texttt{SNR}) of the perturbed gradients. Specifically, we find that the \texttt{SNR} increases linearly with $r$, implying that clients' learning rates must be dynamically scaled with their adaptation ranks for higher utility. Building on these foundations, we introduce \algname{SPRA-LoRA} (\textbf{S}tochastic \textbf{P}rojection and \textbf{R}ank-\textbf{A}ware \algname{LoRA}), a federated rank-aware fine-tuning strategy that utilizes frozen, resampled $\mathbf{A}$ matrices and rank-dependent learning rates for clients. We also prove its convergence with heterogeneous client resources. Our experiments across benchmark datasets demonstrate that \algname{SPRA-LoRA} consistently outperforms state-of-the-art baselines with half the overhead on clients. The implementation codes for the project are available at this link:  \href{https://github.com/Saber-mm/SPRA-LoRA}{https://github.com/Saber-mm/SPRA-LoRA}. 

\end{abstract}

\section{Introduction}
Adapting foundation models (e.g., \texttt{GPT} \citep{openai2024gpt4technicalreport}, \texttt{BERT} \citep{devlin-etal-2019-bert} and \texttt{ViT} \citep{dosovitskiy2021an}) to downstream tasks requires the use of fine-tuning datasets that are often unavailable to or represented by just a single client. Federated learning (\texttt{FL}) \citep{flmcmahan} addresses this by enabling collaborative fine-tuning across multiple clients in a privacy-preserving manner. Despite the promise of \texttt{FL}, the large number of parameters in foundation models can make  federated fine-tuning expensive.

To reduce the computational and communication costs of federated fine-tuning, Parameter Efficient Fine-Tuning (\texttt{PEFT}) methods \citep{prompttuning2021, prefixtuning2021, lora} freeze most pre-trained parameters, updating only a small subset. Specifically, \algname{LoRA} \citep{lora} is a popular \texttt{PEFT} technique that has been integrated with \algname{FedAvg} \citep{flmcmahan} for federated fine-tuning of \texttt{LLM}s \citep{FedIT}.  \algname{LoRA} formulates the update to an \emph{adapter} $\mathbf{W}^0 \in \mathbb{R}^{n \times m}$ of a pre-trained model as $\mathbf{W} = \mathbf{W}^0 + s\mathbf{B}\mathbf{A}$, where $\mathbf{B} \in \mathbb{R}^{n \times r}, \mathbf{A} \in \mathbb{R}^{r \times m}$ are matrices of rank $r$ and $r \ll \min \{m, n\}$. Also, $s = \alpha/r^\nu$ is a scaling factor, where $\alpha$ is a hyperparameter. This reduces the number of  trainable parameters from $mn$ to $r(m+n)$, making the rank $r$ a key hyperparameter for controlling client-side computational and communication costs in federated fine-tuning.

Various strategies have been adopted in the literature to combine \algname{LoRA} with \texttt{FL}. Some works directly apply the \algname{FedAvg} algorithm \citep{flmcmahan} to the \algname{LoRA} components $\mathbf{A}$ and $\mathbf{B}$ and average them on the server as $\mathbf{A}=\sum_i w_i\mathbf{A}_i$ and $\mathbf{B}=\sum_i w_i\mathbf{B}_i$, where $w_i$ is the aggregation weight assigned to client $i$ \citep{FedIT}. However, noting that \algname{LoRA} models the adapter update as $\Delta \mathbf{W = s\mathbf{B}\mathbf{A}}$, this aggregation strategy suffers from  an \textbf{\emph{aggregation noise}} \citep{Flexlora, flora2} rooted in the fact that:

\begin{align}\label{eq:nonequality}
    \big(\sum_i w_i \mathbf{B}_i \big)\big(\sum_i w_i\mathbf{A}_i \big) \neq \sum_i w_i \Delta \mathbf{W}_i= \sum_i w_i (s_i \mathbf{B}_i  \mathbf{A}_i).
\end{align}

To address the aggregation noise, existing methods either 1) alternatingly update $\mathbf{A}_i$ and $\mathbf{B}_i$ (between clients' local optimization steps or between communication rounds) and aggregate them separately \citep{RoLoRA, LALoRA} (this requires uniform ranks across clients), or 2) simultaneously update both and aggregate the resulting adapter updates $\{\Delta \mathbf{W}_i\} = \{s_i \mathbf{B}_i \mathbf{A}_i\}$ \citep{Flexlora, flora2}. 

We demonstrate that both the above methods are sub-optimal. The first approach wastes significant client computation and communication on fine-tuning/transmitting the $\mathbf{A}_i$ matrices, which we show, for Gaussian-initialized \algname{LoRA}, function more as random gradient perturbations than as essential knowledge-storing parameters. The second approach, while solving the aggregation noise problem, suffers from a \textbf{\emph{gradient coupling}} issue between the $\mathbf{A}$ and $\mathbf{B}$ matrices: the current gradient update of $\mathbf{A}$ depends on the current $\mathbf{B}$, and vice versa. Hence, when fine-tuned \emph{simultaneously}, one matrix adapts to a basis that the other has already shifted away from and this induces the \textbf{\emph{gradient coupling noise}} \citep{LALoRA}. While the aforementioned alternate optimization of the two matrices can avoid this noise \citep{LALoRA, RoLoRA}, it spends half of the computation/communication resources of each client $i$ on fine-tuning/transmitting its $\mathbf{A}_i$ matrix. By avoiding the noise via freezing matrix $\mathbf{A}_i$, our proposed method spends all computation and communication resources of the client on matrix $\mathbf{B}_i$. 

Freezing client matrices $\mathbf{A}_i$ to their initial value saves computational and communication resources at the cost of a lower model utility \citep{RoLoRA, LALoRA}. To balance efficiency and utility, we propose resampling and freezing $\mathbf{A}_i$ matrices at each round. This approach avoids the gradient coupling noise found in existing resampling strategies—like \algname{FLoRA}~\cite{flora2}—where both matrices are fine-tuned simultaneously. Consequently, our method cuts clients' overhead by half, eliminates gradient interference between $\mathbf{A}_i$ and $\mathbf{B}_i$, and mitigates the low-rank limitations of \algname{LoRA}.

Finally, we demonstrate that per-round resampling and freezing of $\mathbf{A}_i$ approximates full fine-tuning under rank-dependent gradient perturbations, characterized by gradient scaling and additive noise. The signal-to-noise ratio (\texttt{SNR}) of perturbed gradients increases with the adaptation rank $r$. Therefore, clients with heterogeneous ranks experience varying levels of gradient perturbation. This variance renders a uniform learning rate suboptimal. We thus also propose rank-dependent learning rate adjustments to compensate for this heterogeneity and achieve further performance improvement.

Our main contributions are summarized below:

\begin{itemize}[leftmargin=*]
    \item We introduce \algname{SPRA-LoRA}, a federated fine-tuning framework for rank-heterogeneous environments. By resampling and freezing client $\mathbf{A}_i$ matrices, it cuts computation and uplink communication costs by 50\% while mitigating LoRA's low-rank constraints.

    \item \algname{SPRA-LoRA} employs rank-aware learning rates for clients to account for their different levels of gradient perturbation. This tailored approach stabilizes training, leading to faster convergence and superior performance compared to the existing methods using uniform learning rates.
    
    \item We provide a convergence proof for \algname{SPRA-LoRA} in non-convex settings, accounting for variations in rank, data and system resources. Our analysis demonstrates that \algname{SPRA-LoRA} is fundamentally equivalent to performing \algname{Fed-Avg} on adapters with noisier batch gradients. 

    \item Experimental results across multiple models and language tasks show that \algname{SPRA-LoRA} improves utility and convergence speed with 50\% less computation and uplink communication overhead on clients compared to state-of-the-art baselines.
\end{itemize}

\section{Related Works}
\algname{LoRA} is a prominent \texttt{PEFT} method for federated fine-tuning \citep{fedpetuning}. Recent work highlights the asymmetry between its component matrices $\mathbf{A}$ and $\mathbf{B}$: \citet{symmetry2024} evaluates freezing $\mathbf{A}$ versus $\mathbf{B}$, while \citet{lora+} improves performance using matrix-specific learning rates. Recent work by \citet{lee2026learningratemattersvanilla} demonstrates that optimizing the learning rate significantly boosts the performance of \algname{LoRA}. \citet{kalajdzievski2023rankstabilizationscalingfactor} studies the
impact of the scaling factor $s$ on the learning process of \algname{LoRA}. 

Existing \texttt{Fed-LoRA} research follows two categories, and \Cref{tab:fedlora_algs} summarizes their key features:

\begin{table*}[t!]
    \scriptsize
    \caption{Comparison of \texttt{Fed-LoRA} algorithms. \ding{53}: NO. $N_i$ is dataset size of client $i$; $n^e$ is the total clients in round $e$. \textcolor{red}{Red} rows indicate algorithms for uniform ranks; \textcolor{blue}{blue} for heterogeneous ranks.}
    \centering
    \begin{tabular}{lcccc}
    \toprule 
   \bf algorithm & \bf aggregation strategy &
   \bf freezing $\{\mathbf{A}_i\}$ or $\{\mathbf{B}_i\}$ & \bf resampling $\{\mathbf{A}_i$\} & \bf non-uniform learning rate\\
    \midrule
    \bf  \algname{\textcolor{red}{SLoRA}}[\citeyear{Slora}] & \textcolor{red}{$\{\mathbf{A}_i\}$ and $\{\mathbf{B}_i\}$ separately} & 
    \textcolor{red}{\ding{53}} & \textcolor{red}{\ding{53}} & \textcolor{red}{\ding{53}}\\
    
    \bf \algname{\textcolor{blue}{HetLoRA}} [\citeyear{HetLoRA}] & \textcolor{blue}{$\{\mathbf{A}_i\}$ and $\{\mathbf{B}_i\}$ separately} & 
    \textcolor{blue}{\ding{53}} & \textcolor{blue}{\ding{53}} & \textcolor{blue}{\ding{53}}\\
    
    \bf \algname{\textcolor{red}{FeDeRA}} [\citeyear{Federa}] & \textcolor{red}{$\{\mathbf{A}_i\}$ and $\{\mathbf{B}_i\}$ separately} & 
    \textcolor{red}{\ding{53}} & \textcolor{red}{\ding{53}} & \textcolor{red}{\ding{53}}\\
    
    \bf \algname{\textcolor{red}{FedIT}} [\citeyear{FedIT}] & \textcolor{red}{$\{\mathbf{A}_i\}$ and $\{\mathbf{B}_i\}$ separately} & 
    \textcolor{red}{\ding{53}} & \textcolor{red}{\ding{53}} & \textcolor{red}{\ding{53}}\\
    
    \bf \algname{\textcolor{blue}{FlexLoRA}} [\citeyear{Flexlora}] & \textcolor{blue}{$\{s_i \mathbf{B}_i  \mathbf{A}_i$\}} & 
    \textcolor{blue}{\ding{53}} & \textcolor{blue}{\ding{53}} & \textcolor{blue}{\ding{53}}\\
    
    \bf \algname{\textcolor{blue}{FLoRA}} [\citeyear{flora2}] & \textcolor{blue}{$\{s_i \mathbf{B}_i  \mathbf{A}_i$\}} & 
    \textcolor{blue}{\ding{53}} & \textcolor{blue}{every round} & \textcolor{blue}{\ding{53}}\\
    
    \bf \algname{\textcolor{red}{FLASC}} [\citeyear{Flassc}] & \textcolor{red}{$\{\mathbf{A}_i\}$ and $\{\mathbf{B}_i\}$ separately} & 
    \textcolor{red}{\ding{53}} & \textcolor{red}{\ding{53}} & \textcolor{red}{\ding{53}}\\
    
    \bf \algname{\textcolor{red}{FFA-LoRA}} [\citeyear{FFA}] & \textcolor{red}{only $\{\mathbf{B}_i$\}} & 
    \textcolor{red}{\textcolor{red}{  only $\{\mathbf{A}_i$\}}} & \textcolor{red}{\ding{53}} & \textcolor{red}{\ding{53}}\\
    
    
    \bf \algname{\textcolor{red}{RoLoRA}} [\citeyear{RoLoRA}] & \textcolor{red}{$\{\mathbf{A}_i\}$ and $\{\mathbf{B}_i\}$ alternately} & 
    \textcolor{red}{$\{\mathbf{A}_i\}$ and $\{\mathbf{B}_i\}$ alternately} & \textcolor{red}{\ding{53}} & \textcolor{red}{\ding{53}}\\
    
    \bf \algname{\textcolor{blue}{Replicate}} [\citeyear{replication}] & \textcolor{blue}{$\{\mathbf{A}_i\}$ and $\{\mathbf{B}_i\}$ separately} & 
    \textcolor{blue}{\ding{53}} & \textcolor{blue}{\ding{53}} & \textcolor{blue}{\ding{53}}\\
    
    \bf \algname{\textcolor{red}{FedSA}} [\citeyear{Fedsa}] & \textcolor{red}{only $\{\mathbf{A}_i$\}} & 
    \textcolor{red}{\ding{53}} & \textcolor{red}{\ding{53}} & \textcolor{red}{\ding{53}}\\
    
    \bf \algname{\textcolor{red}{LoRA-A$^2$}} [\citeyear{LoRAA2}] & \textcolor{red}{$\{\mathbf{A}_i\}$ and $\{\mathbf{B}_i\}$ alternately} & 
    \textcolor{red}{ $\{\mathbf{A}_i\}$ and $\{\mathbf{B}_i\}$ alternately} & \textcolor{red}{\ding{53}} & \textcolor{red}{\ding{53}}\\
    
    \bf \algname{\textcolor{red}{EcoLoRA}} [\citeyear{Ecolora}] & \textcolor{red}{$\{\mathbf{A}_i\}$ and $\{\mathbf{B}_i\}$ separately} & 
    \textcolor{red}{\ding{53}} & \textcolor{red}{\ding{53}} & \textcolor{red}{\ding{53}}\\
    
    \bf \algname{\textcolor{blue}{FSLoRA}} [\citeyear{Fslora}] & \textcolor{blue}{$\{\mathbf{A}_i\}$ and $\{\mathbf{B}_i\}$ separately} & 
    \textcolor{blue}{\ding{53}} & \textcolor{blue}{\ding{53}} & \textcolor{blue}{\ding{53}}\\
    
    \bf \algname{\textcolor{red}{LA-LoRA}} [\citeyear{LALoRA}] & \textcolor{red}{$\{\mathbf{A}_i\}$ and $\{\mathbf{B}_i\}$ separately} & 
    \textcolor{red}{$\{\mathbf{A}_i\}$ and $\{\mathbf{B}_i\}$ alternately} & \textcolor{red}{\ding{53}} & \textcolor{red}{\ding{53}}\\
    
    \midrule
    \bf \algname{\textcolor{blue}{SPRA-LoRA}} & \textcolor{blue}{$\{s_i\mathbf{B}_i  \mathbf{A}_i\}$} & 
    \textcolor{blue}{$ $ only $\{\mathbf{A}_i\}$} & \textcolor{blue}{  every round} & \textcolor{blue}{non-uniform}\\
    \bottomrule
    \end{tabular}
    \label{tab:fedlora_algs}
\end{table*}

\subsection{Federated \algname{LoRA} with homogeneous client ranks}

The first category of research assumes homogeneous resources where all $n$ clients use a uniform rank $r$, with matrices $\{\mathbf{A}_i\} \in \mathbb{R}^{r \times m}$ and $\{\mathbf{B}_i\} \in \mathbb{R}^{n \times r}$. Early methods like \algname{SLoRA} \citep{Slora}, \algname{FedIT} \citep{FedIT}, and \algname{FeDeRA} \citep{Federa} independently apply \algname{FedAvg} to these components. However, this causes client-side gradient coupling noise from simultaneous update of the components \citep{LALoRA} and server-side aggregation noise \citep{Flexlora, flora2}. To reduce coupling, \algname{LA-LoRA} \citep{LALoRA} and \algname{RoLoRA} \citep{RoLoRA} use alternating optimization, though separate component aggregation still incurs noise. \algname{LoRA-A$^2$} \citep{LoRAA2} extends this with mask-based adaptive rank selection. Conversely, \algname{FFA} \citep{FFA} and \algname{LoRA-FA} \citep{LoRAFA} freeze $\{\mathbf{A}_i\}$ to aggregate only $\{\mathbf{B}_i\}$ for improved stability at the cost of model utility. While \algname{FedSA-LoRA} \citep{Fedsa} aggregates only $\{\mathbf{A}_i\}$—claiming it learns general global knowledge—we show that $\mathbf{A}_i$ acts primarily as a random projection for gradients rather than a knowledge-capturing parameter. Finally, \algname{EcoLoRA} \citep{Ecolora} and \algname{FLASC} \citep{Flassc} focus on communication efficiency via segment sharing and sparsity.

The aforementioned algorithms aggregate $\{\mathbf{A}_i\}$ and $\{\mathbf{B}_i\}$ separately, making them limited to homogeneous ranks and vulnerable to the aggregation noise \citep{Flexlora, flora2}.

\subsection{Federated \algname{LoRA} with heterogeneous client ranks}

Varied client resources often require heterogeneous ranks $r_i$, causing dimension mismatches in $\{\mathbf{A}_i\} \in \mathbb{R}^{r_i \times m}$ and $\{\mathbf{B}_i\} \in \mathbb{R}^{n \times r_i}$ that prevent direct component aggregation. To resolve this, \algname{HetLoRA} \citep{HetLoRA} uses zero-padding with sparsity-weighted aggregation, while \cite{replication} employs replication-based padding; both require matrix truncation for redistribution. \algname{FSLoRA} \citep{Fslora} uses a fixed global rank and separate aggregation of components on the server side, and each client uses sketching to update submatrices of the global \algname{LoRA} modules. While addressing rank heterogeneity, the fixed global rank limits \algname{FSLoRA}'s rank heterogeneity. Aggregating components separately and fine-tuning them simultaneously, all three methods suffer from the aggregation and coupling noises. \algname{FlexLoRA} \citep{Flexlora} and \algname{FLoRA} \citep{flora2} avoid the aggregation noise by aggregating reconstructed updates $\{\Delta \mathbf{W}_i = s_i\mathbf{B}_i \mathbf{A}_i\} \in \mathbb{R}^{n \times m}$. \algname{FlexLoRA} then redistributes components through computationally expensive truncated \texttt{SVD} of the aggregated update. \algname{FLoRA} implements the update aggregation by stacking clients' \algname{LoRA components} and avoids the \texttt{SVD} by merging the aggregated update into clients' local adapters and reinitializing both the \algname{LoRA} matrices for the next round. Yet, both \algname{FLoRA} and \algname{FlexLoRA} still suffer from the coupling noise, due to simultaneous fine-tuning of \algname{LoRA} components.

In contrast, by treating $\{\mathbf{A}_i\}$ as periodically resampled, frozen random projection matrices and fine-tuning only $\{\mathbf{B}_i\}$, \algname{SPRA-LoRA} totally eliminates the gradient coupling noise as well as the aggregation noise, while halving clients' overhead. This random-$\mathbf{A}$ framework paves the way to precisely quantify a rank-dependent gradient perturbation existing in \algname{LoRA} dynamics, leading to a theoretically grounded strategy where clients adapt their learning rates based on their specific ranks. Unlike heuristic alternatives, our approach accelerates convergence and improves model utility.

\section{Notations}

Bold uppercase and lowercase letters denote matrices and vectors. For $\mathbf{A} \in \mathbb{R}^{r \times m}$, $\|\mathbf{A}\|_F$ is the Frobenius norm. $\mathbb{I}_k$ is the $k \times k$ identity, and $\mathbf{0}_{k_1 \times k_2}$ is the zero matrix. $\mathcal{N}(\mathbf{0}, \mathbb{I}_{k_1 \times k_2})$ denotes a $k_1 \times k_2$ matrix with i.i.d. standard normal entries.

In a \texttt{FL} system of $n$ clients (dataset $\mathcal{D}_i$, size $N_i$), $\mathcal{S}^e$ is the set of $n^e$ clients participating in round $e \in \{1, \dots, E\}$. Local objective function of client $i$ is $f_i(\mathbf{W})=\frac{1}{N_i}\sum_{x\in \mathcal{D}_i}l(x;\mathbf{W})$, with minimum value $f_i^*$. Starting from $\mathbf{W}_i^{e,0} = \mathbf{W}^{e-1}$, client $i \in \mathcal{S}^e$ performs $T$ local optimization steps (indexed by $t$). We denote adapter batch gradients with $\nabla_{\mathbf{W}_i^{e,t}}\mathcal{L}_i^t(\mathbf{W}_i^{e,t}) = \frac{1}{b_i}\sum_{x \in \mathcal{B}_i^{e,t}}\nabla_{\mathbf{W}_i^{e,t}}l(x; \mathbf{W}_i^{e,t})$, where $l$ is the used loss function and $b_i$ is the batch size. Server aggregates the resulting adapter updates and updates the global adapter as $\mathbf{W}^e = \mathbf{W}^{e-1} + \sum_{i \in \mathcal{S}^e} \Delta \mathbf{W}_i^{e}/n^e$.

\section{\algname{LoRA} randomly perturbs adapter batch gradients}\label{sec:perturbation}

\begin{figure*}[t]
\centering
 \includegraphics[width=0.9\columnwidth]{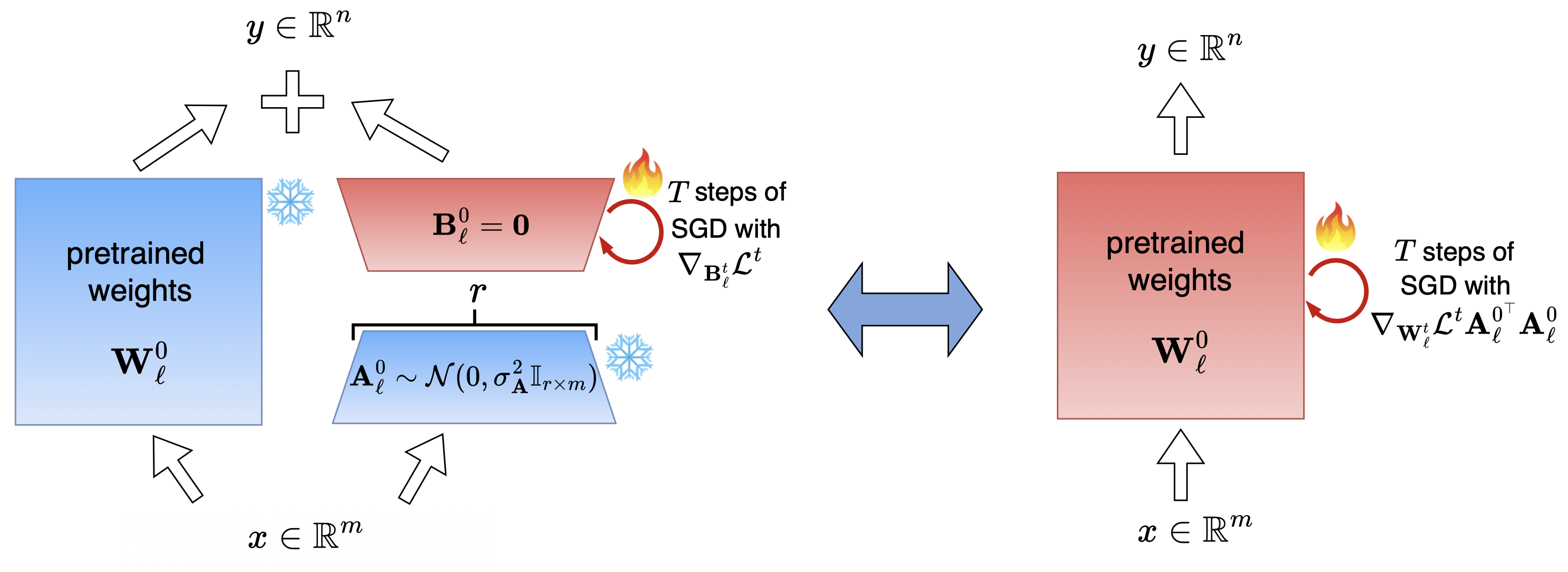}
 \caption{\algname{LoRA} with frozen component $\mathbf{A}_\ell$. \textbf{Left:} Standard \algname{LoRA} with a frozen $\mathbf{A}_\ell$ and fine-tuned $\mathbf{B}_\ell$. \textbf{Right:} Equivalent full-weight fine-tuning of $\mathbf{W}_\ell^0$ using structured noisy gradients $\nabla_{\mathbf{W}_\ell^t} \mathcal{L}^t \mathbf{A}_\ell^{0\top} \mathbf{A}_\ell^0$. Scaling factor $s$ is not shown in the figure for simplicity.}
\label{fig:lora}
\end{figure*}

To motivate the proposed \algname{SPRA-LoRA} algorithm, we analyze the structural mechanism of \algname{LoRA}. This section is summarized in \Cref{fig:lora}. Consider a pretrained model with layers $\ell \in \{1, \dots, L\}$ and weights $\mathbf{W}_\ell^0 \in \mathbb{R}^{n \times m}$. For a low rank $r$, \algname{LoRA} introduces component matrices $\mathbf{B}_\ell \in \mathbb{R}^{n \times r}$ and $\mathbf{A}_\ell \in \mathbb{R}^{r \times m}$ in the $\ell$-th adaptation layer of the model, defining the forward pass as:
\begin{align}\label{eq:forward_short}
    y = (\mathbf{W}_\ell^0 + s\mathbf{B}_\ell^t \mathbf{A}_\ell^t)x = \mathbf{W}_\ell^0 x + s\mathbf{B}_\ell^t \mathbf{A}_\ell^t x.
\end{align}
Standard \algname{LoRA} implementations, e.g., Hugging Face \citep{HF}, initialize $\mathbf{B}_\ell^0$ with an all-zero matrix and sample elements of $\mathbf{A}_\ell^0$ from $\mathcal{N}(0, \sigma_{\mathbf{A}}^2)$, with $\sigma_{\mathbf{A}}^2 = 1/r^2$. Also, $s=\alpha/r^\nu$, with $\nu=1$ for regular \algname{LoRA} and $\nu=0.5$ for rank-stabilized \algname{LoRA} \citep{kalajdzievski2023rankstabilizationscalingfactor}. 

Prior work suggests that while $\mathbf{B}_\ell$ is updated significantly, the randomly initialized $\mathbf{A}_\ell^0$ remains nearly frozen during fine-tuning \citep{flora}. On the other hand, at every optimization step, the gradient of the batch loss $\mathcal{L}^t$ with respect to the current $\mathbf{B}_\ell^t$ becomes a projection of the full-weight gradient:
\begin{align}\label{eq:down_projectionB}
\color{blue}{\nabla_{\mathbf{B}_{\ell}^t} \mathcal{L}^t} &=  \frac{\partial \mathcal{L}^t}{\partial s\mathbf{B}_{\ell}^t\mathbf{A}_{\ell}^t} \cdot \frac{\partial s\mathbf{B}_{\ell}^t\mathbf{A}_{\ell}^t}{\partial \mathbf{B}_{\ell}^t} = \frac{\partial \mathcal{L}^t}{\partial y} \cdot \frac{\partial y}{\partial s\mathbf{B}_{\ell}^t\mathbf{A}_{\ell}^t} \cdot s\mathbf{A}_{\ell}^{t^\top} = \frac{\partial \mathcal{L}^t}{\partial y} \cdot x^\top \cdot s\mathbf{A}_{\ell}^{t^\top} = \color{blue}{(\nabla_{\mathbf{W}_{\ell}^t} \mathcal{L}^t) (s\mathbf{A}_{\ell}^t)^{\top}}.
\end{align}
In other words, every batch gradient computation w.r.t. the component matrix $\mathbf{B}_\ell^t$ at time step $t$ is equal to projection of batch gradient $\nabla_{\mathbf{W}_{\ell}^t} \mathcal{L}^t$, w.r.t. the equivalent forward-pass adapter $\mathbf{W}_\ell^t=  \mathbf{W}_\ell^0+ s\mathbf{B}_\ell^t\mathbf{A}_\ell^t$, with $(s\mathbf{A}_{\ell}^t)^{\top}$. Similarly, \textcolor{blue}{$\nabla_{\mathbf{A}_{\ell}^t} \mathcal{L}^t = (s\mathbf{B}_{\ell}^t)^\top(\nabla_{\mathbf{W}_{\ell}^t} \mathcal{L}^t)$}. Therefore, assuming freezing $\mathbf{A}_\ell^0$, after $T$ local \texttt{SGD} steps with learning rate $\eta$ for fine-tuning $\mathbf{B}_\ell^0$, the resulting adapter weight is:
\begin{align}\label{eq:adapterupdate}
    \mathbf{W}_\ell^T=  \mathbf{W}_\ell^0+ s
    \mathbf{B}_\ell^T\mathbf{A}_\ell^0 = \mathbf{W}_\ell^0 - s\eta \sum_{t=0}^{T-1} \nabla_{\mathbf{B}_{\ell}^t} \mathcal{L}^t\mathbf{A}_\ell^0 = \mathbf{W}_\ell^0 - \eta \sum_{t=0}^{T-1} \nabla_{\mathbf{W}_{\ell}^t} \mathcal{L}^t (s\mathbf{A}_\ell^0)^\top (s\mathbf{A}_\ell^0).
\end{align}

This formulation reveals that \algname{LoRA} with frozen $\{\mathbf{A}_\ell^0\}_{\ell=1}^L$ is equivalent to full fine-tuning $\{\mathbf{W}_\ell^0\}_{\ell=1}^L$ with \emph{randomly perturbed} batch gradients, where $s^2\mathbf{A}_\ell^{0^\top} \mathbf{A}_\ell^0$ acts as a structural perturbation. While derived under a frozen-$\mathbf{A}$ assumption, we show in \Cref{lem:lora_final_full_forwardpass} (Appendix) that this interpretation holds within $\mathcal{O}(\eta^3)$ for the original \algname{LoRA}. Along with the lemma, this shows that \emph{$\mathbf{A}_\ell$ matrices function more as random low-rank perturbation matrices than essential knowledge-capturing parameters.}

\subsection{Analysis of the perturbed batch gradients in \cref{eq:adapterupdate}}\label{sec:perturbation_analysis}

The columns of $\mathbf{A}_\ell^{0\top}$ are $m$-dimensional Gaussian variables sampled from $\mathcal{N}(\mathbf{0}, \sigma_{\mathbf{A}}^2\mathbb{I}_{m \times m})$, and the product $\mathbf{A}_\ell^{0\top}\mathbf{A}_\ell^{0}$ follows a Wishart distribution with $r$ degrees of freedom. For any statistically independent vector $\mathbf{q} \in \mathbb{R}^{1 \times m}$, the term $\mathbf{q}(s^2\mathbf{A}_\ell^{0\top}\mathbf{A}_\ell^{0})$ represents a weighted sum of chi-squared variables. Based on the Lindeberg's Central Limit Theorem (\Cref{thm:Lindeberg_CLT}), this sum approximately follows a Gaussian distribution, a property we leverage to characterize the gradient perturbation.

\begin{restatable}{lemma}{noisevar}\label{lem:noise_var}
Consider a fixed $\mathbf{q} \in \mathbb{R}^{1 \times m}$ and $\mathbf{A}_\ell^0 \in \mathbb{R}^{r \times m}\sim \mathcal{N}(0, \sigma_{\mathbf{A}}^2\mathbb{I}_{r \times m})$, with $\sigma_{\mathbf{A}}^2 = 1/r^\mu$ $(2\nu+\mu> 0.5)$. Random perturbation $\mathbf{q}(s^2\mathbf{A}_\ell^{0\top}\mathbf{A}_\ell^0)$ approaches the $m$-dimensional Gaussian $\mathcal{N}\big(\alpha^2 \mathbf{q}/r^{2\nu+\mu-1}, \alpha^4(\|\mathbf{q}\|^2 \mathbb{I}_m + \mathbf{D})/r^{4\nu+2\mu-1}\big)$ as $r$ increases, where $\mathbf{D} = \text{diag}(\mathbf{q}^2)$.
\end{restatable}

The above lemma shows that, for a fixed $\mathbf{q}$, the term $\mathbf{q}(s^2\mathbf{A}_\ell^{0\top}\mathbf{A}_\ell^0)$ is a \emph{biased} perturbation of $\mathbf{q}$ with a distribution centered at mean $\alpha^2 \mathbf{q}/r^{2\nu +\mu - 1}$. Also, it has the following signal-to-noise ratio (\texttt{SNR}): 

\begin{align}\label{eq:linearsnr}
    \forall~ \alpha, \nu, \mu: \texttt{SNR}\big(\mathbf{q}(s^2\mathbf{A}_\ell^{0\top}\mathbf{A}_\ell^0)\big) = \frac{\|\frac{\alpha^2 \mathbf{q}}{r^{2\nu +\mu - 1}}\|^2}{Tr\big(\frac{\alpha^4}{r^{4\nu+2\mu-1}}(\|\mathbf{q}\|^2 \mathbb{I}_m + \mathbf{D})\big)} = \frac{\|\frac{\alpha^2 \mathbf{q}}{r^{2\nu +\mu- 1}}\|^2}{\frac{\alpha^4(m+1)\|\mathbf{q}\|^2}{r^{4\nu +2\mu-1}}} = \frac{r}{m+1}.
\end{align}

As $\nu$ decreases, the scaling effect decreases, while the additive noise magnitude increases. For example, at $\nu=-0.5$, the perturbation is \emph{unbiased} with asymptotic distribution $\mathcal{N}\big(\mathbf{q}, (\|\mathbf{q}\|^2 \mathbb{I}_m + \mathbf{D})/r\big)$ (see \Cref{fig:lora_scaling_v-0.5} in the Appendix). Previous work \cite{kalajdzievski2023rankstabilizationscalingfactor} showed that rank-stabilized \algname{LoRA} with $\nu=0.5$, i.e., $s=\alpha/r^{0.5}$, enhances performance in high-rank scenarios compared to regular \algname{LoRA} with $\nu=1$. 
\begin{wrapfigure}{r}{0.4\textwidth}
\vspace{-10pt}
\centering
\includegraphics[width=0.329\columnwidth]{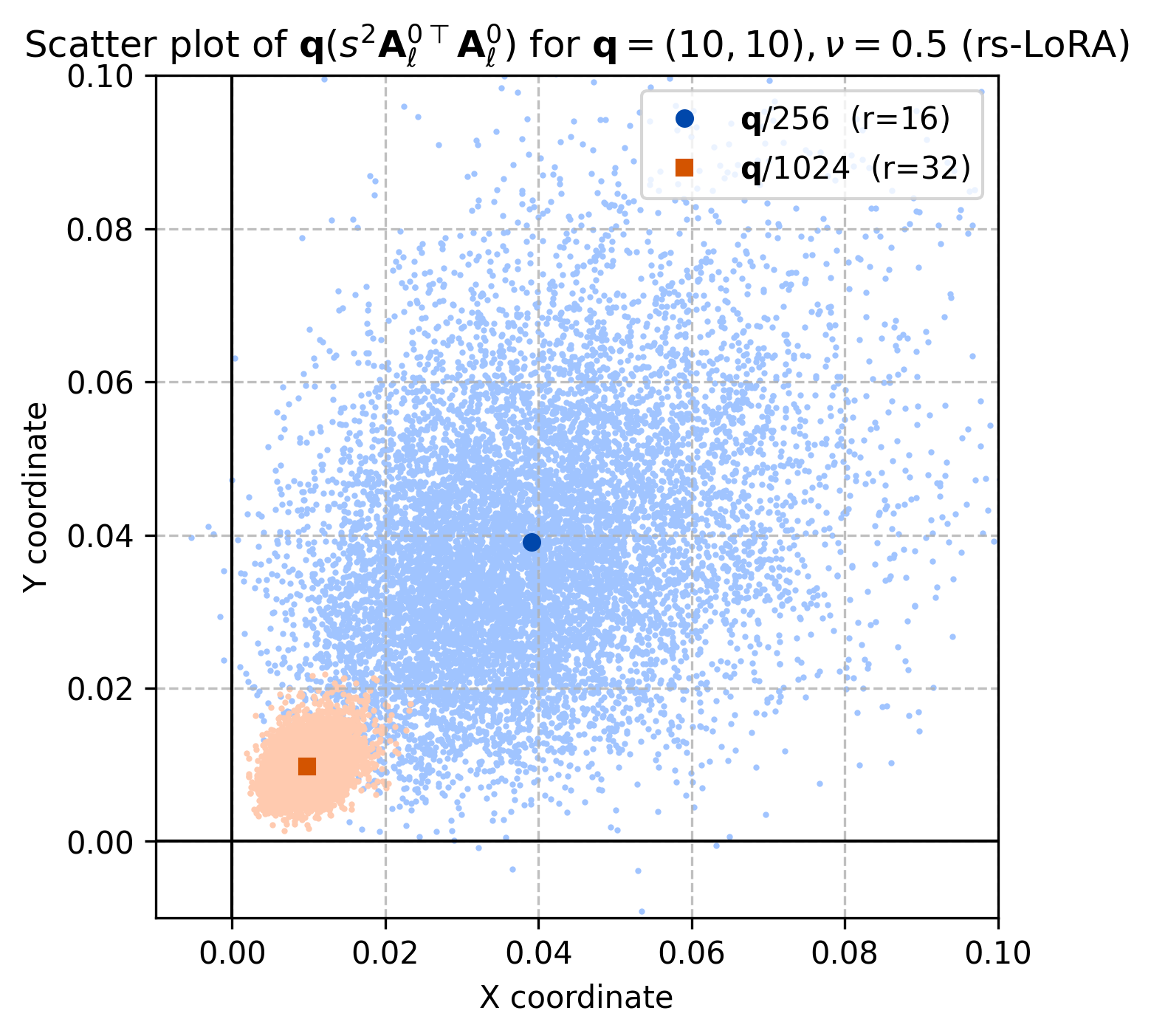}
\vspace{-3pt}
 \caption{Scatter plot of $\mathbf{q}(s^2\mathbf{A}_\ell^{0\top}\mathbf{A}_\ell^0)$, for example $\mathbf{q}=(10,10)$, $10^4$ samples $\mathbf{A}_\ell^0 \sim \mathcal{N}(0, \mathbb{I}_{r \times 2}/r^2)$ and $\nu = {0.5}$.}
\label{fig:scatterplot}
\vspace{-22pt}
\end{wrapfigure}
Accordingly, we adopt rank-stabilized \algname{LoRA} with $\alpha=1$. Also, let us use $\mu=2$, following the Gaussian initialization of Hugging Face \citep{HF}, resulting in distribution $\mathcal{N}\big(\mathbf{q}/r^2, (\|\mathbf{q}\|^2 \mathbb{I}_m + \mathbf{D})/r^{5}\big)$. We have visualized this in \Cref{fig:scatterplot} for an example $\mathbf{q}=(10,10)$. In Appendix \ref{app:app_kaiming_init}, we show that the above rank-dependent perturbation of gradients similarly exists for Kaiming-Uniform initialization \citep{kaiming} of $\{\mathbf{A}_\ell^0\}_{\ell=1}^L$ (\texttt{PEFT} default), with a rank-linear \texttt{SNR}.

\begin{figure*}[t]
\centering
\includegraphics[width=0.47\columnwidth]{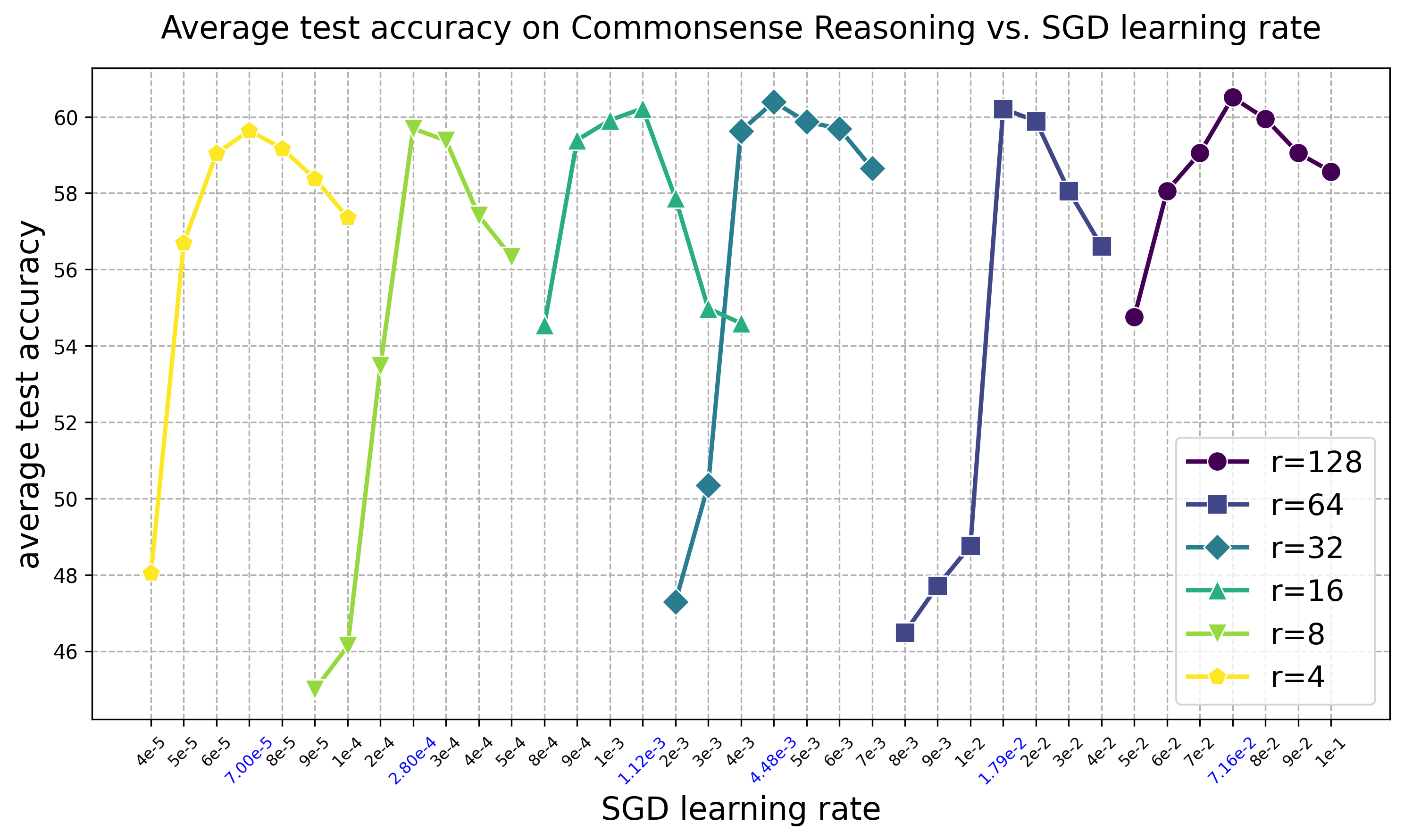}
 \includegraphics[width=0.47\columnwidth]{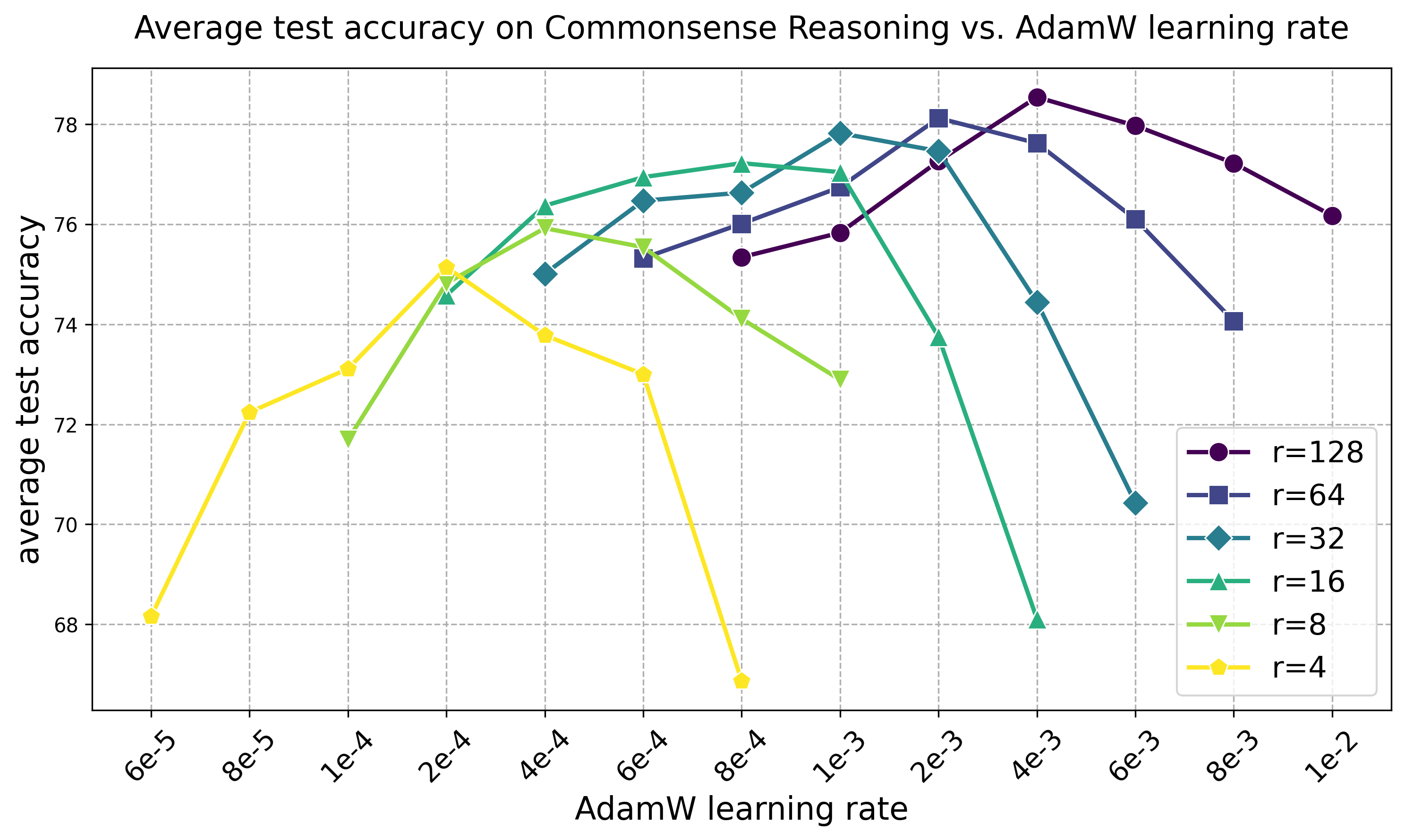}
 \caption{Average test accuracy for \texttt{LLaMA-3.2-3B} on Commonsense Reasoning obtained with \texttt{SGD} optimizer (left) and \texttt{AdamW} optimizer (right) with frozen and resampled $\mathbf{A}$.}
\label{fig:ranklr}
\vspace{-1em}
\end{figure*}

While the previous lemma assumes independence between $\mathbf{q}$ and $\mathbf{A}_\ell^{0}$, in \cref{eq:adapterupdate}, the gradient $\nabla_{\mathbf{W}_\ell^t} \mathcal{L}^t$ ($t \geq 1$) depends on $\mathbf{A}_\ell^{0}$ (weakly). Given the \emph{few local steps per round}, we use the lemma to approximate the distribution of the perturbed gradients in \cref{eq:adapterupdate}\footnote{For completeness, we show in Appendix \ref{app:meanshift} that the true distribution of the perturbed gradients introduces bounded mean and covariance shifts to the distribution returned by \Cref{lem:noise_var}, where the bounds decrease with fewer local steps per round ($T$) and larger rank ($r$). We also use this bound in our convergence proof.}. Consequently, the distribution of the perturbed batch gradient $\mathbf{Q} = \nabla_{\mathbf{W}_\ell^t} \mathcal{L}^t (s^2\mathbf{A}_\ell^{0\top}\mathbf{A}_\ell^0)$ approaches a Gaussian where row $i$ converges to $\mathcal{N}\big(\mathbf{g}_{i,:}/r^2, ~ (\|\mathbf{g}_{i,:}\|^2 \mathbb{I}_m + \mathbf{D})/r^5\big)$. Thus, all gradient elements share a $1/r^2$ scaling factor but have distinct noise levels.  Varying rank r causes inconsistent gradient scales and perturbation levels. As \Cref{fig:ranklr} (left) confirms, optimal \texttt{SGD} learning rates follow the square law $\eta_r^*=\eta_c r^2$ and increase with adaptation rank $r$.

\subsection{Extension of \Cref{eq:adapterupdate} to \texttt{Adam} optimizer}\label{app:adam_perturbation}
The gradient perturbation persists in optimizers like \texttt{Adam} \citep{adam} and \texttt{AdamW} \citep{adamw}, as it affects the entire optimization trajectory.
We now extend our previous analysis to \texttt{Adam} (and \texttt{AdamW}) optimizer. Under the frozen-$\mathbf{A}$ assumption ($\mathbf{B}_\ell^0 = 0$), \texttt{Adam} w.r.t $\mathbf{B}_\ell^t$ computes gradient moments as:
\begin{align}\label{eq:adam}
    \textit{first moments:}~~ &m_{-1} = 0 \to m_t = \beta_1 m_{t-1} + (1 - \beta_1) \nabla_{\mathbf{B}_{\ell}^t} \mathcal{L}^t \to \hat m_t = m_t / (1-\beta_1) \nonumber \\
    \textit{second moments:}~~ &v_{-1} = 0 \to v_t = \beta_2 v_{t-1} + (1 - \beta_2) (\nabla_{\mathbf{B}_{\ell}^t} \mathcal{L}^t \odot \nabla_{\mathbf{B}_{\ell}^t} \mathcal{L}^t) \to \hat v_t = v_t / (1-\beta_2) \nonumber \\
    \textit{update:}~~ &\mathbf{B}_\ell^{t+1} = \mathbf{B}_\ell^{t} - \frac{\eta}{\sqrt{\hat v}+\epsilon} \odot \hat m_t,
\end{align}
where $\odot$ denotes the Hadamard product. As $\nabla_{\mathbf{B}_{\ell}^t} \mathcal{L}^t = (\nabla_{\mathbf{W}_{\ell}^t} \mathcal{L}^t) (s\mathbf{A}_{\ell}^0)^{\top}$, we have $\hat m_t = \Bar{m}_t (s\mathbf{A}_{\ell}^0)^{\top}$ as well ($\Bar{m}_t$ being the first moment of $\nabla_{\mathbf{W}_{\ell}^t} \mathcal{L}^t$). Therefore, the final parameter update becomes:
$$
    \mathbf{W}_\ell^{T} = \mathbf{W}_\ell^{0} + s\mathbf{B}_\ell^T\mathbf{A}_\ell^0=\mathbf{W}_\ell^{0} - \sum_{t=0}^{T-1}\bigg(\frac{\eta}{\sqrt{\hat v_t}+\epsilon} \odot \Bar{m}_t (s\mathbf{A}_{\ell}^0)^{\top}\bigg) (s\mathbf{A}_{\ell}^0).
$$

This shows that \texttt{Adam} w.r.t $\mathbf{B}_\ell^t$ is equivalent to \texttt{Adam} w.r.t $\mathbf{W}_\ell^t$, with the first and second moments computed using perturbed/projected adapter gradients. Unlike the square rule $\eta_r^* = \eta_c r^2$ for \texttt{SGD}, \texttt{Adam}'s element-wise normalization via $\sqrt{\hat v_t}$ complicates the rank dependence (otherwise there would be only the first moments being computed using the perturbed batch gradients $\nabla_{\mathbf{W}_{\ell}^t} \mathcal{L}^t  (s\mathbf{A}_{\ell}^0)^{\top} (s\mathbf{A}_{\ell}^0)$), justifying a grid search. Yet, the optimum learning rate still increases with rank $r$ (\Cref{fig:ranklr}, right). 

\begin{figure}[H]
\centering
\includegraphics[width=0.5\columnwidth]{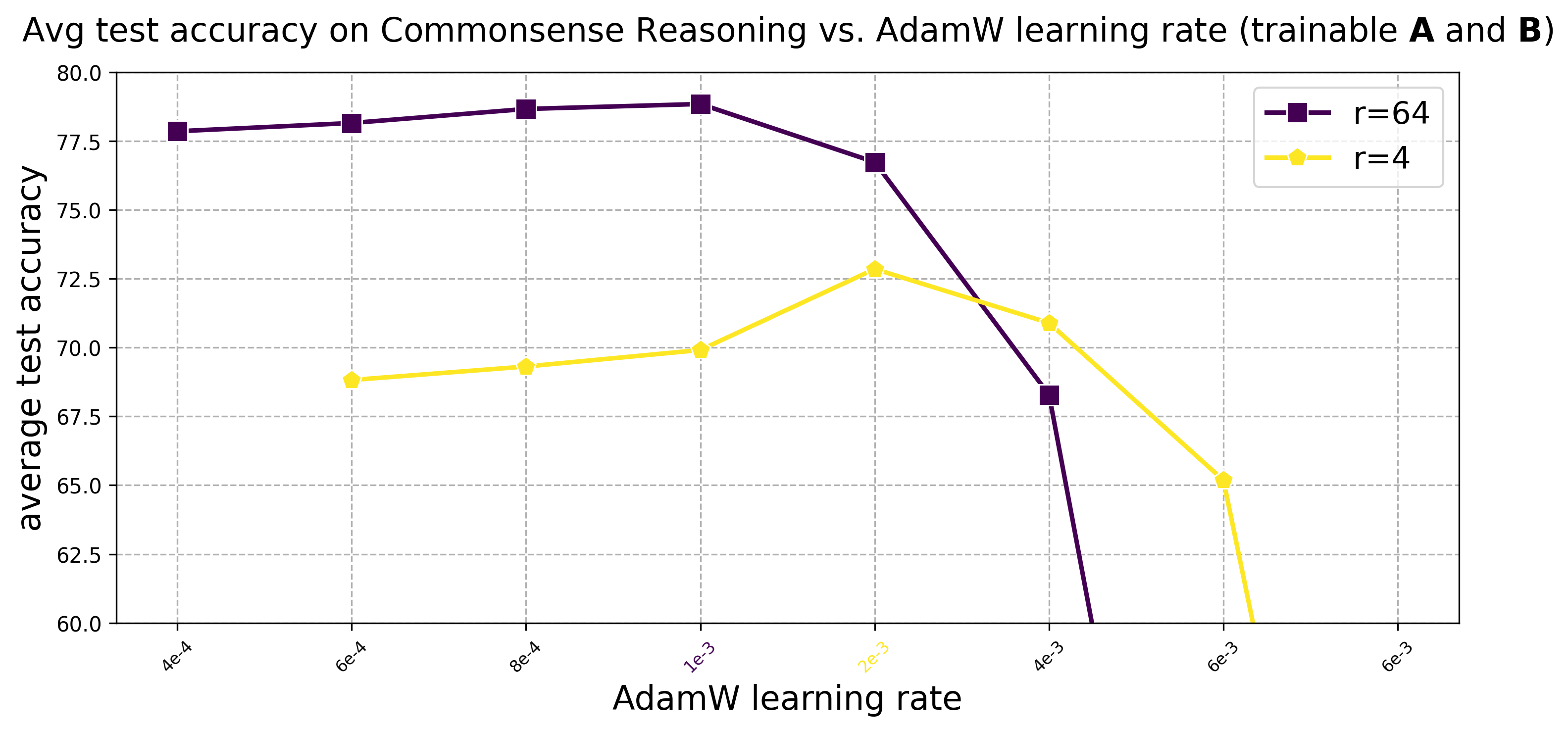}
 \caption{Average test accuracy for \texttt{LLaMA-3.2-3B} on Commonsense Reasoning obtained with \texttt{AdamW} optimizer and rank-stabilized \algname{LoRA} ($\nu=0.5$) and both $\mathbf{A}$ and $\mathbf{B}$ being trainable. Unlike when $\mathbf{A}$ is frozen (\Cref{fig:ranklr}, right), optimum learning rate \emph{decreases slowly} (at rate $r^{-1/4}$) with $r$. When $\nu=1$ (not shown), optimum learning rate does not vary (see \citep{lrvsrICML2026} for a detailed analysis).}
\label{fig:ranklr_trainA}
\vspace{-1em}
\end{figure}

Our derivations above were for frozen-$\mathbf{A}$. It is noteworthy that when both $\mathbf{A}$ and $\mathbf{B}$ are trainable, the optimal learning rate behaves differently and \emph{decreases slowly} with rank $r$ \cite{lrvsrICML2026} (see \Cref{fig:ranklr_trainA}).

We propose freezing and resampling the $\mathbf{A}$ matrices with rank-aware client learning rates. 

\section{\algname{SPRA-LoRA} algorithm}\label{sec:frank}

Our previous findings suggest that rather than applying a uniform learning rate—a standard practice in all previous \algname{Fed-LoRA} methods—clients should employ learning rates tailored to their specific ranks. Applying a single, global learning rate forces some clients to operate outside their ideal operation zone, leading to diminished performance compared to their rank-specific optimum.

\subsection{\algname{SPRA-LoRA}: Per-round sampling and freezing of $\mathbf{A}_i^{e,0}$ with rank-aware learning rates}

For simplicity, we drop the layer subscript $\ell$ in the next sections. Assuming that client $i$ with rank $r_i$ uses a rank-aware learning rate $\eta_{r_i}$, we describe the proposed \algname{SPRA-LoRA} in the following. In round $e$, a participating client $i \in \mathcal{S}^e$ initializes with the last global model as $\mathbf{W}_i^{e,0} = \mathbf{W}^{e-1}$, resamples $\mathbf{A}_i^{e,0} \sim \mathcal{N}(\mathbf{0}, \frac{1}{r_i^2}\mathbb{I}_{r_i \times m})$ and resets $\mathbf{B}_i^{e,0}=\mathbf{0}_{n \times r_i}$ (participating clients can be synchronized at this stage by using the same random seed). Freezing $\mathbf{A}_i^{e,0}$ during the round, it uploads only the updated $\mathbf{B}_i^{e,T}$ after $T$ local steps using \texttt{SGD} (or any other optimizer, e.g., \texttt{AdamW}): 

\begin{align}\label{frankffa+}
    \mathbf{W}_i^{e,T} = \mathbf{W}_i^{e,0} + s_i \mathbf{B}_i^{e,T}\mathbf{A}_i^{e,0} = \mathbf{W}_i^{e,0} - \eta_{r_i} \sum_{t=0}^{T-1} \nabla_{\mathbf{W}_i^{e,t}}\mathcal{L}_i^{e,t} (s_i \mathbf{A}_i^{{e,0}})^\top(s_i\mathbf{A}_i^{e,0})
\end{align}

So far, by freezing $\mathbf{A}_i^{e,0}$, the client's computation and uplink communication overheads (measured by the number of trainable parameters) have been cut by 50\%. Also, the gradient coupling noise has been completely removed. In order to eliminate the aggregation noise, the server aggregates $\{\Delta \mathbf{W}_i^e\}=\{s_i\mathbf{B}_i^{e,T} \mathbf{A}_i^{e,0}\}$ into $\Delta \mathbf{W}^e = \sum_{i\in \mathcal{S}^e} \Delta \mathbf{W}_i^e/n^e$ and updates the global adapter parameter as $\mathbf{W}^{e} = \mathbf{W}^{e-1} + \Delta \mathbf{W}^{e}$, which will be downloaded by clients for the next round. 

Freezing matrices $\mathbf{A}_i^{e,0}$ resembles the \algname{FFA-LoRA} algorithm\footnote{While \algname{FFA} \citep{FFA} was originally proposed for settings with homogeneous client ranks, it can be extended to heterogeneous settings by aggregating $\{\Delta \mathbf{W}_i^e = s_i \mathbf{B}_i^{e,T} \mathbf{A}_i^0\}$, instead of aggregating $\{\mathbf{B}_i^{e,T}$\}.}, but \algname{SPRA-LoRA} extends it by resampling the matrices $\mathbf{A}_i^{e,0}$ at every round $e$ during fine-tuning, which diminishes the \algname{LoRA} low-rank constraints. This approach also improves \algname{FLoRA} \citep{flora2} and \algname{FlexLoRA} \citep{Flexlora}, which update both $\mathbf{A}_i^{e,0}$ and $\mathbf{B}_i^{e,0}$ simultaneously, suffering from the gradient coupling between them and the coupling noise \citep{LALoRA}.

Assuming that there are $R$ different values of adaptation ranks in the system and that $C_j^e$ denotes the set of clients $i \in \mathcal{S}^e$ using the $j$-th rank value, the aggregation on the server can be re-written as:

\begin{align}
    \Delta \mathbf{W}^e = \sum_{i\in \mathcal{S}^e} \Delta \mathbf{W}_i^e/n^e = \sum_{j=1}^R \frac{|C_j^e|}{n^e}\bigg(\sum_{i \in C_j^e} \Delta \mathbf{W}_i^e/|C_j^e|\bigg).
\end{align}

Specifically, participating clients $i \in \mathcal{S}^e$ are partitioned into $R$ distinct clusters according to their respective adaptation ranks. Rather than applying a uniform learning rate across the entire clusters, our rank-aware \algname{SPRA-LoRA} framework ensures that clients within each cluster $j$ utilize a specific learning rate optimized to maximize fine-tuned model utility for that specific rank. The server subsequently aggregates these $R$ cluster-specific models via the averaging process defined above.

\subsection{Rank-aware learning rate selection for \algname{SPRA-LoRA}}
To determine the optimal learning rate $\eta_{r_i}$ for individual ranks, we propose the following simple strategy: Each cluster of clients with a common rank performs federated fine-tuning and an \emph{intra-cluster} evaluation over a grid of \emph{cluster-specific} learning rates. Note that this method has \emph{\textbf{no additional}} communication overhead compared to existing baselines. Additionally, when using \texttt{SGD} optimizer, the square law $\eta_r^* = \eta_c r^2$ could be used to further diminish the overhead of learning rate tuning.

\subsection{Convergence analysis of \algname{SPRA-LoRA}}
In this section, we prove the convergence of \algname{SPRA-LoRA} in non-convex settings under mild assumptions. Let us define global objective function $f(\mathbf{W}):= \sum_{i=1}^n  f_i(\mathbf{W})/n$, with minimum value $f^\star$ and minimizer $\mathbf{W}^*$. We also make some mild assumptions about the loss functions $f_i$, including their Lipschitz continuity and smoothness (see Assumptions \ref{assump:lipschitz_smooth_bounded} in the Appendix).

\begin{restatable}[\textbf{Convergence of \algname{SPRA-LoRA} with \texttt{SGD}}]{theorem}{franklora}\label{thm:franklora}
Assume using rank-stabilized \algname{LoRA} ($\nu=0.5$), assumptions \ref{assump:lipschitz_smooth_bounded}, and full participation with client $i$ using learning rate $\eta_{r_i} = \eta_c r_i^2$. If the common learning rate satisfies $\eta_c \le \dfrac{1}{4\beta T}$, then for any number of communication rounds $E\ge1$,
$$
\boxed{\;
\min_e \mathbb E\big[\normF{\nabla f(\W^e)}^2\big]
\;\le\;
\frac{2\big(f(\W^0)-f^\star\big)}{\eta_c TE}
\;+\;\Psi
\;}
$$
where $\{\W^e\}_{e=0}^{E}$ are global iterates, and $\Psi = \mathcal{O}(\eta_c)$ is a constant. Choosing
$\eta_c=\min\big(\tfrac1{4\beta T},\tfrac1{\sqrt{E}}\big)$
$$
\min_e \mathbb E\big[\normF{\nabla f(\W^e)}^2\big] = \mathcal{O}\!\left(\frac{1}{\sqrt{E}}\right),
$$
so the squared gradient norm of the global iterates converges, as $E\to\infty$, to a
stationary point of $f$, recovering the classical
$O(1/\sqrt{E})$ non-convex FedAvg rate \citep{malekmohammadi2024noiseaware, zhang2023proportional}.
\end{restatable}

\paragraph{Discussion.} The term $\Psi$ decreases with smaller $T$, more i.i.d.\ client data and larger batch sizes. Unlike prior work, e.g.\ \cite{flora2} (\algname{FLoRA}), which assumes strong convexity of $\{f_i\}$, our results hold under more realistic assumptions. Under i.i.d.\ data, our result reduces to \texttt{SGD} \citep{ghadimi2013stochastic} on $\mathbf{W}^e$, with rate $\frac{1}{\sqrt{E}}$.

\section{Experimental results}\label{sec:exps}
We evaluate our approach on language generation and understanding tasks using \texttt{LLaMA-3.2-3B} \citep{llama3} (on Commonsense Reasoning dataset \citep{commonsensereasoning}) and \texttt{RoBERTa-base} \citep{roberta} (on GLUE benchmark \citep{gluebenchmark}). Considering our \emph{heterogeneous-rank} setting, only the baselines highlighted in blue in \Cref{tab:fedlora_algs} are the main baselines comparable to \algname{SPRA-LoRA}. Yet, we extend the aggregation strategies of the recent algorithms \algname{FedSA} \cite{Fedsa}, \algname{RoLoRA} \cite{RoLoRA} and \algname{LA-LoRA} \cite{LALoRA} with zero padding (similar to \algname{HetLoRA} \cite{HetLoRA}) and use them as baselines as well. Also, we generalize the aggregation strategy of \algname{FFA} to aggregate $\{\Delta \mathbf{W}_i^e = s_i\mathbf{B}_i^{e,T}\mathbf{A}_i^{e,0}\}$ at the end of each round $e$ (instead of aggregating $\{\mathbf{B}_i^{e,T}\}$) to extend it to settings with heterogeneous client ranks. For Commonsense Reasoning (GLUE), data is partitioned across $n=20$ ($n=40$) clients with a Dirichlet distribution with $\alpha=2$ ($\alpha=0.8$) by default \citep{flora2, Flexlora, Fslora}. Client ranks are sampled uniformly and non-uniformly from $\{\texttt{4}, \texttt{8}, \dots, \texttt{128}\}$, with identical configurations applied to all baselines to ensure a fair comparison. We use a cluster with 4 \texttt{NVIDIA H100 80 GB} GPUs for running our experiments. Each experiment is run on a single GPU. Further details about our experimental setup is explained in Appendix \ref{app:expsetup}.

\begin{table*}
\vspace{-1em}
\centering
\setlength\tabcolsep{2pt}
\begin{minipage}[c]{1.0\textwidth}
\centering
\caption{Average test accuracy (5 runs) with sample \texttt{std} error bars on Commonsense Reasoning, with Dirichlet distribution ($\alpha=2$), \texttt{LLaMA-3.2-3B} model and \texttt{AdamW} optimizer. Clients' ranks follow \textbf{\emph{uniform}} distribution over $\{\texttt{4}, \texttt{8}, \dots, \texttt{128}\}$. \algname{SPRA-LoRA} and its doubled-rank variant consistently outperform baselines while maintaining 50\% less or equal client overhead, respectively.}
\label{tab:common}
\resizebox{\textwidth}{!}{
\begin{tabular}{c|c|cccccccc|c}
\toprule
\bf{Method}& Run time & \bf{BoolQ} & \bf{WinoGrande} & \bf{OBQA} & \bf{ARC-e} & \bf{ARC-c} & \bf{SIQA} & \bf{PIQA} & \bf{HellaSwag}& \bf{Avg} 
\\ 
\toprule
\algname{Base Pretrained Model}& --- & 47.28& 04.94 &52.01  &60.89 &51.05 &53.36 & 20.67 &55.98 &43.27 \\
\midrule
\algname{FFA} [\citeyear{FFA}]& 0.81h& 62.67$_{\pm 0.67}$ & 64.06$_{\pm 0.31}$ &64.66$_{\pm 1.81}$  &76.46$_{\pm 0.52}$ &60.66$_{\pm 2.15}$ &63.05$_{\pm 2.23}$ & 73.19$_{\pm 2.18}$&64.53$_{\pm1.62}$ &66.15$_{\pm 0.61}$ \\
\algname{HetLoRA} [\citeyear{HetLoRA}]& 0.91h& 65.52$_{\pm 0.27}$&70.74$_{\pm 0.84}$ &76.01$_{\pm 0.35}$ &85.27$_{\pm 0.22}$ &72.69$_{\pm 0.41}$ &73.26$_{\pm 0.63}$ &79.10$_{\pm 0.46}$ & 72.87$_{\pm 0.40}$& 74.43$_{\pm 0.32}$\\
\algname{FlexLoRA} [\citeyear{Flexlora}]& 5.11h & 65.87$_{\pm 0.52}$&71.04$_{\pm 0.40}$ &78.01$_{\pm 0.69}$ &86.56$_{\pm 0.17}$ &73.04$_{\pm 0.01}$ &72.62$_{\pm 0.17}$ &80.81$_{\pm 0.35}$ & 71.51$_{\pm 1.39}$& 74.93$_{\pm 0.14}$\\
\algname{FLoRA} [\citeyear{flora}]& 1.51h& 67.06$_{\pm 0.88}$&70.59$_{\pm 1.67}$ &75.94$_{\pm 0.42}$ &85.70$_{\pm 0.58}$ &72.33$_{\pm 1.05}$ &73.90$_{\pm 0.49}$ &80.45$_{\pm 0.63}$ & 73.02$_{\pm 0.82}$& 74.87$_{\pm 0.40}$\\
\algname{FSLoRA} [\citeyear{Fslora}]& 0.92h& 66.83$_{\pm 0.33}$&70.30$_{\pm 0.50}$ &77.14$_{\pm 0.42}$ &$\underline{87.26}_{\pm 0.36}$ &72.87$_{\pm 0.90}$ &73.90$_{\pm 0.27}$ &81.06$_{\pm 0.47}$ & 71.52$_{\pm 0.36}$& 75.11$_{\pm 0.14}$\\
\algname{FedSA} [\citeyear{Fedsa}]& 1.08h& $59.45_{\pm 9.51}$&$50.18_{\pm4.72 }$ &$62.44_{\pm 11.81}$ &$79.53_{\pm 8.11}$ &$63.48_{\pm 7.91}$ &$60.57_{\pm 10.49}$ &$67.64_{\pm 15.78}$ & $48.12_{\pm 12.03}$& $61.42_{\pm 8.81}$\\
\algname{RoLoRA} [\citeyear{RoLoRA}]& 0.91h& $66.21_{\pm 2.23}$&$73.48_{\pm 1.51 }$ &$76.94_{\pm 1.72}$ &$87.12_{\pm 0.33}$ &$73.94_{\pm 0.77}$ &$75.18_{\pm 0.89}$ &$80.68_{\pm 0.34}$ & $75.01_{\pm 0.30}$& $76.07_{\pm 0.22}$\\
\algname{LA-LoRA} [\citeyear{LALoRA}]& 0.95h& $67.06_{\pm 2.07}$&$74.19_{\pm 0.67}$ &$78.07_{\pm 0.83}$ &$\bf87.79_{\pm 0.19}$ &$\bf74.81_{\pm 0.56}$ &$75.62_{\pm 0.49}$ &$81.06_{\pm 0.91}$ & $76.21_{\pm 0.07}$& $76.84_{\pm 0.23}$\\
\midrule
\algname{SPRA-LoRA} (uniform LR)& 1.36h &66.72$_{\pm 0.99}$ & 71.27$_{\pm 1.06}$ &75.14$_{\pm 1.53}$ & 86.77$_{\pm 0.27}$ & 73.18$_{\pm 0.18}$ & 73.49$_{\pm 0.67}$ & 80.78$_{\pm 0.36}$ & 73.94$_{\pm 0.26}$& 75.16$_{\pm 0.30}$\\
\rowcolor{lightgray}\algname{SPRA-LoRA}& 1.38h& \bf 68.54$_{\pm 2.17}$& $\underline{75.90}_{\pm 0.18}$ & $\underline{78.34}_{\pm 1.40}$ &86.39$_{\pm 0.54}$ & 73.75$_{\pm 0.81}$ & $\underline{75.78}_{\pm 0.21}$ &$\underline{81.85}_{\pm 0.93}$ & $\underline{77.69}_{\pm 0.44}$& $\underline{77.28}_{\pm 0.24}$\\
\rowcolor{lightgray} \algname{SPRA-LoRA} (doubled ranks)& 1.52h& $\underline{67.59}_{\pm 0.63}$& \bf76.27$_{\pm 1.14}$ & \bf79.01$_{\pm 0.60}$ & 86.96$_{\pm 0.21}$ & $\underline{74.46}_{\pm 0.39}$ & \bf76.12$_{\pm 0.72}$ & \bf82.03$_{\pm 0.53}$ & \bf 77.80$_{\pm 0.21}$& \bf77.53$_{\pm 0.08}$\\
\bottomrule
\end{tabular}}
\end{minipage}
\hfill 
\begin{minipage}[c]{1.0\textwidth}
\centering
\caption{Average test accuracy (5 runs) on Commonsense Reasoning dataset, with Dirichlet distribution ($\alpha=2$), \texttt{LLaMA-3.2-3B} model and \texttt{AdamW} optimizer. Clients' ranks follow \textbf{\emph{log-normal}} with heavy low tail distribution over $\{\texttt{4}, \texttt{8}, \dots, \texttt{128}\}$. \algname{SPRA-LoRA} and its doubled-rank variant consistently outperform baselines while maintaining 50\% less or equal client overhead, respectively.}
\label{tab:common_lognormal}
\resizebox{\textwidth}{!}{
\begin{tabular}{c|c|cccccccc|c}
\toprule
\bf{Method}& Run time & \bf{BoolQ} & \bf{WinoGrande} & \bf{OBQA} & \bf{ARC-e} & \bf{ARC-c} & \bf{SIQA} & \bf{PIQA} & \bf{HellaSwag}& \bf{Avg} 
\\ 
\toprule
\algname{Base Pretrained Model}& --- & 47.28& 04.94 &52.01  &60.89 &51.05 &53.36 & 20.67 &55.98 &43.27 \\
\midrule
\algname{FFA} [\citeyear{FFA}]& 0.71h& 61.43$_{\pm 0.40}$ & 63.90$_{\pm 0.28}$ &61.93$_{\pm 2.04}$  &76.54$_{\pm 2.04}$ &59.52$_{\pm 1.35}$ &63.54$_{\pm 0.51}$ & 71.93$_{\pm 3.10}$&63.00$_{\pm0.93}$ &65.22$_{\pm 0.88}$ \\
\algname{HetLoRA} [\citeyear{HetLoRA}]& 0.85h& 65.40$_{\pm 1.19}$&71.93$_{\pm 1.15}$ &78.34$_{\pm 0.64}$ &\bf87.78$_{\pm 0.41}$ &\bf74.52$_{\pm 0.52}$ &75.66$_{\pm 0.91}$ &81.28$_{\pm 0.38}$ & 73.86$_{\pm 2.18}$& 76.09$_{\pm 0.25}$\\
\algname{FlexLoRA} [\citeyear{Flexlora}]& 5.01h & 66.37$_{\pm 1.91}$&72.09$_{\pm 1.57}$ &77.01$_{\pm 0.72}$ &87.07$_{\pm 0.41}$ &72.87$_{\pm 1.41}$ &73.88$_{\pm 0.42}$ &\bf81.52$_{\pm 0.71}$ & 72.99$_{\pm 0.69}$& 75.47$_{\pm 0.51}$\\ 
\algname{FLoRA} [\citeyear{flora}]& 1.42h& 67.32$_{\pm 1.51}$&72.56$_{\pm 1.23}$ &77.14$_{\pm 1.10}$ &86.34$_{\pm 0.77}$ &72.10$_{\pm 0.94}$ &74.90$_{\pm 0.62}$ &80.20$_{\pm 0.80}$ & 73.26$_{\pm 2.19}$& 75.47$_{\pm 0.40}$\\
\algname{FSLoRA} [\citeyear{Fslora}]& 0.93h& 67.83$_{\pm 0.30}$&70.33$_{\pm 0.21}$ &77.54$_{\pm 0.42}$ &87.25$_{\pm 0.07}$ &73.14$_{\pm 0.05}$ &73.30$_{\pm 0.15}$ &80.78$_{\pm 0.28}$ & 71.10$_{\pm 0.29}$& 74.87$_{\pm 0.53}$\\
\algname{FedSA} [\citeyear{Fedsa}]& 1.11h& $60.79_{\pm 6.38}$&$53.56_{\pm 3.45}$ &$63.04_{\pm 7.39}$ &$78.76_{\pm 7.01}$ &$63.79_{\pm 6.85}$ &$60.13_{\pm 5.48}$ &$63.96_{\pm 16.56}$ & $46.23_{\pm 13.81 }$& $61.28_{\pm 7.41}$\\
\algname{RoLoRA} [\citeyear{RoLoRA}]& 0.92h& $67.77_{\pm 0.43}$&$71.21_{\pm 1.22}$ &$77.81_{\pm 1.31}$ &$\underline{87.54}_{\pm 0.14}$ &$\underline{74.34}_{\pm 0.86}$ &$74.58_{\pm 1.38}$ &$80.26_{\pm 0.84}$ & $73.18_{\pm 1.46}$& $75.83_{\pm 0.86}$\\
\algname{LA-LoRA} [\citeyear{LALoRA}]& 0.96h& $67.98_{\pm 0.86}$&$71.82_{\pm 0.13}$ &$\underline{78.61}_{\pm 0.87}$ &$87.50_{\pm 0.41}$ &$74.08_{\pm 0.58}$ &$75.71_{\pm 0.81}$ &$81.04_{\pm 0.53}$ & $73.97_{\pm 3.41}$& $76.34_{\pm 0.31}$\\
\midrule
\algname{SPRA-LoRA} (uniform LR)& 1.26h &68.10$_{\pm 0.31}$ &72.22$_{\pm 1.23}$ &76.01$_{\pm 0.87}$ & 86.24$_{\pm 0.64}$ & 71.96$_{\pm 1.09}$ & 74.91$_{\pm 0.46}$ & 80.36$_{\pm 1.32}$ & 73.66$_{\pm 1.34}$& 75.24$_{\pm 0.86}$\\
\rowcolor{lightgray}\algname{SPRA-LoRA}& 1.27h& $\underline{68.40}_{\pm 0.18}$& $\underline{75.88}_{\pm 0.92}$ & 77.94$_{\pm 0.92}$ &86.40$_{\pm 0.15}$ &72.87$_{\pm 0.38}$ &\bf 76.09$_{\pm 0.44}$ &80.56$_{\pm 0.47}$ & $\underline{76.82}_{\pm 0.05}$& \bf $\underline{76.87}_{\pm 0.05}$\\
\rowcolor{lightgray} \algname{SPRA-LoRA} (doubled ranks)& 1.35h& \bf68.82$_{\pm 2.65}$& \bf76.26$_{\pm 0.29}$ & \bf79.48$_{\pm 0.61}$ & 86.49$_{\pm 0.08}$ & 73.15$_{\pm 0.73}$ & $\underline{75.99}_{\pm 0.54}$ & $\underline{81.34}_{\pm 1.30}$ & \bf 76.94$_{\pm 0.51}$& \bf77.26$_{\pm 0.54}$\\
\bottomrule
\end{tabular}}
\end{minipage}
\vspace{-1em}
\end{table*}

The Commonsense benchmark results, detailed in Tables \ref{tab:common} and \ref{tab:common_lognormal}, compare uniform and log-normal client rank distributions across three \algname{SPRA-LoRA} variants: a uniform learning rate version designed to isolate the effects of per-round resampling and weight freezing against the \algname{FFA} baseline, the standard rank-aware learning rate configuration, and a "double-rank" alternative.

Despite 50\% lower overhead, \algname{SPRA-LoRA} with a uniform learning rate matches or exceeds most baselines. Its significant performance advantage over \algname{FFA} stems from its per-round resampling strategy. Both the methods share the same number of trainable parameters, and the marginal increase in run time for \algname{SPRA-LoRA} reflects the time for per-round parameter merging, saving and reloading in our experiments, rather than increased computational complexity.

\algname{HetLoRA} suffers from both aggregation noise (due to aggregating $\mathbf{A}$ and $\mathbf{B}$ matrices separately) and coupling noise (due to simultaneous optimization of $\mathbf{A}$ and $\mathbf{B}$ matrices). While \algname{FLoRA} shares \algname{SPRA-LoRA}'s per-round resampling and resetting strategy, it still suffers from the coupling noise, due to fine-tuning both the component matrices simultaneously with a uniform learning rate. In contrast, \algname{SPRA-LoRA} demonstrates superior performance by resampling and freezing $\{\mathbf{A}_i^{e,0}\}$ and utilizing rank-aware learning rates for clients. Remarkably, \algname{SPRA-LoRA} achieves these gains while reducing computational and communication overhead by 50\% compared to \algname{FLoRA} and other baselines. 

To avoid coupling noise, \algname{LA-LoRA} and \algname{RoLoRA} alternately optimize $\mathbf{A}$ and $\mathbf{B}$ between local steps and rounds, respectively, matching the computational overhead of \algname{SPRA-LoRA}. However, \algname{SPRA-LoRA} outperforms them by treating $\mathbf{A}$ as a resampled random projection matrix and allocating the full computational budget to $\mathbf{B}$. This is motivated by the fact that with zero-initialized $\mathbf{B}$, matrix $\mathbf{A}$ remains nearly frozen during fine-tuning (\Cref{sec:perturbation}, \Cref{lem:lora_final_full_forwardpass} in Appendix, and \cite{lrvsrICML2026}).

\begin{table}[t]
\centering
\setlength\tabcolsep{2pt}
\begin{minipage}[c]{1\textwidth}
\centering
\caption{Average test accuracy (5 runs) on GLUE language understanding benchmark with Dirichlet distribution ($\alpha=0.8$), \texttt{RoBERTa} model and \texttt{AdamW} optimizer. Clients' ranks follow \textbf{\emph{uniform}} distribution over $\{\texttt{4}, \texttt{8}, \dots, \texttt{128}\}$. \algname{SPRA-LoRA} and its doubled-rank variant outperform baselines while maintaining 50\% less or equal client overhead, respectively.}
\label{tab:Glue_iid_alpha_20_uniformrank}
\resizebox{\textwidth}{!}{
\begin{tabular}{c|c|ccccccc|c}
\toprule
\bf{Method}& Run time & \bf{QNLI} & \bf{MRPC} & \bf{CoLA} & \bf{MNLI} & \bf{RTE} & \bf{SST-2} & \bf{QQP}& \bf{Avg} 
\\ 
\toprule
\algname{Base Pretrained Model}& --- & 49.46 & 31.62 & 30.87 & 35.45 & 52.71 & 49.08 & 63.18 & 44.62 \\
\midrule
\algname{FFA} [\citeyear{FFA}] & 0.90h & 83.23$_{\pm 0.60}$ & 80.15$_{\pm 1.07}$ & 77.66$_{\pm 0.89}$ & 73.22$_{\pm 0.95}$ & 66.64$_{\pm 1.78}$ & 91.01$_{\pm 0.25}$ & 79.62$_{\pm 0.70}$ & 78.79$_{\pm 0.38}$ \\
\algname{HetLoRA} [\citeyear{HetLoRA}] & 0.62h & 81.33$_{\pm 1.51}$ & 85.29$_{\pm 0.17}$ & 79.94$_{\pm 0.29}$ & 77.25$_{\pm 0.11}$ & 64.40$_{\pm 0.48}$ & 92.61$_{\pm 0.19}$ & 81.25$_{\pm 0.64}$ & 80.30$_{\pm 0.25}$ \\
\algname{FlexLoRA} [\citeyear{Flexlora}] & 1.11h & \bf 87.71$_{\pm 0.20}$ & 84.61$_{\pm 1.61}$ & \bf 81.36$_{\pm 0.41}$ & 78.02$_{\pm 0.76}$ & 69.53$_{\pm 1.38}$ & 93.12$_{\pm 0.19}$ & 81.12$_{\pm 2.18}$ & 82.21$_{\pm 0.45}$ \\
\algname{FLoRA} [\citeyear{flora}] & 0.92h & 84.47$_{\pm 1.25}$ & 85.05$_{\pm 0.23}$ & 79.23$_{\pm 1.15}$ & 76.20$_{\pm 0.70}$ & 67.87$_{\pm 0.62}$ & 92.36$_{\pm 0.45}$ & 81.60$_{\pm 0.40}$ & 80.97$_{\pm 0.29}$ \\
\algname{FSLoRA} [\citeyear{Fslora}] & 0.84h & 76.02$_{\pm 2.29}$ & 83.43$_{\pm 0.64}$ & 79.31$_{\pm 0.63}$ & 60.40$_{\pm 2.88}$ & 65.06$_{\pm 1.87}$ & 92.00$_{\pm 0.13}$ & 79.13$_{\pm 0.96}$ & 76.48$_{\pm 0.62}$ \\
\algname{FedSA} [\citeyear{Fedsa}] & 1.28h & 69.65$_{\pm 2.36}$ & 68.53$_{\pm 1.52}$ & 60.86$_{\pm 1.87}$ & 46.46$_{\pm 1.18}$ & 51.02$_{\pm 0.45}$ & 87.28$_{\pm 1.02}$ & 65.29$_{\pm 0.59}$ & 64.16$_{\pm 0.54}$ \\
\algname{RoLoRA} [\citeyear{RoLoRA}] & 0.60h & 74.65$_{\pm 2.01}$ & 82.99$_{\pm 0.79}$ & 76.91$_{\pm 0.65}$ & 52.92$_{\pm 2.61}$ & 62.53$_{\pm 1.42}$ & 90.78$_{\pm 0.15}$ & 77.97$_{\pm 0.64}$ & 74.11$_{\pm 0.54}$ \\
\algname{LA-LoRA} [\citeyear{LALoRA}] & 0.61h & 74.96$_{\pm 2.54}$ & 83.97$_{\pm 0.46}$ & 76.24$_{\pm 0.84}$ & 67.57$_{\pm 1.78}$ & 64.26$_{\pm 0.16}$ & 91.17$_{\pm 0.06}$ & 78.64$_{\pm 0.60}$ & 76.69$_{\pm 0.47}$ \\
\midrule
\algname{SPRA-LoRA} (uniform LR) & 0.86h & 84.77$_{\pm 0.55}$ & 83.68$_{\pm 0.46}$ & 78.69$_{\pm 1.19}$ & 74.51$_{\pm 0.97}$ & 66.42$_{\pm 0.56}$ &92.55$_{\pm 0.21}$ & 81.27$_{\pm 0.46}$ & 80.27$_{\pm 0.26}$ \\
\rowcolor{lightgray}\algname{SPRA-LoRA} & 0.86h & $\underline{87.52}_{\pm 1.20}$ & \textbf{85.58}$_{\pm 0.57}$ & $\underline{80.88}_{\pm 0.50}$ & $\underline{80.70}_{\pm 0.43}$ & $\underline{70.54}_{\pm 1.27}$ & \textbf{93.60}$_{\pm 0.30}$ & \textbf{84.26}$_{\pm 0.32}$ & $\underline{83.30}_{\pm 0.29}$ \\
\rowcolor{lightgray}\algname{SPRA-LoRA} (doubled ranks) & 0.96h & 86.51$_{\pm 2.03}$ & $\underline{85.39}_{\pm 0.31}$ & 80.63$_{\pm 0.88}$ & \textbf{81.80}$_{\pm 0.21}$ & \textbf{71.84}$_{\pm 0.82}$ & $\underline{93.14}_{\pm 0.30}$ & $\underline{84.20}_{\pm 0.37}$ & \textbf{83.36}$_{\pm 0.35}$ \\
\bottomrule
\end{tabular}}
\end{minipage}\hfill
\vspace{-2em}
\end{table}

\algname{FedSA} performs the weakest. By training both $\mathbf{A}$ and $\mathbf{B}$ locally while aggregating only $\mathbf{A}$ on the server, it leaves $\mathbf{B}$ local. Hence, clients never receive an aggregated $\mathbf{B}$ matrix. This necessitates strict full client participation \cite{Fedsa}. Furthermore, because each client maintains a personalized model, \algname{FedSA}'s results are averaged across random seeds and clients. As shown, under partial participation, the overall performance for \algname{FedSA} is low and there is a high variance across clients (large \texttt{std}s).

Finally, to ensure a fair comparison, the final row of results evaluates \algname{SPRA-LoRA} with doubled client ranks to match the trainable parameters and overhead of the baselines. Under these equivalent resource constraints, \algname{SPRA-LoRA} achieves its most significant performance gains, highlighting its superior efficiency and utility. Similar trends are observed for the GLUE benchmark in \Cref{tab:Glue_iid_alpha_20_uniformrank}.

\begin{figure}[t]
     \centering
     \begin{subfigure}[b]{0.48\textwidth}
         \centering
         \includegraphics[width=\textwidth]{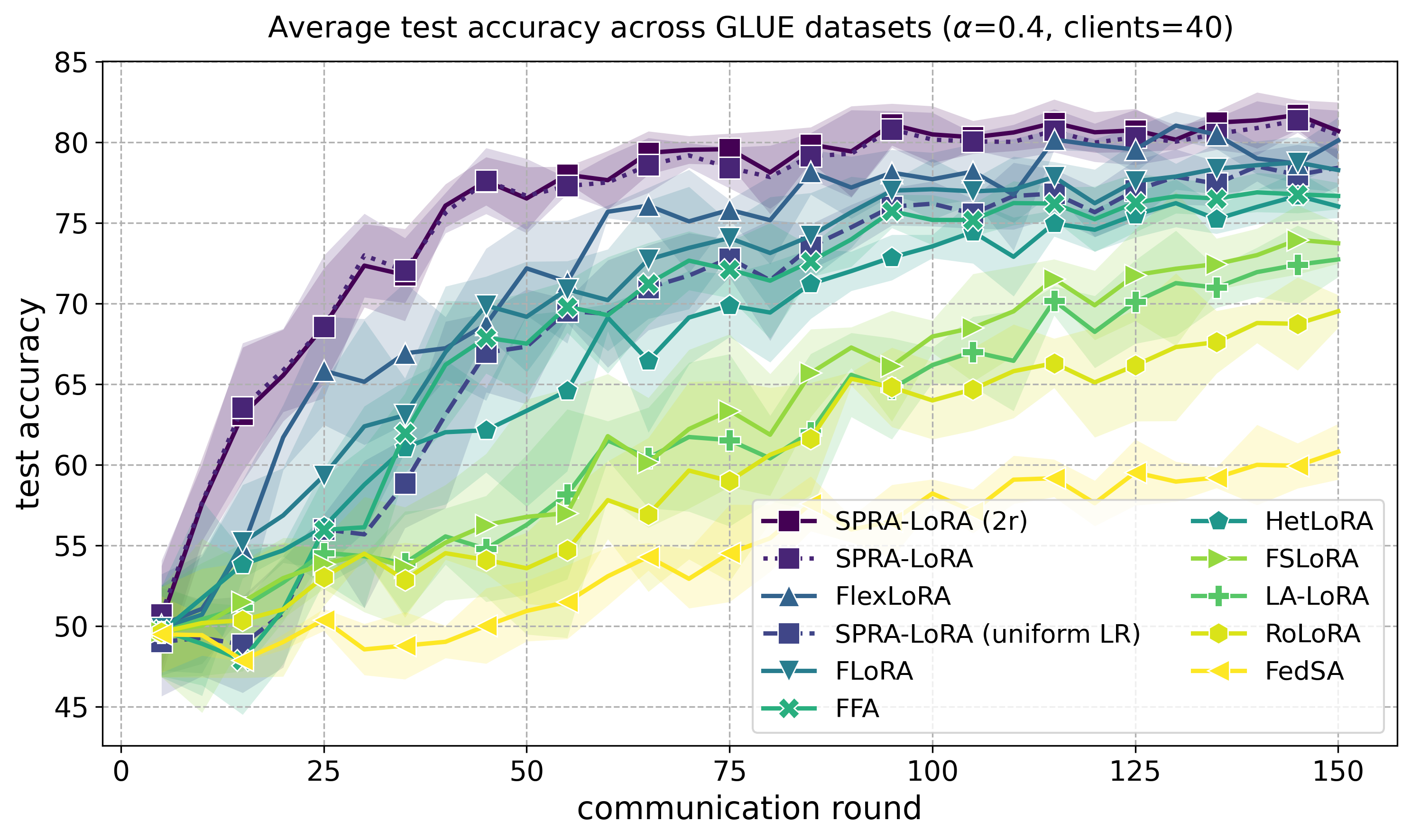}
     \end{subfigure}
     \hfill
     \begin{subfigure}[b]{0.48\textwidth}
         \centering
         \includegraphics[width=\textwidth]{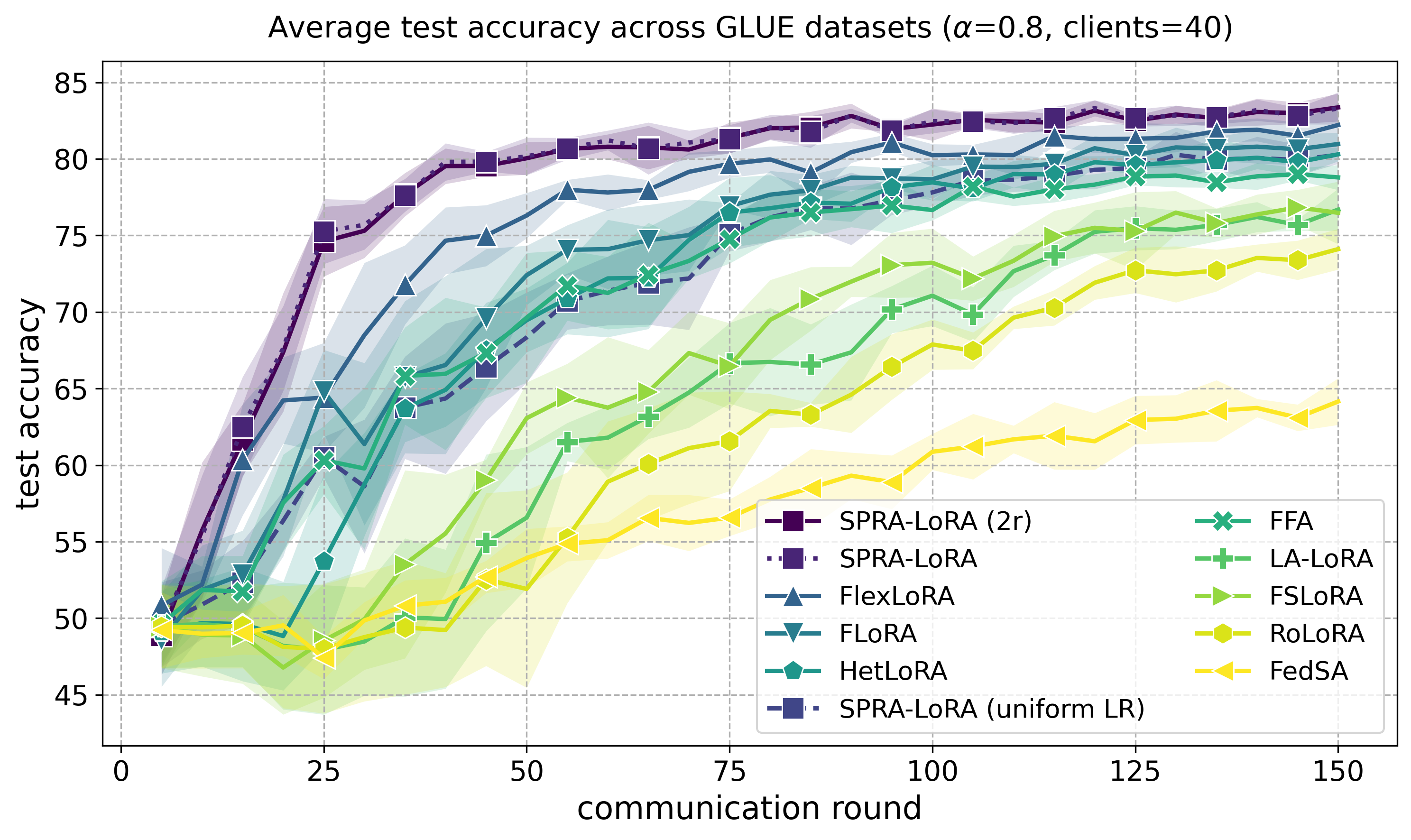}
     \end{subfigure}
     \caption{\textbf{Left:} Average test accuracy across 5 runs on GLUE with Dirichlet distribution ($\alpha=0.4$), \texttt{RoBERTa} model and \texttt{AdamW} optimizer. \textbf{Right:} Same as left, with $\alpha=0.8$.}
     \label{fig:glueconvergence}
     \vspace{-0.5em}
\end{figure}

\begin{figure}[t]
     \centering
     \begin{subfigure}[b]{0.48\textwidth}
         \centering
         \includegraphics[width=\textwidth]{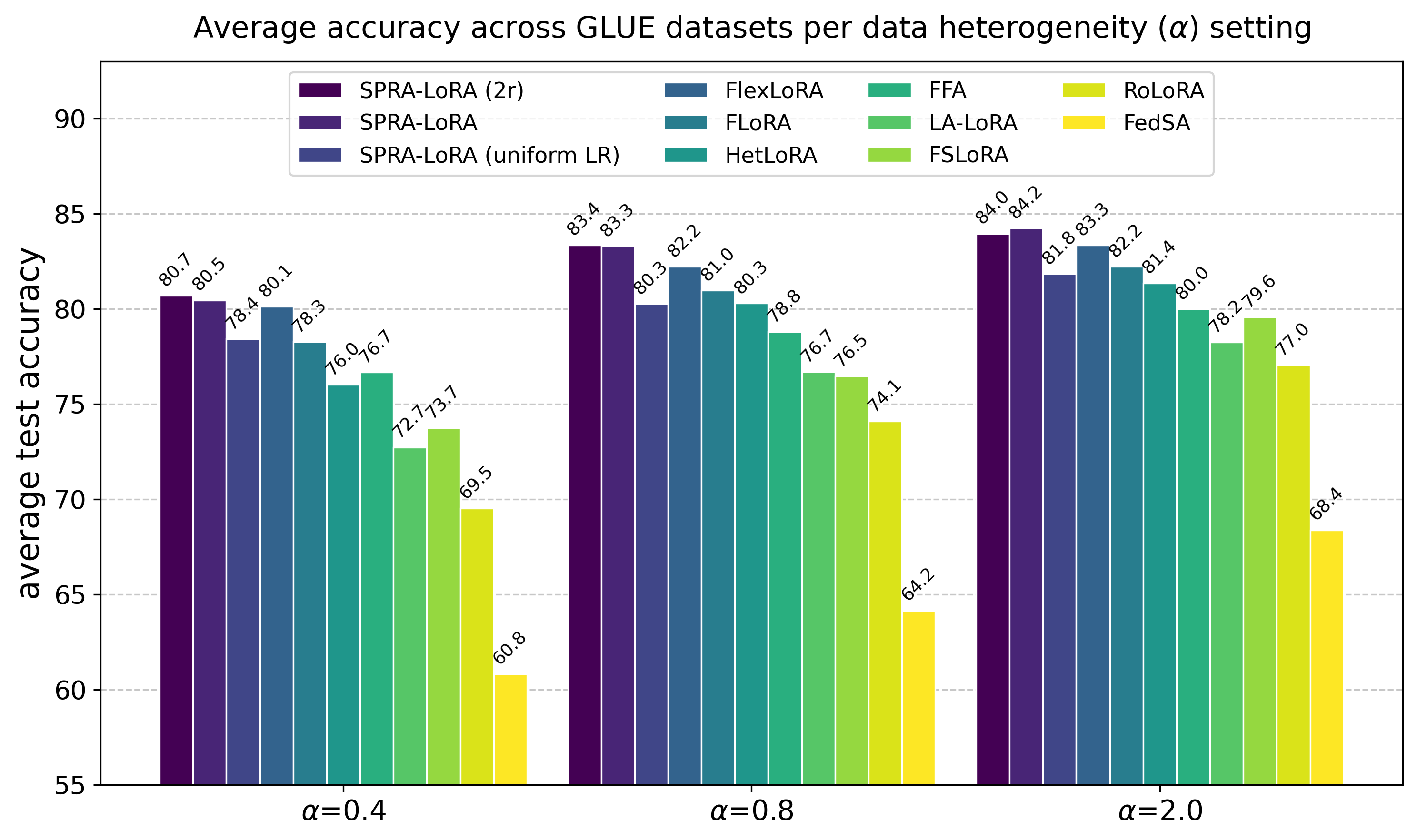}
     \end{subfigure}
     \hfill
     \begin{subfigure}[b]{0.48\textwidth}
         \centering
         \includegraphics[width=\textwidth]{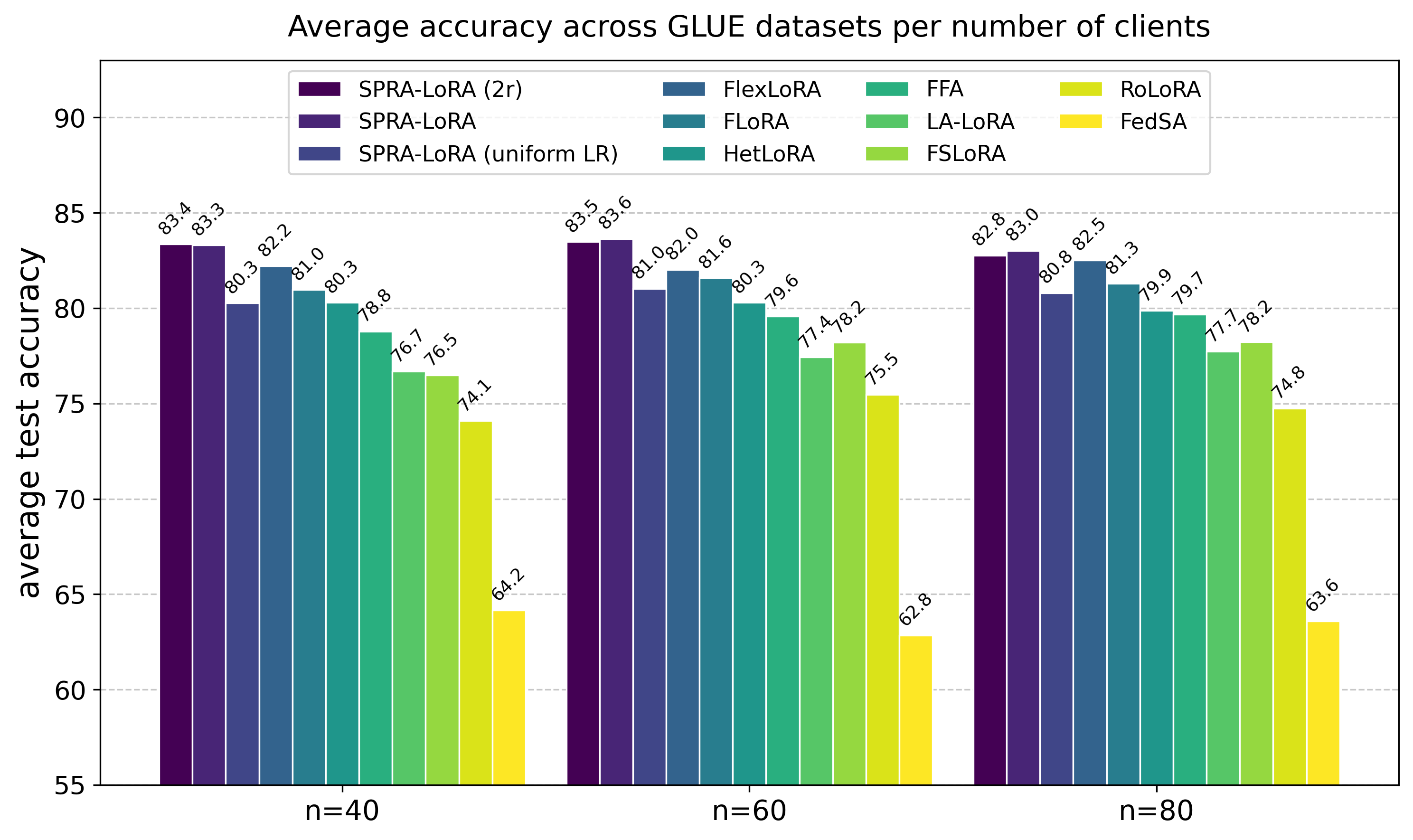}
     \end{subfigure}
     \caption{\textbf{Left:} Average test accuracy across 5 runs on GLUE language understanding benchmark with $n=40$ and Dirichlet distribution with $\alpha\in\{\texttt{0.4},\texttt{0.8},\texttt{2.0}\}$, \texttt{RoBERTa} model and \texttt{AdamW} optimizer. \textbf{Right:} Average test accuracy across 5 runs on GLUE and Dirichlet distribution with $\alpha=0.8$, number of clients $n\in\{\texttt{40}, \texttt{60}, \texttt{80}\}$, \texttt{RoBERTa} model and \texttt{AdamW} optimizer.}
     \label{fig:ablation}
     \vspace{-1em}
\end{figure}

\begin{table*}[h!]
\centering
\setlength\tabcolsep{2pt}
\begin{minipage}[c]{1\textwidth}
\centering
\caption{Average test accuracy (10 runs) on Commonsense Reasoning dataset (\texttt{LLaMA-3.2-3B} model and \texttt{AdamW} optimizer) in a \emph{centralized} setting with fixed adaptation rank $r=64$. Total number of gradient steps is $1000$, and the frozen matrix $\mathbf{A}$ is resampled every $T$ steps.}
\label{tab:ablation_T}
\resizebox{\textwidth}{!}{
\begin{tabular}{c|cccccccc|c}
\toprule
\bf{Period}& \bf{BoolQ} & \bf{WinoGrande} & \bf{OBQA} & \bf{ARC-e} & \bf{ARC-c} & \bf{SIQA} & \bf{PIQA}& \bf{HellaSwag} & \bf{Avg} 
\\ 
\toprule
$T=20$& 
$66.11_{\pm 0.95}$ & 
$67.45_{\pm 0.72}$ & 
$73.12_{\pm 1.01}$ & 
$85.24_{\pm 0.48}$ & 
$70.26_{\pm 1.06}$ & 
$71.52_{\pm 0.71}$ & 
$79.54_{\pm 0.68}$ & 
$70.34_{\pm 1.35}$ & 
$72.94_{\pm 0.43}$ \\
$T=10$& 
$64.58_{\pm 1.06}$ & 
$69.99_{\pm 1.19}$ & 
$73.54_{\pm 1.28}$ & 
$84.27_{\pm 0.55}$ & 
$68.15_{\pm 1.14}$ & 
$71.97_{\pm 0.72}$ & 
$79.35_{\pm 1.06}$ &
$71.31_{\pm 1.14}$ & 
$72.89_{\pm 0.23}$ \\
$T=5$& 
$65.65_{\pm 1.81}$ & 
$68.08_{\pm 1.05}$ & 
$72.81_{\pm 1.01}$ & 
$85.61_{\pm 0.51}$ & 
$69.99_{\pm 0.92}$ & 
$72.06_{\pm 0.53}$ & 
$79.24_{\pm 1.11}$ & 
$69.84_{\pm 1.77}$ & 
$72.91_{\pm 0.43}$ \\
$T=1$& 
$64.72_{\pm 2.23}$ & 
$69.59_{\pm 2.41}$ & 
$73.46_{\pm 0.41}$ & 
$84.59_{\pm 0.64}$ & 
$69.86_{\pm 1.31}$ & 
$73.33_{\pm 0.68}$ & 
$79.27_{\pm 0.47}$ & 
$72.33_{\pm 1.62}$ & 
$73.39_{\pm 0.31}$ \\
\bottomrule
\end{tabular}}
\end{minipage}\hfill
\vspace{-1em}
\end{table*}

 \algname{SPRA-LoRA}’s gains derive from per-round resampling of $\mathbf{A}$ matrices, focusing computational resources on optimizing $\mathbf{B}$ matrices and rank-aware heterogeneous learning rates. This approach avoids both aggregation and coupling noises. Convergence results in \Cref{fig:glueconvergence} demonstrate that both variants of \algname{SPRA-LoRA} consistently achieve higher average test accuracy across the GLUE benchmark throughout the communication rounds compared to the baselines. 

 \Cref{fig:ablation} shows an ablation study on data heterogeneity ($\alpha$) and number of clients ($n$), with consistently superior performance of \algname{SPRA-LoRA} variants. See Appendix \ref{app:ablations_details} for detailed results.

Finally, \Cref{tab:ablation_T} examines $T$ (local steps for \algname{SPRA-LoRA} and matrix $\mathbf{A}$ resampling period). Using a centralized setting ($r=64$, 1000 steps with \texttt{AdamW}) to isolate federated optimization effects, we find minimal difference between $T>1$ and $T=1$ for small $T$, showing negligible impact from the weak dependence between matrices $\mathbf{A}$ and batch gradients. Appendix \ref{app:meanshift} proves that per-$T$-steps resampling introduces a bounded mean shift to per-step sampling (\Cref{lem:noise_var}), decaying with $T$.

\section{Conclusion}
We proposed \algname{SPRA-LoRA} to address update scale imbalances across heterogeneous client ranks. By freezing and resampling matrices $\mathbf{A}$ and applying rank-scaled learning rates, \algname{SPRA-LoRA} balances gradient perturbations for superior efficient performance. Future work will explore dynamic rank adaptation and continuous learning rate schedules.

\newpage
\section*{Impact Statement}
The goal of this paper is to advance the field of Machine Learning. Our paper is foundational research and not tied to any particular application with negative impacts. There are many potential societal consequences of our work, none which we feel must be specifically highlighted here.

\section*{Acknowledgement and Disclosure of Funding}
Funding support for project activities has been partially provided by NSERC Canada, Canada CIFAR AI Chair, Facebook award, and Google award.

\bibliography{neurips}

\newpage
\appendix

\begin{center}
\Large
\bf
Appendix: \emph{``Stochastic Projections and Rank-Aware Learning Rates for Federated LoRA"}
\end{center}

\section{Further experimental details}\label{app:exp_setup}
We provide more details about our experiments in this section. 

\subsection{Experimental setup}\label{app:expsetup}
In all experiments, we sample client ranks from $\{\texttt{4},\texttt{8},\texttt{16},\texttt{32},\texttt{64},\texttt{128}\}$. For each training experiment, we run 5 seeds and report the mean and sample standard deviation (\texttt{std}) across them. We use $p=0.2$ participation rate, uniform batch size of $b_i=16$, and \texttt{AdamW} optimizer. Next, we describe specific configuration choices for the GLUE and Commonsense reasoning benchmarks.

\textbf{GLUE.} The GLUE benchmark is a collection of natural language understanding tasks, formatted as classification tasks with 2 or 3 classes~\citep{gluebenchmark}. We randomly initialize and freeze the classification layer of a \texttt{RoBERTa-base} model. We then apply rank-stabilized \algname{LoRA} (rs-\algname{LoRA}) with $\alpha=1$ to the \texttt{["query", "value", "key", "dense"]} parameters, where the first three refer to the attention input projections and ``dense'' is the second-to-last fully connected layer (before the final classifier layer). We use a maximum sequence length of 128 tokens. We tune learning rate over $\{\texttt{1e-4},\texttt{2e-4},\texttt{4e-4},\texttt{1e-3},\texttt{2e-3},\texttt{4e-3}\}$ and select the best performing learning rate based on the average across all 7 benchmarks. We use the same grid of learning rates when searching for the client-specific ranks for \algname{SPRA-LoRA}.

We do an ablation study on the GLUE benchmark with varying data heterogeneity (Dirichlet parameter $\alpha$) and number of clients ($n$), as shown in Table~\ref{tab:glue_settings}. For all experiments with GLUE, we use a fixed number of communication rounds $E=150$, with $T=20$ gradient steps during each round.

\begin{table}[h!]
    \centering
     \caption{Studied values of client population size and data heterogeneity $(n, \alpha)$ on GLUE.}
    \label{tab:glue_settings}
    \begin{tabular}{r|c|c|c|c|c}
    \toprule
        $n$ (clients)            & \multicolumn{3}{c|}{40} & 60 & 80 \\
        \midrule
        $\alpha$ (heterogeneity) & \texttt{0.4} & \texttt{0.8} & \texttt{2.0} & \multicolumn{2}{c}{\texttt{0.8}}  \\
        \bottomrule
    \end{tabular}
\end{table}

\textbf{Commonsense Reasoning.} The training set of the commonsense reasoning benchmark is a mixture of 8 datasets including about 170K training samples \citep{commonsensereasoning}. For fine-tuning, we use rs-\algname{LoRA} with $\alpha=1$. We apply \algname{LoRA} to the \texttt{["q proj", "k proj", "v proj", "up proj", "down proj"]} parameters of a pretrained \texttt{LLaMA-3.2-3B} model. We use a maximum sequence length of 128 tokens. We tune learning rate over $\{\texttt{1e-4},\texttt{2e-4},\texttt{4e-4},\texttt{6e-4},\texttt{8e-4},\texttt{1e-3},\texttt{2e-3},\texttt{4e-3}, \texttt{6e-3}, \texttt{8e-3}\}$ and select the best performing learning rate based on the average across all the 8 datasets. We split data between $n=20$ clients with Dirichlet distribution with $\alpha=2$ and use a fixed number of communication rounds $E=100$, with $T=20$ gradient steps during each round.
\newpage
\subsection{Further detailed results}\label{app:ablations_details}

\begin{table*}[h!]
\centering
\setlength\tabcolsep{2pt}
\begin{minipage}[c]{1\textwidth}
\centering
\caption{Test accuracy averaged over 5 runs on GLUE benchmark with Dirichlet distribution ($\mathbf{\alpha=0.4}$), \texttt{RoBERTa} model, \texttt{AdamW} optimizer and $\mathbf{n=40}$ clients. Clients' ranks follow \textbf{\emph{uniform}} distribution over $\{\texttt{4}, \texttt{8}, \dots, \texttt{128}\}$.}
\label{tab:iid_alpha_04_n40}
\resizebox{\textwidth}{!}{
\begin{tabular}{c|c|ccccccc|c}
\toprule
\bf{Method}& Run time & \bf{QNLI} & \bf{MRPC} & \bf{CoLA} & \bf{MNLI} & \bf{RTE} & \bf{SST-2} & \bf{QQP}& \bf{Avg} 
\\ 
\toprule
\algname{Base Pretrained Model}& --- & 49.46 & 31.62 & 30.87 & 35.45 & 52.71 & 49.08 & 63.18 & 44.62 \\
\midrule
\algname{FFA} [\citeyear{FFA}] & 0.82h & 81.28$_{\pm 0.70}$ & 77.80$_{\pm 1.70}$ & 77.66$_{\pm 0.33}$ & 70.19$_{\pm 1.15}$ & 66.14$_{\pm 0.82}$ & 90.48$_{\pm 0.19}$ & 73.14$_{\pm 3.35}$ & 76.67$_{\pm 0.58}$ \\
\algname{HetLoRA} [\citeyear{HetLoRA}] & 0.61h & 83.61$_{\pm 0.78}$ & 81.42$_{\pm 1.82}$ & 78.01$_{\pm 0.95}$ & 65.98$_{\pm 1.89}$ & 53.93$_{\pm 2.70}$ & 91.42$_{\pm 0.29}$ & \underline{77.78}$_{\pm 2.37}$ & 76.02$_{\pm 0.66}$ \\
\algname{FlexLoRA} [\citeyear{Flexlora}] & 1.05h & 83.68$_{\pm 3.44}$ & 83.67$_{\pm 0.59}$ & \underline{80.81}$_{\pm 0.44}$ & 77.05$_{\pm 0.82}$ & 65.34$_{\pm 3.48}$ & 90.83$_{\pm 0.84}$ & \textbf{79.42}$_{\pm 1.49}$ & 80.11$_{\pm 0.76}$ \\
\algname{FLoRA} [\citeyear{flora}] & 0.84h & 82.15$_{\pm 2.16}$ & 80.78$_{\pm 1.97}$ & 79.64$_{\pm 0.93}$ & 73.24$_{\pm 1.20}$ & 66.50$_{\pm 1.66}$ & 91.77$_{\pm 0.35}$ & 73.85$_{\pm 3.96}$ & 78.28$_{\pm 0.77}$ \\
\algname{FSLoRA} [\citeyear{Fslora}] & 0.78h & 79.48$_{\pm 1.47}$ & 78.72$_{\pm 1.32}$ & 74.27$_{\pm 0.89}$ & 58.83$_{\pm 2.35}$ & 56.17$_{\pm 3.95}$ & 91.88$_{\pm 0.47}$ & 76.80$_{\pm 1.02}$ & 73.74$_{\pm 0.74}$ \\
\algname{FedSA} [\citeyear{Fedsa}] & 1.27h & 64.43$_{\pm 1.17}$ & 62.03$_{\pm 4.38}$ & 65.54$_{\pm 1.19}$ & 43.35$_{\pm 0.99}$ & 51.28$_{\pm 0.85}$ & 73.45$_{\pm 3.53}$ & 65.60$_{\pm 1.61}$ & 60.81$_{\pm 0.89}$ \\
\algname{RoLoRA} [\citeyear{RoLoRA}] & 0.58h & 79.49$_{\pm 0.90}$ & 75.49$_{\pm 1.39}$ & 74.50$_{\pm 1.59}$ & 39.96$_{\pm 2.27}$ & 53.72$_{\pm 3.18}$ & 90.00$_{\pm 0.33}$ & 73.46$_{\pm 2.48}$ & 69.52$_{\pm 0.74}$ \\
\algname{LA-LoRA} [\citeyear{LALoRA}] & 0.60h & 80.52$_{\pm 0.97}$ & 77.16$_{\pm 2.27}$ & 76.55$_{\pm 1.14}$ & 56.71$_{\pm 2.45}$ & 54.94$_{\pm 3.29}$ & 90.50$_{\pm 0.44}$ & 72.74$_{\pm 3.50}$ & 72.73$_{\pm 0.86}$ \\
\midrule
\algname{SPRA-LoRA} (uniform LR) & 0.85h & 83.30$_{\pm 1.39}$ & 80.39$_{\pm 1.66}$ & 79.10$_{\pm 0.45}$ & 71.05$_{\pm 1.95}$ & \textbf{68.23}$_{\pm 1.36}$ & 91.42$_{\pm 0.23}$ & 75.48$_{\pm 2.77}$ & 78.42$_{\pm 0.61}$ \\
\rowcolor{lightgray}\algname{SPRA-LoRA} & 0.79h & \textbf{85.59}$_{\pm 2.09}$ & \textbf{84.66}$_{\pm 0.75}$ & \underline{80.81}$_{\pm 0.89}$ & \underline{78.28}$_{\pm 0.86}$ & 64.47$_{\pm 2.36}$ & \textbf{92.32}$_{\pm 0.58}$ & 77.10$_{\pm 3.86}$ & \underline{80.46}$_{\pm 0.75}$ \\
\rowcolor{lightgray}\algname{SPRA-LoRA} (doubled ranks) & 0.83h & \underline{85.53}$_{\pm 2.36}$ & \underline{83.82}$_{\pm 2.18}$ & \textbf{81.07}$_{\pm 0.79}$ & \textbf{78.62}$_{\pm 0.86}$ & \underline{67.00}$_{\pm 1.66}$ & \underline{92.29}$_{\pm 0.38}$ & 76.54$_{\pm 4.22}$ & \textbf{80.70}$_{\pm 0.81}$ \\
\bottomrule
\end{tabular}}
\end{minipage}\hfill
\label{table:iid_alpha_04_n40}
\end{table*}

\begin{table*}[h!]
\centering
\setlength\tabcolsep{2pt}
\begin{minipage}[c]{1\textwidth}
\centering
\caption{Test accuracy averaged over 5 runs on GLUE benchmark with Dirichlet distribution ($\mathbf{\alpha=2.0}$), \texttt{RoBERTa} model, \texttt{AdamW} optimizer, and $\mathbf{n=40}$ clients. Clients' ranks follow \textbf{\emph{uniform}} distribution over $\{\texttt{4}, \texttt{8}, \dots, \texttt{128}\}$.}
\label{tab:iid_alpha_20_n40}
\resizebox{\textwidth}{!}{
\begin{tabular}{c|c|ccccccc|c}
\toprule
\bf{Method}& Run time & \bf{QNLI} & \bf{MRPC} & \bf{CoLA} & \bf{MNLI} & \bf{RTE} & \bf{SST-2} & \bf{QQP}& \bf{Avg} 
\\ 
\toprule
\algname{Base Pretrained Model}& --- & 49.46 & 31.62 & 30.87 & 35.45 & 52.71 & 49.08 & 63.18 & 44.62 \\
\midrule
\algname{FFA} [\citeyear{FFA}] & 0.76h & 84.94$_{\pm 0.19}$ & 84.51$_{\pm 0.80}$ & 79.77$_{\pm 0.77}$ & 74.95$_{\pm 0.37}$ & 67.51$_{\pm 0.52}$ & 91.22$_{\pm 0.11}$ & 77.11$_{\pm 0.93}$ & 80.00$_{\pm 0.23}$ \\
\algname{HetLoRA} [\citeyear{HetLoRA}] & 0.59h & 86.18$_{\pm 0.48}$ & 83.53$_{\pm 1.44}$ & 81.72$_{\pm 0.29}$ & 78.44$_{\pm 0.12}$ & 70.18$_{\pm 0.54}$ & 92.39$_{\pm 0.15}$ & 77.09$_{\pm 0.46}$ & 81.36$_{\pm 0.24}$ \\
\algname{FlexLoRA} [\citeyear{Flexlora}] & 1.02h & 87.31$_{\pm 0.34}$ & \textbf{86.42}$_{\pm 0.33}$ & 82.26$_{\pm 0.21}$ & 80.20$_{\pm 0.09}$ & \underline{73.86}$_{\pm 0.80}$ & 92.98$_{\pm 0.11}$ & 80.35$_{\pm 0.70}$ & 83.34$_{\pm 0.17}$ \\
\algname{FLoRA} [\citeyear{flora}] & 0.83h & 86.20$_{\pm 0.60}$ & 84.95$_{\pm 0.80}$ & 81.84$_{\pm 0.30}$ & 78.50$_{\pm 0.23}$ & 71.48$_{\pm 0.59}$ & 92.75$_{\pm 0.23}$ & 79.78$_{\pm 0.76}$ & 82.21$_{\pm 0.21}$ \\
\algname{FSLoRA} [\citeyear{Fslora}] & 0.78h & 85.19$_{\pm 0.04}$ & 83.28$_{\pm 1.56}$ & 81.05$_{\pm 0.25}$ & 74.15$_{\pm 0.61}$ & 65.34$_{\pm 1.67}$ & 92.00$_{\pm 0.17}$ & 76.01$_{\pm 0.75}$ & 79.57$_{\pm 0.36}$ \\
\algname{FedSA} [\citeyear{Fedsa}] & 1.28h & 72.01$_{\pm 2.13}$ & 67.55$_{\pm 1.60}$ & 72.39$_{\pm 0.99}$ & 52.24$_{\pm 1.90}$ & 52.37$_{\pm 0.41}$ & 90.46$_{\pm 0.24}$ & 71.63$_{\pm 1.21}$ & 68.38$_{\pm 0.52}$ \\
\algname{RoLoRA} [\citeyear{RoLoRA}] & 0.62h & 82.42$_{\pm 0.54}$ & 80.98$_{\pm 2.29}$ & 80.86$_{\pm 0.37}$ & 64.38$_{\pm 2.24}$ & 66.64$_{\pm 0.57}$ & 91.21$_{\pm 0.12}$ & 72.80$_{\pm 1.25}$ & 77.04$_{\pm 0.51}$ \\
\algname{LA-LoRA} [\citeyear{LALoRA}] & 0.60h & 82.25$_{\pm 0.70}$ & 80.78$_{\pm 2.62}$ & 80.54$_{\pm 0.30}$ & 71.18$_{\pm 0.56}$ & 68.30$_{\pm 0.97}$ & 91.72$_{\pm 0.10}$ & 72.89$_{\pm 1.02}$ & 78.24$_{\pm 0.45}$ \\
\midrule
\algname{SPRA-LoRA} (uniform LR) & 0.77h & 86.12$_{\pm 0.53}$ & 84.66$_{\pm 0.58}$ & 81.65$_{\pm 0.31}$ & 77.75$_{\pm 0.19}$ & 70.25$_{\pm 0.90}$ & 92.73$_{\pm 0.19}$ & 79.78$_{\pm 0.90}$ & 81.85$_{\pm 0.22}$ \\
\rowcolor{lightgray}\algname{SPRA-LoRA} & 0.81h & \underline{88.57}$_{\pm 0.60}$ & \underline{86.03}$_{\pm 0.64}$ & \underline{82.38}$_{\pm 0.35}$ & \textbf{82.73}$_{\pm 0.12}$ & \textbf{73.87}$_{\pm 0.18}$ & \underline{93.37}$_{\pm 0.09}$ & \textbf{82.77}$_{\pm 0.49}$ & \textbf{84.25}$_{\pm 0.16}$ \\
\rowcolor{lightgray}\algname{SPRA-LoRA} (doubled ranks) & 0.95h & \textbf{89.45}$_{\pm 0.48}$ & 85.93$_{\pm 1.15}$ & \textbf{82.72}$_{\pm 0.19}$ & \underline{82.71}$_{\pm 0.11}$ & 71.91$_{\pm 1.02}$ & \textbf{93.56}$_{\pm 0.10}$ & \underline{81.43}$_{\pm 0.93}$ & \underline{83.96}$_{\pm 0.27}$ \\
\bottomrule
\end{tabular}}
\end{minipage}\hfill
\label{table:iid_alpha_20_n40}
\end{table*}

\begin{table*}[h!]
\centering
\setlength\tabcolsep{2pt}
\begin{minipage}[c]{1\textwidth}
\centering
\caption{Test accuracy averaged over 5 runs on GLUE benchmark with Dirichlet distribution ($\mathbf{\alpha=0.8}$), \texttt{RoBERTa} model, \texttt{AdamW} optimizer and $\mathbf{n=60}$ clients. Clients' ranks follow \textbf{\emph{uniform}} distribution over $\{\texttt{4}, \texttt{8}, \dots, \texttt{128}\}$.}
\label{tab:iid_alpha_08_n60}
\resizebox{\textwidth}{!}{
\begin{tabular}{c|c|ccccccc|c}
\toprule
\bf{Method}& Run time & \bf{QNLI} & \bf{MRPC} & \bf{CoLA} & \bf{MNLI} & \bf{RTE} & \bf{SST-2} & \bf{QQP}& \bf{Avg} 
\\ 
\toprule
\algname{Base Pretrained Model}& --- & 49.46 & 31.62 & 30.87 & 35.45 & 52.71 & 49.08 & 63.18 & 44.62 \\
\midrule
\algname{FFA} [\citeyear{FFA}] & 1.26h & 85.15$_{\pm 0.43}$ & 83.48$_{\pm 0.28}$ & 79.06$_{\pm 0.37}$ & 74.04$_{\pm 0.58}$ & 63.83$_{\pm 1.54}$ & 91.49$_{\pm 0.28}$ & 79.95$_{\pm 0.79}$ & 79.57$_{\pm 0.28}$ \\
\algname{HetLoRA} [\citeyear{HetLoRA}] & 0.85h & 84.75$_{\pm 0.92}$ & 85.05$_{\pm 0.28}$ & 81.15$_{\pm 0.31}$ & 77.41$_{\pm 0.84}$ & 61.45$_{\pm 0.91}$ & 92.09$_{\pm 0.25}$ & 80.21$_{\pm 0.94}$ & 80.30$_{\pm 0.27}$ \\
\algname{FlexLoRA} [\citeyear{Flexlora}] & 1.45h & 86.22$_{\pm 0.85}$ & 84.75$_{\pm 0.46}$ & 80.79$_{\pm 0.33}$ & 78.30$_{\pm 0.69}$ & \underline{69.03}$_{\pm 0.79}$ & 92.06$_{\pm 0.38}$ & 83.02$_{\pm 0.20}$ & 82.02$_{\pm 0.22}$ \\
\algname{FLoRA} [\citeyear{flora}] & 1.31h & 85.48$_{\pm 0.85}$ & \underline{86.13}$_{\pm 0.28}$ & 79.79$_{\pm 0.61}$ & 77.06$_{\pm 0.32}$ & 68.45$_{\pm 0.67}$ & 93.09$_{\pm 0.07}$ & 81.17$_{\pm 0.87}$ & 81.60$_{\pm 0.23}$ \\
\algname{FSLoRA} [\citeyear{Fslora}] & 1.23h & 81.31$_{\pm 2.57}$ & 84.31$_{\pm 0.77}$ & 80.04$_{\pm 0.25}$ & 67.99$_{\pm 2.77}$ & 62.60$_{\pm 2.12}$ & 91.56$_{\pm 0.24}$ & 79.56$_{\pm 0.75}$ & 78.20$_{\pm 0.64}$ \\
\algname{FedSA} [\citeyear{Fedsa}] & 1.86h & 69.12$_{\pm 1.35}$ & 59.31$_{\pm 0.86}$ & 61.77$_{\pm 2.35}$ & 46.83$_{\pm 1.89}$ & 51.00$_{\pm 0.52}$ & 84.87$_{\pm 2.20}$ & 66.97$_{\pm 1.52}$ & 62.84$_{\pm 0.62}$ \\
\algname{RoLoRA} [\citeyear{RoLoRA}] & 0.89h & 80.48$_{\pm 1.31}$ & 82.35$_{\pm 0.52}$ & 79.44$_{\pm 0.49}$ & 59.10$_{\pm 4.99}$ & 60.00$_{\pm 1.56}$ & 91.12$_{\pm 0.25}$ & 75.85$_{\pm 1.60}$ & 75.48$_{\pm 0.81}$ \\
\algname{LA-LoRA} [\citeyear{LALoRA}] & 0.87h & 80.85$_{\pm 1.66}$ & 83.92$_{\pm 0.84}$ & 79.06$_{\pm 0.45}$ & 69.49$_{\pm 3.42}$ & 60.65$_{\pm 1.42}$ & 91.77$_{\pm 0.22}$ & 76.24$_{\pm 1.87}$ & 77.42$_{\pm 0.65}$ \\
\midrule
\algname{SPRA-LoRA} (uniform LR) & 1.18h & 85.82$_{\pm 0.66}$ & 85.44$_{\pm 0.72}$ & 79.52$_{\pm 0.75}$ & 75.46$_{\pm 0.40}$ & 66.86$_{\pm 1.93}$ & 92.91$_{\pm 0.08}$ & 81.21$_{\pm 0.62}$ & 81.03$_{\pm 0.34}$ \\
\rowcolor{lightgray}\algname{SPRA-LoRA} & 1.17h & \textbf{89.32}$_{\pm 0.58}$ & 85.98$_{\pm 0.38}$ & \textbf{81.32}$_{\pm 0.32}$ & \underline{81.31}$_{\pm 0.28}$ & \textbf{69.89}$_{\pm 0.72}$ & \textbf{93.33}$_{\pm 0.13}$ & \textbf{84.32}$_{\pm 0.67}$ & \textbf{83.64}$_{\pm 0.18}$ \\
\rowcolor{lightgray}\algname{SPRA-LoRA} (doubled ranks) & 1.32h & \underline{89.02}$_{\pm 0.57}$ & \textbf{86.91}$_{\pm 0.21}$ & \underline{81.23}$_{\pm 0.22}$ & \textbf{81.66}$_{\pm 0.28}$ & 68.09$_{\pm 1.47}$ & \underline{93.16}$_{\pm 0.09}$ & \underline{84.30}$_{\pm 0.85}$ & \underline{83.48}$_{\pm 0.26}$ \\
\bottomrule
\end{tabular}}
\end{minipage}\hfill
\label{table:iid_alpha_08_n60}
\end{table*}

\begin{table*}[h!]
\centering
\setlength\tabcolsep{2pt}
\begin{minipage}[c]{1\textwidth}
\centering
\caption{Test accuracy averaged over 5 runs on GLUE benchmark with Dirichlet distribution ($\mathbf{\alpha=0.8}$), \texttt{RoBERTa} model, \texttt{AdamW} optimizer and $\mathbf{n=80}$ clients. Clients' ranks follow \textbf{\emph{uniform}} distribution over $\{\texttt{4}, \texttt{8}, \dots, \texttt{128}\}$.}
\label{tab:iid_alpha_08_n80}
\resizebox{\textwidth}{!}{
\begin{tabular}{c|c|ccccccc|c}
\toprule
\bf{Method}& Run time & \bf{QNLI} & \bf{MRPC} & \bf{CoLA} & \bf{MNLI} & \bf{RTE} & \bf{SST-2} & \bf{QQP}& \bf{Avg} 
\\ 
\toprule
\algname{Base Pretrained Model}& --- & 49.46 & 31.62 & 30.87 & 35.45 & 52.71 & 49.08 & 63.18 & 44.62 \\
\midrule
\algname{FFA} [\citeyear{FFA}] & 1.51h & 84.68$_{\pm 0.55}$ & 82.45$_{\pm 0.85}$ & 79.46$_{\pm 0.50}$ & 74.76$_{\pm 0.80}$ & 66.57$_{\pm 1.59}$ & 91.72$_{\pm 0.32}$ & 78.12$_{\pm 1.44}$ & 79.68$_{\pm 0.37}$ \\
\algname{HetLoRA} [\citeyear{HetLoRA}] & 1.08h & 86.10$_{\pm 0.33}$ & 84.07$_{\pm 0.58}$ & \underline{80.71}$_{\pm 0.35}$ & 77.74$_{\pm 0.48}$ & 58.77$_{\pm 2.02}$ & 92.27$_{\pm 0.11}$ & 79.46$_{\pm 1.06}$ & 79.87$_{\pm 0.35}$ \\
\algname{FlexLoRA} [\citeyear{Flexlora}] & 1.54h & 87.36$_{\pm 0.20}$ & 85.73$_{\pm 0.40}$ & 80.59$_{\pm 1.01}$ & 78.88$_{\pm 0.23}$ & \textbf{70.47}$_{\pm 1.99}$ & 92.98$_{\pm 0.25}$ & 81.66$_{\pm 1.15}$ & 82.52$_{\pm 0.37}$ \\
\algname{FLoRA} [\citeyear{flora}] & 1.48h & 85.87$_{\pm 0.63}$ & 84.85$_{\pm 0.66}$ & 80.36$_{\pm 0.23}$ & 76.45$_{\pm 0.39}$ & \underline{68.52}$_{\pm 1.30}$ & 92.61$_{\pm 0.23}$ & 80.30$_{\pm 1.23}$ & 81.28$_{\pm 0.30}$ \\
\algname{FSLoRA} [\citeyear{Fslora}] & 1.40h & 84.50$_{\pm 0.35}$ & 82.69$_{\pm 0.91}$ & 80.50$_{\pm 0.41}$ & 71.83$_{\pm 1.22}$ & 58.70$_{\pm 2.63}$ & 91.79$_{\pm 0.18}$ & 77.55$_{\pm 0.72}$ & 78.22$_{\pm 0.45}$ \\
\algname{FedSA} [\citeyear{Fedsa}] & 2.64h & 68.73$_{\pm 2.79}$ & 61.69$_{\pm 1.12}$ & 65.24$_{\pm 1.32}$ & 47.62$_{\pm 0.67}$ & 50.85$_{\pm 0.24}$ & 84.40$_{\pm 0.54}$ & 66.58$_{\pm 1.62}$ & 63.59$_{\pm 0.54}$ \\
\algname{RoLoRA} [\citeyear{RoLoRA}] & 1.09h & 82.10$_{\pm 0.49}$ & 81.62$_{\pm 0.63}$ & 78.18$_{\pm 0.47}$ & 56.60$_{\pm 2.51}$ & 58.85$_{\pm 1.39}$ & 91.17$_{\pm 0.17}$ & 74.79$_{\pm 1.33}$ & 74.76$_{\pm 0.47}$ \\
\algname{LA-LoRA} [\citeyear{LALoRA}] & 1.07h & 82.60$_{\pm 0.74}$ & 83.18$_{\pm 0.99}$ & 78.62$_{\pm 0.59}$ & 69.94$_{\pm 1.56}$ & 61.95$_{\pm 1.50}$ & 91.42$_{\pm 0.17}$ & 76.43$_{\pm 1.18}$ & 77.73$_{\pm 0.40}$ \\
\midrule
\algname{SPRA-LoRA} (uniform LR) & 1.50h & 85.62$_{\pm 0.55}$ & 84.95$_{\pm 0.24}$ & 80.04$_{\pm 0.37}$ & 75.03$_{\pm 0.61}$ & 68.09$_{\pm 1.96}$ & 92.59$_{\pm 0.21}$ & 79.31$_{\pm 1.51}$ & 80.80$_{\pm 0.38}$ \\
\rowcolor{lightgray}\algname{SPRA-LoRA} & 1.53h & \textbf{89.03}$_{\pm 0.84}$ & \underline{86.03}$_{\pm 0.42}$ & \textbf{82.09}$_{\pm 0.30}$ & \textbf{81.54}$_{\pm 0.26}$ & 65.63$_{\pm 1.80}$ & \underline{93.19}$_{\pm 0.10}$ & \textbf{83.65}$_{\pm 0.58}$ & \textbf{83.02}$_{\pm 0.31}$ \\
\rowcolor{lightgray}\algname{SPRA-LoRA} (doubled ranks) & 1.66h & \underline{88.94}$_{\pm 0.67}$ & \textbf{86.22}$_{\pm 0.28}$ & 82.09$_{\pm 0.20}$ & \underline{81.50}$_{\pm 0.41}$ & 64.69$_{\pm 2.41}$ & \textbf{93.26}$_{\pm 0.14}$ & \underline{82.73}$_{\pm 0.86}$ & \underline{82.78}$_{\pm 0.39}$ \\
\bottomrule
\end{tabular}}
\end{minipage}\hfill
\label{table:iid_alpha_08_n80}
\end{table*}

\subsection{Visualization of \Cref{lem:noise_var} for $\nu=-0.5$ and $\mu=2$}
 \begin{figure}[H]
  \centering
\includegraphics[width=0.48\textwidth]{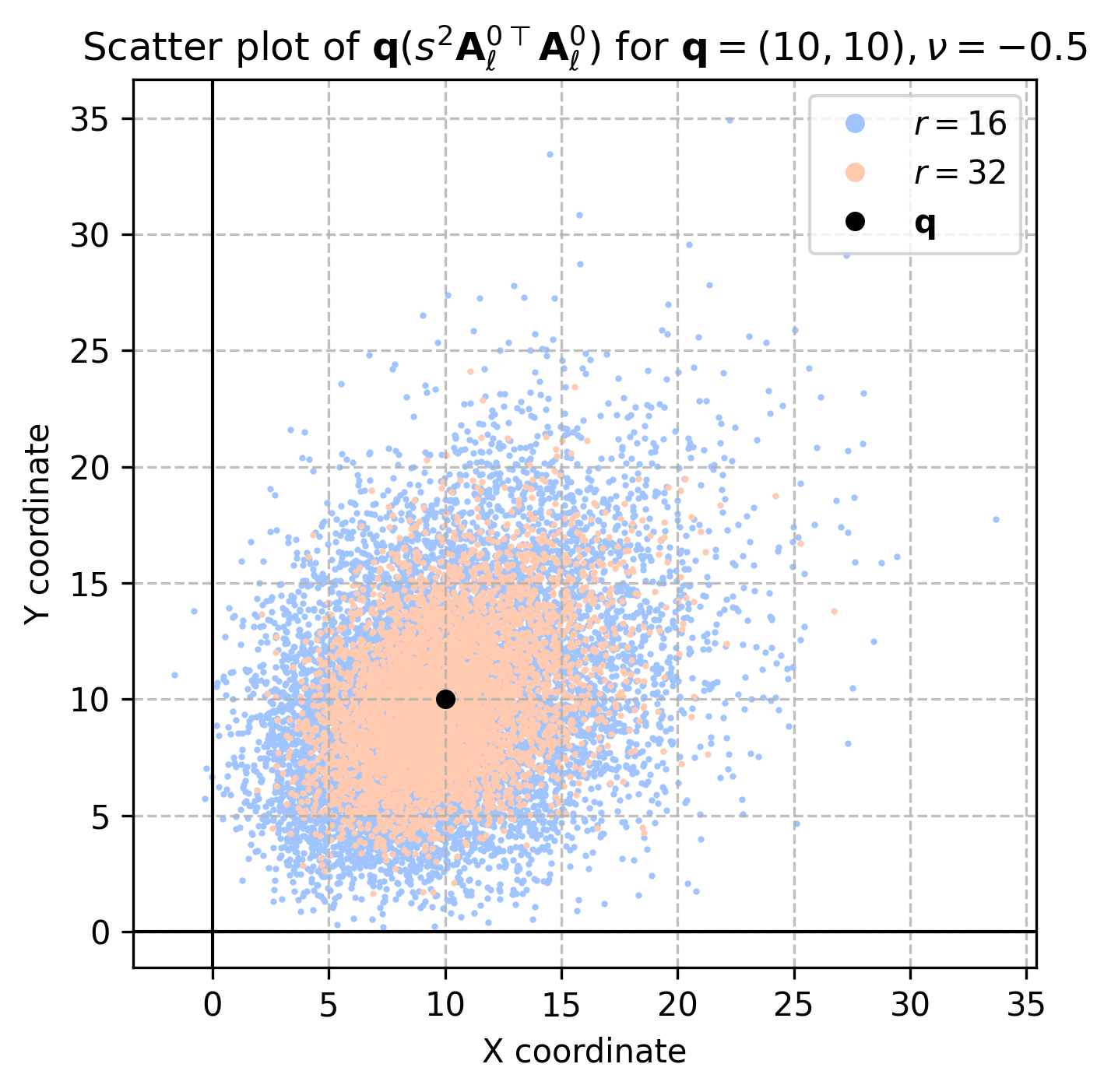}
  \caption{Scatter plot of $\mathbf{q}(s^2\mathbf{A}_\ell^{0\top}\mathbf{A}_\ell^0)$ plotted for $\mathbf{q}=(10,10)$, $10^4$ samples $\mathbf{A}_\ell^0 \sim \mathcal{N}(0, \mathbb{I}_{r \times 2}/r^2)$, $s=\alpha/r^\nu = r^{0.5}$, and two values of adaptation rank $r \in \{\texttt{16,32}\}$. As shown, unlike when $\nu>-0.5$, the perturbation is unbiased for every value of $r$. As $r$ increases, the additive noise magnitude decreases inversely proportional to $r$ and \texttt{SNR} increases linearly with $r$.}
  \label{fig:lora_scaling_v-0.5}
  \vspace{-1em}
\end{figure}

\clearpage
\newpage

\section{Component $\mathbf{A}$ is almost frozen when fine-tuning with \algname{LoRA}}
The following lemma shows that when fine-tuning with \algname{LoRA}, the component matrix $\mathbf{A}$ is almost frozen, compared to the component matrix $\mathbf{B}$.

\begin{restatable}{lemma}{lorafinalforwardpass}
\label{lem:lora_final_full_forwardpass}
Initializing $\mathbf{B}_{\ell}$ with a zero matrix, let \algname{LoRA} update matrices $\mathbf{A}_{\ell}$ and $\mathbf{B}_{\ell}$ with \texttt{SGD} and learning rate $\eta$ for $T$ steps. Also, let us assume $s=1$ for simplicity. Let $\mathbf{W}_{\ell}^t = \mathbf{W}_{\ell}^0 + \mathbf{B}_{\ell}^t\mathbf{A}_{\ell}^t$ be the \textbf{equivalent forward pass parameter} at the end of step $t$. The final equivalent parameter $\mathbf{W}_{\ell}^T$ can be written as: 
\begin{align}\label{eq:noisy_sgd}
    \mathbf{W}_{\ell}^T = \mathbf{W}_{\ell}^0 + \mathbf{B}_{\ell}^T \mathbf{A}_{\ell}^T = \mathbf{W}_{\ell}^0 - \eta \sum_{t=0}^{T-1} \nabla_{\mathbf{W}_{\ell}^t} \mathcal{L}^t \mathbf{A}_{\ell}^0{^\top} \mathbf{A}_{\ell}^0  + \mathcal{O}(\eta^3 ),
\end{align}

where the term $\nabla_{\mathbf{W}_{\ell}^t} \mathcal{L}^t$ in the sum is the full batch gradient at time step $t$ w.r.t. $\mathbf{W}_{\ell}^t = \mathbf{W}_{\ell}^0 + \mathbf{B}_{\ell}^t \mathbf{A}_{\ell}^t$.
\end{restatable}

\begin{proof}
    
Our proof is directly based on the derivations by \cite{flora}. We first restate the following theorem from the work without restating its proof:

\begin{theorem}[\citet{flora}]
\label{thm:lora_dynamics}
Let \algname{LoRA} update matrices $\mathbf{A}_{\ell}$ and $\mathbf{B}_\ell$ with \texttt{SGD} as:

\begin{align*}
    \mathbf{A}_{\ell}^{t+1} \gets \mathbf{A}_{\ell}^t - \eta \frac{\partial \mathcal{L}}{\partial \mathbf{A}_{\ell}^t} = \mathbf{A}_{\ell}^t - \eta \mathbf{B}_{\ell}^t{^\top} (\nabla_{\mathbf{W}_{\ell}^t} \mathcal{L}^t),
\end{align*}

\begin{align*}
    \mathbf{B}_{\ell}^{t+1} \gets \mathbf{B}_{\ell}^t - \eta \frac{\partial \mathcal{L}}{\partial \mathbf{B}_{\ell}^t} = \mathbf{B}_{\ell}^t - \eta (\nabla_{\mathbf{W}_{\ell}^t} \mathcal{L}^t)\mathbf{A}_{\ell}^t{^\top},
\end{align*}

where $\eta$ is the learning rate. We assume $\|\sum_{t=0}^{\tau} \nabla_{\mathbf{W}_{\ell}^t}\mathcal{L}^t\|_F \leq L$ for every $\tau\leq T-1$ during training, implying that the model stays within a finite Euclidean ball from $\mathbf{W}_{\ell}^0$. In this case, the dynamics of $\mathbf{A}_{\ell}^t$ and $\mathbf{B}_{\ell}^t$ are given by
\begin{align}\label{eq:AB_dynamics}
    \mathbf{A}_{\ell}^t = \mathbf{A}_{\ell}^0 + \eta \mathbf{A}_{\ell}^0f_{\mathbf{A}_{\ell}}(t), ~~~ \mathbf{B}_{\ell}^t = \eta f_{\mathbf{B}_{\ell}}(t)\mathbf{A}_{\ell}^0{^\top},
\end{align}

where the forms of $f_{\mathbf{A}_{\ell}}(t)\in \mathbb{R}^{m \times m}$ and $f_{\mathbf{B}_{\ell}}(t) \in \mathbb{R}^{n \times m}$ are as follows for $t\geq 1$:

\begin{align*}
    f_{\mathbf{A}_{\ell}}(t)&= -\eta \sum_{i=0}^{t-1} f_{\mathbf{B}_{\ell}}^{\top}(i)(\nabla_{W^i} \mathcal{L}^i), \\
    f_{\mathbf{B}_{\ell}}(t)&= -\sum_{i=0}^{t-1} (\nabla_{W^i} \mathcal{L}^i)(\eta f_{\mathbf{A}_{\ell}}^{\top}(i)+I),
\end{align*}
and $f_{\mathbf{A}_{\ell}}(0) = f_{\mathbf{B}_{\ell}}(0) = 0$. In particular, $\|f_{\mathbf{A}_{\ell}}(t)\| \leq \frac{\eta L^2 \big(1-(\eta^2 L^2)^t\big)}{1-\eta^2L^2}$ for every $t$.
\end{theorem}

Therefore, we have:

\begin{align*}
    \mathbf{W}_{\ell}^T = \mathbf{W}_{\ell}^0 + \mathbf{B}_{\ell}^T\mathbf{A}_{\ell}^T &= \mathbf{W}_{\ell}^0 + (\eta f_B(T)\mathbf{A}_{\ell}^0{^\top})(\mathbf{A}_{\ell}^0 + \eta \mathbf{A}_{\ell}^0 f_A(T)) \nonumber \\
    & = \mathbf{W}_{\ell}^0 + \eta f_B(T)\mathbf{A}_{\ell}^0{^\top}\mathbf{A}_{\ell}^0 + \eta^2 f_B(T) \mathbf{A}_{\ell}^0{^\top} \mathbf{A}_{\ell}^0 f_A(T).
\end{align*}

After substituting $f_B(T)$ into the last term of the above equation and noting that $f_A(T) \in \mathcal{O}(\eta)$, we get to:

\begin{align*}
    \mathbf{W}_{\ell}^T = \mathbf{W}_{\ell}^0 + \mathbf{B}_{\ell}^T \mathbf{A}_{\ell}^T = \mathbf{W}_{\ell}^0 - \eta \bigg(\sum_{t=0}^{T-1} \nabla_{\mathbf{W}_{\ell}^t} \mathcal{L}^t \mathbf{A}_{\ell}^0{^\top} \mathbf{A}_{\ell}^0 \bigg ) + \mathcal{O}(\eta^3 ),
\end{align*}

which completes the proof.
\end{proof}

\section{Useful lemmas and results}

\begin{theorem}[Chi-Squared distribution: \citep{AlexMood}, Section 4.3, Theorem 7]\label{thm:chisquared}
If the random variables $X_i$, $i=1, \ldots, k$, are normally and independently distributed with means $\mu_i$ and variances $\sigma_i^2$, then

$$
    U = \sum_{i=1}^k \big(\frac{X_i- \mu_i}{\sigma_i}\big)^2
$$

has a chi-squared distribution with $k$ degrees of freedom: $U \sim \mathcal{X}_k^2$. Also, $\mathbb{E}[U]=k$ and $\texttt{Var}[U]=2k$. 
\end{theorem}
The theorem above states that sum of the squares of $k$ standard normal random variables is a chi-squared distribution with $k$ degrees of freedom.

\begin{lemma}[Raw moments of Chi-Squared distribution]\label{lemma:rw_moment_chisquared}
Suppose $X \sim \mathcal{X}_k^2$. Then, the $m$-th raw moment of $X$ can be found as follows;

$$
    \mathbb{E}[X^m] = \prod_{i=0}^{m-1} (k + 2i)
$$
\end{lemma}

\begin{proof}
    From the definition of Chi-Squared distribution with $r$ degrees of reddom, $X$ has the following probability density function:

 $$
        f_{X}(x) = \frac{1}{2^{\frac{k}{2}} \Gamma(\frac{k}{2})} x^{\frac{k}{2}-1}  e^{-\frac{x}{2}}
$$

    Therefore, we have:

    $$
        \mathbb{E}[X^m] = \frac{1}{2^{\frac{k}{2}} \Gamma(\frac{k}{2})} \int_0^{+\infty} x^{\frac{k}{2}+m-1} e^{-\frac{x}{2}} dx = \frac{2}{2^{\frac{k}{2}} \Gamma(\frac{k}{2})} \int_0^{+\infty} (2u)^{\frac{k}{2}+m-1} e^{-u} du 
    $$
    $$
        =\frac{2^{\frac{k}{2}+m-1+1}}{2^{\frac{k}{2}} \Gamma(\frac{k}{2})} \int_0^{+\infty} u^{\frac{k}{2}+m-1} e^{-u} du = \frac{2^m}{\Gamma(\frac{k}{2})}\Gamma (\frac{k}{2}+m) = \frac{2^m \Gamma(\frac{k}{2})}{\Gamma(\frac{k}{2})}\prod_{i=0}^{m-1} (\frac{k}{2}+i) = \prod_{i=0}^{m-1} (k+2i).
    $$
    Note that the fifth equality directly results from the property of gamma function that for $z>0$, $\Gamma(1+z) = z\Gamma(z)$.
\end{proof}

\begin{theorem}[Classical Central Limit Theorem: \citep{patrickB}, Theorem 27.1]\label{thm:CLT}
Suppose that $\{X_i\}_{i=1}^n$, is an independent sequence of random variables having the same distribution with mean $\mu$ and positive variance $\sigma^2$. Define $S_n = \sum_{i=1}^n X_i$ as their sum. Let $Z_n$ be defined by

$$
    Z_n = \frac{S_n - n \mu}{\sqrt{n} \sigma}.
$$

Then, the distribution of $Z_n$ approaches standard normal distribution as $n$ approaches infinity. 

\end{theorem}
The theorem above states that $ S_n$ is approximately, or asymptotically, distributed as a normal distribution with mean $n\mu$ and variance $n \sigma^2$.

The next theorem is about the Lindeberg's condition, which is a sufficient (and under certain conditions also a necessary condition) for the Central Limit Theorem (CLT) to hold for a sequence of independent random variables $\{X_i\}_{i=1}^n$. Unlike the classical CLT stated above, which requires the sequence of random variables to have a finite variance and be both independent and identically distributed (\emph{i.i.d}), Lindeberg's CLT only requires the sequence of random variables to have finite variance, be independent and also satisfy the Lindeberg's condition. The following states the theorem.

\begin{theorem}[Lindeberg and Lyapounov Theorem: \citep{patrickB}, Theorem 27.2]\label{thm:Lindeberg_CLT}
Suppose $X_1, \ldots, X_n$ are $n$ independent random variables with $\mathbb{E}[X_i] = \mu_i$ and $\texttt{Var}[X_i]=\sigma_i^2>0$. Define $S_n = \sum_{i=1}^n X_i$ and let  $s_n^2 = \sum_{i=1}^n \sigma_i^2$. 
Also assume the following condition holds for all $\epsilon >0$:

$$
    \textit{Lindeberg's condition:}~~~\lim_{n \to \infty} \sum_{i=1}^n \frac{1}{s_n^2} \int_{|x-\mu_i|\geq \epsilon s_n} (x - \mu_i)^2 P_{X_i}(x) dx =  0,
$$

where $P_{X_i}$ is the pdf of variable $X_i$. Assuming $Z_n =\frac{S_n - \sum_{i=1}^n \mu_i}{s_n}$, the distribution of $Z_n$ approaches standard normal distribution as $n$ approaches infinity. 

\end{theorem}

The theorem above states that, given that Lindeberg's condition is satisfied, $ S_n$ is approximately, or asymptotically, distributed as a normal distribution with mean $\sum_{i=1}^n \mu_i$ and variance $s_n^2$, even if the sequence of variables are not identically distributed.

\section{Proof of \Cref{lem:noise_var}, with Gaussian initialization of $\{\mathbf{A}_\ell\}_{\ell=1}^L$}

\noisevar*
\begin{proof}
Let us denote $\mathbf{A}_\ell^0$ with $\mathbf{A}_\ell$ for simplicity. We first derive the distribution of $\mathbf{q} \cdot (\mathbf{A}_{\ell}^\top \mathbf{A}_{\ell}-\mathbb{I}_m)$ for simplicity of notations. Then, we translate it to the distribution of $\mathbf{q} \cdot ((s\mathbf{A}_{\ell}^\top) (s\mathbf{A}_{\ell})-\mathbb{I}_m) \in \mathbb{R}^{1 \times m}$. Finally, distribution of $\mathbf{q} (s^2\mathbf{A}_{\ell}^\top\mathbf{A}_{\ell}) \in \mathbb{R}^{1 \times m}$ will be obtained after a shift in the mean with $\mathbf{q}$.

From the theorem's assumption, we know that elements of $\mathbf{A}_{\ell}$ are from $\mathcal{N}(0, \sigma_{\mathbf{A}}^2\mathbb{I}_{r \times m})$. Therefore, we can rewrite the product $\mathbf{q} \cdot (\mathbf{A}_{\ell}^\top \mathbf{A}_{\ell}-\mathbb{I}_m) \in \mathbb{R}^{1 \times m}$ as the following product:

\begin{align}\label{eq:reparam}
    \mathbf{q}\cdot (\sigma_{\mathbf{A}}^2 \mathbf{A}_{\ell}^\top \mathbf{A}_{\ell}-\mathbb{I}_m) \in \mathbb{R}^{1 \times m}
\end{align}
where \emph{the elements of $\mathbf{A}_{\ell}$ are now from standard normal distribution}. Let $a_{i,j}$ denote the element in $i$-th row and $j$-th column of this new standard normal $\mathbf{A}_{\ell}$. Therefore, for all $i$ and $j$, $a_{i,j}$ has distribution $\mathcal{N}(0, 1)$. Let $\mathbf{B}=\sigma_{\mathbf{A}}^2 \mathbf{A}_{\ell}^\top \mathbf{A}_{\ell}-\mathbb{I}_m$. Also, let $\mathbf{A}_{i,:}$ and $\mathbf{A}_{:,j}$ denote the $i$-th row and $j$-th column of the new $\mathbf{A}_{\ell}$, respectively. We have:

$$
    \mathbf{B}_{i,i} = \sigma_{\mathbf{A}}^2[\mathbf{A}_{\ell}^{\top} \mathbf{A}_\ell]_{i,i}-1 = \sigma_{\mathbf{A}}^2 \mathbf{A}_{:,i}^{\top} \mathbf{A}_{:,i}-1 = \sigma_{\mathbf{A}}^2 \|\mathbf{A}_{:,i}\|_2^2 -1 = (\sigma_{\mathbf{A}}^2 \sum_{l=1}^r a_{l,i}^2) -1
$$

From \cref{eq:reparam}, we know that $a_{l,i}$ is from standard normal distribution. Hence, $a_{l,i}^2$ is a chi-squared with 1 degree of freedom: $a_{l,i}^2 \sim \mathcal{X}_1^2$. Therefore, $\sum_{l=1}^r a_{l,i}^2$, which is the sum of $r$ independent chi-squared variables with 1 degree of freedom, is a chi-squared with $r$ degrees of freedom: $\sum_{l=1}^r a_{l,i}^2 \sim \mathcal{X}_{r}^2$ (see \cref{thm:chisquared}). Therefore, for $i \in \{1,\ldots,m\}$, we have: 

\begin{align} \label{eq:diagonal_mean_var}
    &\mathbb{E}[\mathbf{B}_{i,i}] = \mathbb{E}\big[\sigma_{\mathbf{A}}^2 \sum_{l=1}^r a_{l,i}^2\big] -1  = r \sigma_{\mathbf{A}}^2 -1, \nonumber
    \\
    &\texttt{Var}[\mathbf{B}_{i,i}] = \texttt{Var}[\sigma_{\mathbf{A}}^2 \sum_{l=1}^r a_{l,i}^2] = \sigma_\mathbf{A}^4 \texttt{Var}(\mathcal{X}_r^2) = 2r\sigma_\mathbf{A}^4.
\end{align}

Similarly, we find the mean and variance of the non-diagonal elements $\mathbf{B}_{i,j} (i\neq j)$ of $\mathbf{B}$. We have:

\begin{align}\label{eq:nondiagonal}
    \mathbf{B}_{i,j} = \sigma_{\mathbf{A}}^2[\mathbf{A}_{\ell}^{\top} \mathbf{A}_\ell]_{i,j} = \sigma_{\mathbf{A}}^2 \mathbf{A}_{:,i}^{\top} \mathbf{A}_{:,j} = \sigma_{\mathbf{A}}^2 \sum_{l=1}^r a_{l,i} a_{l,j},
\end{align}

where $a_{l,i}$ and $a_{l,j}$ are independent and standard normal. Therefore, $a_{l,i} + a_{l,j} \sim \mathcal{N}(0,2)$. Similarly, $a_{l,i} - a_{l,j} \sim \mathcal{N}(0,2)$. So we can rewrite $a_{l,i} a_{l,j}$ as:

$$
    a_{l,i} a_{l,j} = \frac{1}{4}(a_{l,i} + a_{l,j})^2 - \frac{1}{4}(a_{l,i} - a_{l,j} )^2 = \frac{1}{2}z_1^2 - \frac{1}{2}z_2^2,
$$
where $z_1$ and $z_2$ are from standard normal. Therefore, $a_{l,i} a_{l,j} = \frac{\nu_1 - \nu_2}{2}$, where $\nu_1, \nu_2 \sim \mathcal{X}_1^2$. Also, $a_{l,i} + a_{l,j}$ and $a_{l,i} - a_{l,j}$ are independent variables (because both $a_{l,i}$ and $a_{l,j}$ are Gaussian, independent, and have equal variance). Hence, $z_1$ and $z_2$ are independent, and likewise $\nu_1$ and $\nu_2$ are independent. We conclude that: 

\begin{align}\label{eq:cross_product}
a_{l,i} a_{l,j} = \frac{1}{2}(\nu_1 - \nu_2),
\end{align}

 where $\nu_1, \nu_2 \sim \mathcal{X}_1^2$, and are independent. 
 
 Now, let us assume $\nu_1, \nu_2 \sim \mathcal{X}_k^2$ (a more general case), and let $M_{\nu_1}(t) = \mathbb{E}[e^{t\nu_1}]$ be the moment generating function (MGF) of $\nu_1$. In this case, we know that $M_{\nu_1}(t) = M_{\nu_2}(t) = (1-2t)^{-\frac{k}{2}}$ (MGF of $\mathcal{X}_k^2$). Hence, $M_{\nu_1-\nu_2}(t) = M_{\nu_1}(t)\cdot M_{\nu_2}(-t) = (1-4t^2)^{-\frac{k}{2}} = \big( \frac{\frac{1}{4}}{\frac{1}{4}-t^2}\big)^{\frac{k}{2}}$, which is the MGF of a symmetric about origin \texttt{Variance-Gamma} distribution with parameters $\lambda = \frac{k}{2}, \alpha=\frac{1}{2}, \beta=0, \mu=0, \gamma = \frac{1}{2}$. Therefore, when $\nu_1, \nu_2 \sim \mathcal{X}_k^2$, then $\nu_1 - \nu_2$ has this distribution, which has mean $\mu + 2 \beta \lambda/\gamma^2=0$ and variance $2 \lambda (1+2 \beta^2/\gamma^2)/\gamma^2=4k$. 
 
 In \cref{eq:cross_product}, we had $k=1$, as we had $\nu_1, \nu_2 \sim \mathcal{X}_1^2$. Hence, based on the discussion above, we have:

 $$
     \mathbb{E}[a_{l,i} a_{l,j}] = 0 
$$
$$
     \texttt{Var}[a_{l,i} a_{l,j}] = \frac{1}{4}\texttt{Var}[\nu_1 - \nu_2] = \frac{4k}{4}|_{k=1} = 1
$$

Consequently, based on \cref{eq:nondiagonal} and from the results above, we can compute the mean and variance of the non-diagonal elements of $\mathbf{B}$ ($i \neq j$):

\begin{align}\label{eq:nondiagonal_mean_var}
    \mathbb{E}[\mathbf{B}_{i,j}] &= \mathbb{E}\big[\sigma_{\mathbf{A}}^2 \sum_{l=1}^r a_{l,i}a_{l,j}\big] = \sigma_{\mathbf{A}}^2 \sum_{l=1}^r \mathbb{E}[a_{l,i}a_{l,j}] = 0, \nonumber 
    \\
    \texttt{Var}[\mathbf{B}_{i,j}] &= \texttt{Var}[\sigma_{\mathbf{A}}^2\sum_{l=1}^r a_{l,i}a_{l,j}] = \sigma_\mathbf{A}^4 \sum_{l=1}^r \texttt{Var}[a_{l,i}a_{l,j}] = r \sigma_\mathbf{A}^4.
\end{align}

So far, we have computed the mean and variance of each entry in $\mathbf{B}= \sigma_{\mathbf{A}}^2\mathbf{A}_{\ell}^\top \mathbf{A}_{\ell}-\mathbb{I}_m \in \mathbb{R}^{m \times m}$ in \cref{eq:diagonal_mean_var} and \cref{eq:nondiagonal_mean_var}. Now, for a given $\mathbf{q}\in \mathbb{R}^{1\times m}$, we have:

\begin{align} \label{eq:qB}
    \mathbf{q}\cdot \mathbf{B}  = \sum_{l=1}^m \mathbf{q}_l \mathbf{B}_{l,:},
\end{align}
where $\mathbf{B}_{l,:}$ is row $l$ of $\mathbf{B}$. Let $u_i$ denote the $i$-th element of $\mathbf{q}\cdot \mathbf{B}$. Hence, for each element $u_i$ ($i \in \{1,\ldots, m\}$), we have:

\begin{align}\label{eq:mean_var}
\mathbb{E}[u_i] &= \mathbb{E}\bigg[ \sum_{l=1}^m \mathbf{q}_l \mathbf{B}_{l,i}\bigg] =\mathbf{q}_i \mathbb{E}[\mathbf{B}_{i,i}] + \sum_{l\neq i} \mathbf{q}_l \mathbb{E}[\mathbf{B}_{l,i}] = (r\sigma_{\mathbf{A}}^2 - 1)\mathbf{q}_i, \nonumber\\
\texttt{Var}[u_i] &= \texttt{Var}\bigg[ \sum_{l=1}^m \mathbf{q}_l \mathbf{B}_{l,i}\bigg] =\texttt{Var}\bigg[ \sum_{l=1}^m \mathbf{q}_l \sigma_{\mathbf{A}}^2 \sum_{t=1}^r a_{t,l} a_{t,i}\bigg] \nonumber \\
& = \sum_{l=1}^m \mathbf{q}_l^2\sigma_\mathbf{A}^4 \sum_{t=1}^r \texttt{Var}[a_{t,l} a_{t,i}] = 2 \mathbf{q}_i^2 r\sigma_\mathbf{A}^4 + \sum_{l\neq i} \mathbf{q}_l^2 r\sigma_\mathbf{A}^4\nonumber\\
& = \mathbf{q}_i^2 r\sigma_\mathbf{A}^4 + \sum_{l=1}^m \mathbf{q}_l^2 r\sigma_\mathbf{A}^4 = r\sigma_\mathbf{A}^4 (\mathbf{q}_i^2 + \|\mathbf{q}\|_2^2). 
\nonumber \\
\end{align}

In the second last line we could take the \texttt{Var} into the summations because every pair of the $mr$ summands are uncorrelated (so variance of their sum is sum of their variances). 

Finally, we check if the conditions of \Cref{thm:Lindeberg_CLT} hold for each element $u_i$ of $\mathbf{q}\cdot \mathbf{B}$. According to \cref{eq:qB}, each element $u_i$ of $\mathbf{q}\cdot \mathbf{B}$ can be rewritten as the following sum:

\begin{align}
    u_i = \sum_{l=1}^m \mathbf{q}_l \mathbf{B}_{l,i} = \sum_{l=1}^m \mathbf{q}_l \sigma_{\mathbf{A}}^2 \sum_{t=1}^r a_{t,l} a_{t,i} = \sum_{t=1}^r \bigg(\sigma_{\mathbf{A}}^2\sum_{l=1}^m \mathbf{q}_l a_{t,l} a_{t,i}\bigg).
\end{align}

Note every pair of the $mr$ summands are uncorrelated. However, not all of them are independent. Yet, every pair of the $r$ summands $\sum_{l=1}^m \mathbf{q}_l a_{t,l} a_{t,i}$ and $\sum_{l=1}^m \mathbf{q}_l a_{t',l} a_{t',i}$ are independent (for $t\neq t'$). Therefore, each $u_i$ is the sum of $r$ \emph{independent} random variables. Also, each of the $r$ summands has variance $\sigma_\mathbf{A}^4 (\mathbf{q}_i^2 + \|\mathbf{q}\|_2^2)$. Hence, the sum of their variances of the $r$ independent variables constituting $u_i$ is $s_r^2 = r\sigma_\mathbf{A}^4 (\mathbf{q}_i^2 + \|\mathbf q\|_2^2)$, which is the same as the variance we found in \Cref{eq:mean_var}. Assuming a non-zero $\mathbf{q}$, $s_r^2 \to \infty$, as $r \to \infty$. As each of the $r$ summands $\sum_{l=1}^m \mathbf{q}_l a_{t,l} a_{t,i}$, for $t \in \{1,\cdots,r\}$, has a constant variance $\sigma_\mathbf{A}^4 (\mathbf{q}_i^2 + \|\mathbf{q}\|_2^2)$ and a vanishing distribution tail, the Lindeberg's condition is also satisfied for every element $u_i$. Therefore, from \cref{thm:Lindeberg_CLT}, we conclude that as $r \to \infty$, each of the $m$ elements of $\mathbf{q}\cdot \mathbf{B}$ (i.e., each element in $\{u_i\}_{i=1}^m$) approaches a Gaussian with its specific mean and variance found in \cref{eq:mean_var}.

However, the $m$ elements of $\mathbf{q}\cdot \mathbf{B}$ are not completely independent and are positively correlated. So the covariance matrix of $\mathbf{q}\cdot \mathbf{B}$ is not diagonal. Yet, for $\sigma_{\mathbf{A}}^2 = 1/r^\mu$ with $\mu>0.5$, the relative impact of the dependency between the elements vanishes as $r$ grows: according to \Cref{eq:mean_var}, $\mathbf{q} \cdot \mathbf{B}$ becomes increasingly "deterministic" and aligned with $-\mathbf{q}$, making its $m$ elements effectively independent of each other. Hence, as $r$ grows, the covariance matrix $\mathbf{\Sigma}\in \mathbb{R}^{m \times m}$ of $\mathbf{q}\cdot \mathbf{B}$ approaches a diagonal matrix with off-diagonal elements approaching zero and the $i$-th diagonal element coming from \Cref{eq:mean_var}. Therefore, we conclude that for $\mathbf{A}_{\ell} \sim \mathcal{N}(0, \sigma_{\mathbf{A}}^2\mathbb{I}_{r \times m})$ with $\sigma_{\mathbf{A}}^2 = 1/r^\mu$ $(\mu>0.5)$, $\mathbf{q}\cdot (\mathbf{A}_{\ell}^\top \mathbf{A}_{\ell}-\mathbb{I}_m) \in \mathbb{R}^{1 \times m}$ approaches the Gaussian distribution $\mathcal{N}\big((r\sigma_{\mathbf{A}}^2 - 1)~\mathbf{q}, ~r\sigma_{\mathbf{A}}^4 (\|\mathbf{q}\|_2^2 ~ \mathbb{I}_m + \mathbf{D})\big)$, with $\mathbf{D} = diag(\mathbf{q}^2)$, as $r$ increases. 

Having found the asymptotic distribution of $\mathbf{q}\cdot (\mathbf{A}_{\ell}^\top \mathbf{A}_{\ell}-\mathbb{I}_m)$ above, we can directly translate it to the distribution of $\mathbf{q}\cdot \big((s\mathbf{A}_{\ell})^\top (s\mathbf{A}_{\ell})-\mathbb{I}_m \big)$, by replacing $\sigma_{\mathbf{A}}^2$ with $s^2\sigma_{\mathbf{A}}^2$, resulting in the asymptotic Gaussian distribution $\mathcal{N}\big((rs^2\sigma_{\mathbf{A}}^2 - 1)~\mathbf{q}, ~rs^4\sigma_{\mathbf{A}}^4 (\|\mathbf{q}\|_2^2 ~ \mathbb{I}_m + \mathbf{D})\big)$. 

By shifting the mean by $\mathbf{q}$, we get distribution $\mathcal{N}\big(rs^2\sigma_{\mathbf{A}}^2 ~\mathbf{q}, ~rs^4\sigma_{\mathbf{A}}^4 (\|\mathbf{q}\|_2^2 ~ \mathbb{I}_m + \mathbf{D})\big)$ for $\mathbf{q}\cdot s^2\mathbf{A}_{\ell}^\top\mathbf{A}_{\ell}$. Finally, after replacing $s = \alpha/r^\nu$ for \algname{LoRA} and $\sigma_{\mathbf{A}}^2 = 1/r^\mu$, we get to the following distribution

$$
\mathcal{N}\big(\alpha^2 \mathbf{q}/r^{2\nu+\mu-1}, (\|\mathbf{q}\|^2 \mathbb{I}_m + \mathbf{D})\alpha^4/r^{4\nu+2\mu-1}\big).
$$
Finally, note that $s^2\sigma_{\mathbf{A}}^2 = \alpha^2/r^{2\nu+\mu}$, which imposes the constraint $2\nu+\mu> 0.5$. This completes the proof. 

\end{proof}

\section{Extension of \Cref{lem:noise_var} to Kaiming-Uniform initialization of $\{\mathbf{A}_\ell\}_{\ell=1}^L$}\label{app:app_kaiming_init}

We now extend the result derived in \Cref{lem:noise_var} to Kaiming-Uniform initialization of $\{\mathbf{A}_\ell\}_{\ell=1}^L$ to show that a similar rank-dependent behavior exists.

Kaiming uniform initialization \citep{kaiming}, is primarily designed for neural networks employing \texttt{ReLU}-like activation functions. While it was originally derived for \texttt{ReLU}, its principles can be extended to other non-linear activations like \texttt{GELU}, which is a smoother variant of \texttt{ReLU}.
The core idea of Kaiming initialization is to set the initial weights in a way that helps maintain a consistent variance of activations throughout the network, preventing vanishing or exploding gradients during training. For uniform distribution, the weights are drawn from a uniform distribution within a specific range $\mathcal{U}(-b, b)$, where $b=\sqrt{6/m}$. The value of $b$ is found for each layer such that the variance of each subsequent pre-activation layer is equal, e.g. in \Cref{fig:lora}, the variance of each element in $y$ is equal to that of each element in the pre-activation of input $x$. We have the following lemma about this initialization.

\begin{restatable}{lemma}{noisevar_kaiming}\label{lem:noise_var_kaiming}
Consider a fixed $\mathbf{q} \in \mathbb{R}^{1 \times m}$ independent of $\mathbf{A}_\ell^0 \in \mathbb{R}^{r \times m}\sim \mathcal{U}(-b, b)$, and assume $r \ll m$ holds. Random perturbation $\mathbf{q}(s^2\mathbf{A}_\ell^{0\top}\mathbf{A}_\ell^0)$ approaches the $m$-dimensional distribution $\mathcal{N}\big(\alpha^2\sigma_{\mathbf{A}}^2~\mathbf{q}/r^{2\nu - 1}, ~\alpha^4\sigma_{\mathbf{A}}^4 (\|\mathbf{q}\|_2^2 ~ \mathbb{I}_m + \mathbf{D})/r^{4\nu-1}\big)$, as $r$ increases. Also, $\sigma_{\mathbf{A}}^2=b^2/3$ is the variance of each element in $\mathbf{A}_{\ell}$ and $\mathbf{D}\in \mathbb{R}^{m\times m}$ is a diagonal matrix with $\mathbf{D}_{i,i}=-\mathbf{q}_i^2/5$. 
\end{restatable}

\begin{proof}

Let us denote $\mathbf{A}_\ell^0$ with $\mathbf{A}_\ell$ for simplicity. We first derive the distribution of $\mathbf{q} \cdot (\mathbf{A}_{\ell}^\top \mathbf{A}_{\ell}-\mathbb{I}_m)$ for simplicity of notations. Then, we translate it to the distribution of $\mathbf{q} \cdot ((s\mathbf{A}_{\ell}^\top) (s\mathbf{A}_{\ell})-\mathbb{I}_m) \in \mathbb{R}^{1 \times m}$. Finally, distribution of $\mathbf{q} (s^2\mathbf{A}_{\ell}^\top\mathbf{A}_{\ell}) \in \mathbb{R}^{1 \times m}$ will be obtained after a shift in the mean with $\mathbf{q}$.

The elements of $\mathbf{A}_{\ell}$ are from $\mathcal{U}(-b, b)$, which has variance $\sigma_{\mathbf{A}}^2=b^2/3$. Let $a_{i,j}$ denote the element in $i$-th row and $j$-th column of $\mathbf{A}_{\ell}$. Therefore, for all $i$ and $j$, $a_{i,j}$ has distribution $\mathcal{U}(-b, b)$. Let $\mathbf{B}= \mathbf{A}_{\ell}^\top \mathbf{A}_{\ell}-\mathbb{I}_m$. Also, let $A_{i,:}$ and $A_{:,j}$ denote the $i$-th row and $j$-th column of $\mathbf{A}_{\ell}$, respectively. We have:

$$
    \mathbf{B}_{i,i} = [\mathbf{A}_{\ell}^{\top} \mathbf{A}]_{i,i}-1 = \mathbf{A}_{:,i}^{\top} \mathbf{A}_{:,i}-1 =  \|\mathbf{A}_{:,i}\|_2^2 -1 = \big(\sum_{l=1}^r a_{l,i}^2 \big) -1
$$

We know that $a_{l,i}$ is from $\mathcal{U}(-b,b)$ distribution. Hence, $a_{l,i}^2/b^2$ follows \texttt{Beta}(1/2, 1) distribution. Equivalently, $\mathbb{E}[a_{l,i}^2]= \texttt{Var}[a_{l,i}] = \sigma_{\mathbf{A}}^2 = b^2/3$ and $\texttt{Var}[a_{l,i}^2]= 4b^4/45$. Therefore, for $\mathbf{B}_{i,i} = \sum_{l=1}^r a_{l,i}^2 -1$ ($i \in \{1,\ldots,m\}$),  which is the sum of $r$ independent such variables, we have: 

\begin{align} \label{eq:diagonal_mean_var_uniform}
    &\mathbb{E}[\mathbf{B}_{i,i}] = \mathbb{E}\big[ \sum_{l=1}^r a_{l,i}^2\big] -1  = \frac{rb^2}{3} -1, \nonumber
    \\
    &\texttt{Var}[\mathbf{B}_{i,i}] = \texttt{Var}[\sum_{l=1}^r a_{l,i}^2] = \frac{4rb^4}{45}.
\end{align}

Similarly, we find the mean and variance of the non-diagonal elements $\mathbf{B}_{i,j} (i\neq j)$ of $\mathbf{B}$. We have:

$$
    \mathbf{B}_{i,j} = [\mathbf{A}_{\ell}^{\top} A]_{i,j} =  \mathbf{A}_{:,i}^{\top} \mathbf{A}_{:,j} = \sum_{l=1}^r a_{l,i} a_{l,j},
$$

where $a_{l,i}$ and $a_{l,j}$ are independent. Therefore:

\begin{align*}
    \mathbb{E}[a_{l,i} a_{l,j}] &= 0
    \nonumber \\
    \texttt{Var}[a_{l,i} a_{l,j}] &= \frac{5b^4}{45}.
\end{align*}

Consequently, we can compute the mean and variance of the non-diagonal elements of $\mathbf{B}$ ($i \neq j$):

\begin{align}\label{eq:nondiagonal_mean_var_uniform}
    &\mathbb{E}[\mathbf{B}_{i,j}] = \mathbb{E}\big[ \sum_{l=1}^r a_{l,i}a_{l,j}\big] = 0, \nonumber 
    \\
    &\texttt{Var}[\mathbf{B}_{i,j}] = \texttt{Var}[\sum_{l=1}^r a_{l,i}a_{l,j}] = \sum_{l=1}^r \texttt{Var}[a_{l,i}a_{l,j}] = \frac{5rb^4}{45}.
\end{align}

In the last line, we could take the \texttt{Var} into the summation because the $r$ summands are independent.

So far, we have computed the mean and variance of each entry in $\mathbf{B}= \mathbf{A}_{\ell}^\top \mathbf{A}_{\ell}-\mathbb{I}_m \in \mathbb{R}^{m \times m}$. Now, for a given $\mathbf{q}\in \mathbb{R}^{1\times m}$, we have:

\begin{align}\label{eq:qB_uniform}
    \mathbf{q}\cdot \mathbf{B}  = \sum_{l=1}^m \mathbf{q}_l \mathbf{B}_{l,:},
\end{align}

where $\mathbf{B}_{l,:}$ is row $l$ of $\mathbf{B}$. Let $u_i$ denote the $i$-th element of $\mathbf{q}\cdot \mathbf{B}$. Hence, for each element $u_i$ ($i \in \{1,\ldots, m\}$), we have:

\begin{align}\label{eq:mean_var_uniform}
\mathbb{E}[u_i] &= \mathbb{E}\bigg[ \sum_{l=1}^m \mathbf{q}_l \mathbf{B}_{l,i}\bigg] =\mathbf{q}_i \mathbb{E}[\mathbf{B}_{i,i}] + \sum_{l\neq i} \mathbf{q}_l \mathbb{E}[\mathbf{B}_{l,i}] = (rb^2/3 - 1)\mathbf{q}_i= (r\sigma_{\mathbf{A}}^2 - 1)\mathbf{q}_i, \nonumber\\
\texttt{Var}[u_i] &= \texttt{Var}\bigg[ \sum_{l=1}^m \mathbf{q}_l \mathbf{B}_{l,i}\bigg] =\texttt{Var}\bigg[ \sum_{l=1}^m \mathbf{q}_l  \sum_{t=1}^r a_{t,l} a_{t,i}\bigg] = \texttt{Var}\bigg[\mathbf{q}_i \sum_{t=1}^r a_{t,i}^2 + \sum_{l\neq i, l=1}^m \mathbf{q}_l  \sum_{t=1}^r a_{t,l} a_{t,i}\bigg] \nonumber \\
& = \sum_{t=1}^r \mathbf{q}_i^2 \texttt{Var}[a_{t,i}^2] + \sum_{l\neq i, l=1}^m \mathbf{q}_l^2  \sum_{t=1}^r \texttt{Var}[a_{t,l} a_{t,i}]  \nonumber \\
& = \frac{4rb^4}{45} \mathbf{q}_i^2 + \sum_{l\neq i} \frac{5rb^4}{45}\mathbf{q}_l^2  = \frac{5rb^4}{45} \|\mathbf{q}\|_2^2 - \frac{rb^4}{45}\mathbf{q}_i^2 = r\sigma_{\mathbf{A}}^4 (\|\mathbf{q}\|_2^2 - \frac{\mathbf{q}_i^2}{5}).
\end{align}

In the second last line we could take the \texttt{Var} into the summations because every pair of the $mr$ summands are uncorrelated (so variance of their sum is sum of their variances). 

Finally, we check if the conditions of \Cref{thm:Lindeberg_CLT} hold for each element $u_i$ of $\mathbf{q}\cdot \mathbf{B}$. According to \cref{eq:qB_uniform}, each element $u_i$ of $\mathbf{q}\cdot \mathbf{B}$ can be rewritten as the following sum:

\begin{align}
    u_i = \sum_{l=1}^m \mathbf{q}_l \mathbf{B}_{l,i} = \sum_{l=1}^m \mathbf{q}_l \sigma_{\mathbf{A}}^2 \sum_{t=1}^r a_{t,l} a_{t,i} = \sum_{t=1}^r \bigg(\sigma_{\mathbf{A}}^2\sum_{l=1}^m \mathbf{q}_l a_{t,l} a_{t,i}\bigg).
\end{align}

Note every pair of the $mr$ summands are uncorrelated. However, not all of them are independent. Yet, every pair of the $r$ summands $\sum_{l=1}^m \mathbf{q}_l a_{t,l} a_{t,i}$ and $\sum_{l=1}^m \mathbf{q}_l a_{t',l} a_{t',i}$ are independent (for $t\neq t'$). Therefore, each $u_i$ is the sum of $r$ \emph{independent} random variables. Also, each of the $r$ summands has variance $\sigma_\mathbf{A}^4 ( \|\mathbf{q}\|_2^2 - \mathbf{q}_i^2/5)$. Hence, the sum of the variances of the $r$ independent variables constituting $u_i$ is $s_r^2 = r\sigma_\mathbf{A}^4 ( \|\mathbf q\|_2^2 - \mathbf{q}_i^2/5)$, which is the same as the variance we found in \Cref{eq:mean_var_uniform}. Assuming a non-zero $\mathbf{q}$, $s_r^2 \to \infty$, as $r \to \infty$. As each of the $r$ summands $\sum_{l=1}^m \mathbf{q}_l a_{t,l} a_{t,i}$, for $t \in \{1,\cdots,r\}$, has a constant variance $\sigma_\mathbf{A}^4 ( \|\mathbf{q}\|_2^2 -\mathbf{q}_i^2/5)$ and a vanishing distribution tail, the Lindeberg's condition is also satisfied for every element $u_i$. Therefore, from \cref{thm:Lindeberg_CLT}, we conclude that as $r \to \infty$, each of the $m$ elements of $\mathbf{q}\cdot \mathbf{B}$ (i.e., each element in $\{u_i\}_{i=1}^m$) approaches a Gaussian with its specific mean and variance found in \cref{eq:mean_var_uniform}.

However, the $m$ elements of $\mathbf{q}\cdot \mathbf{B}$ are not independent and are positively correlated. So the covariance matrix of $\mathbf{q}\cdot \mathbf{B}$ is not completely diagonal. Yet, according to \Cref{eq:mean_var_uniform} and as long as $r \sigma_{\mathbf{A}}^4 \ll 1$, $\mathbf{q} \cdot \mathbf{B}$ is almost "deterministic" and aligned with $-\mathbf{q}$, making its $m$ elements effectively independent of each other. In other words, the relative impact of the dependency between the elements vanishes as long as $r \sigma_{\mathbf{A}}^4\ll 1$, or equivalently $rb^4 \ll 1$ (remember that $\sigma_{\mathbf{A}}^2 = b^2/3$). Also, remember that for Kaiming-Uniform initialization with ReLU/GeLU activation, $b=\sqrt{6/m}$. Hence, the condition $r b^4\ll 1$ reduces to condition $r/m^2 \ll 1$, which holds from the assumption $r/m \ll 1$. Under this condition, the covariance matrix $\mathbf{\Sigma}\in \mathbb{R}^{m \times m}$ of $\mathbf{q}\cdot \mathbf{B}$ is almost diagonal with off-diagonal elements approaching zero and the $i$-th diagonal element coming from the variance in \Cref{eq:mean_var_uniform}. Therefore, we conclude that, under $r \ll m$ and as $r$ increases, $\mathbf{q}\cdot (\mathbf{A}_{\ell}^\top \mathbf{A}_{\ell}-\mathbb{I}_m) \in \mathbb{R}^{1 \times m}$ approaches a Gaussian $\mathcal{N}\big((r\sigma_{\mathbf{A}}^2 - 1)~\mathbf{q}, ~r\sigma_{\mathbf{A}}^4 (\|\mathbf{q}\|_2^2 ~ \mathbb{I}_m + \mathbf{D})\big)$, where $\mathcal{D}$ is diagonal and $\mathcal{D
}_{i,i} = -\mathbf{q}_i^2/5$.

Having found the asymptotic distribution of $\mathbf{q}\cdot (\mathbf{A}_{\ell}^\top \mathbf{A}_{\ell}-\mathbb{I}_m)$ above, we can directly translate it to the distribution of $\mathbf{q}\cdot \big((s\mathbf{A}_{\ell})^\top (s\mathbf{A}_{\ell})-\mathbb{I}_m \big)$, by replacing $\sigma_{\mathbf{A}}^2$ with $s^2\sigma_{\mathbf{A}}^2$, resulting in the asymptotic Gaussian distribution $\mathcal{N}\big((rs^2\sigma_{\mathbf{A}}^2 - 1)~\mathbf{q}, ~rs^4\sigma_{\mathbf{A}}^4 (\|\mathbf{q}\|_2^2 ~ \mathbb{I}_m + \mathbf{D})\big)$. 

By shifting the mean of the distribution, we get distribution $\mathcal{N}\big(rs^2\sigma_{\mathbf{A}}^2 ~\mathbf{q}, ~rs^4\sigma_{\mathbf{A}}^4 (\|\mathbf{q}\|_2^2 ~ \mathbb{I}_m + \mathbf{D})\big)$ for $\mathbf{q}\cdot s^2\mathbf{A}_{\ell}^\top\mathbf{A}_{\ell}$. Finally, after replacing $s = \alpha/r^\nu$ for \algname{LoRA}, we get to the following distribution:

$$
\mathcal{N}\big(\alpha^2\sigma_{\mathbf{A}}^2~\mathbf{q}/r^{2\nu - 1}, ~\alpha^4\sigma_{\mathbf{A}}^4 (\|\mathbf{q}\|_2^2 ~ \mathbb{I}_m + \mathbf{D})/r^{4\nu-1}\big).
$$

Finally, note that condition $rs^4\sigma_{\mathbf{A}}^4 \ll 1$ is satisfied if $r \ll m$. This completes the proof. 
\end{proof}

\begin{figure}
\centering
\includegraphics[width=0.329\columnwidth]{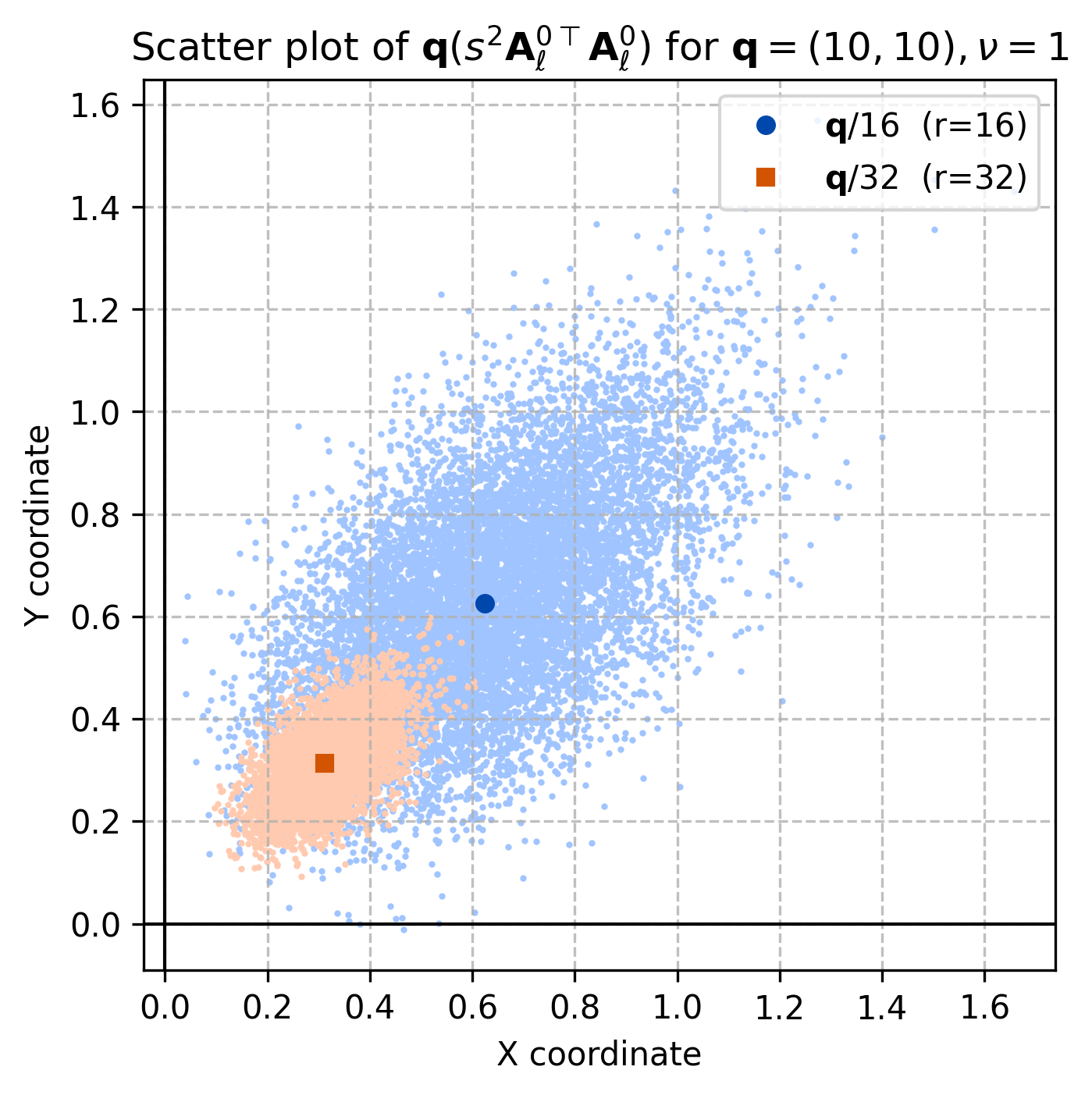}
\caption{Scatter plot of $\mathbf{q}(s^2\mathbf{A}_\ell^{0\top}\mathbf{A}_\ell^0)$, for example $\mathbf{q}=(10,10)$, i.e., $m=2$, $b=\sqrt{6/m}$, $10^4$ samples $\mathbf{A}_\ell^0 \sim \mathcal{U}(-b, b)$, $\alpha=1$, $\nu = {1}$ and $s=\alpha/r^\nu = 1/r$.}
\label{fig:scatterplot_uniform}
\end{figure}

The above lemma shows that, for a fixed $\mathbf{q}$, the term $\mathbf{q}(s^2\mathbf{A}_\ell^{0\top}\mathbf{A}_\ell^0)$ is a \emph{biased} perturbation of $\mathbf{q}$ with a distribution centered at mean $\alpha^2\sigma_{\mathbf{A}}^2~\mathbf{q}/r^{2\nu - 1}$. Also, it has the following signal-to-noise ratio (\texttt{SNR}): 

\begin{align}
    \forall~ \alpha, \nu: \texttt{SNR}\big(\mathbf{q}(s^2\mathbf{A}_\ell^{0\top}\mathbf{A}_\ell^0)\big) = \frac{\|\alpha^2\sigma_{\mathbf{A}}^2~\frac{\mathbf{q}}{r^{2\nu - 1}}\|^2}{Tr\big(\frac{\alpha^4\sigma_{\mathbf{A}}^4}{r^{4\nu -1}} (\|\mathbf{q}\|_2^2 ~ \mathbb{I}_m + \mathbf{D})\big)} = \frac{\|\alpha^2 \sigma_{\mathbf{A}}^2\frac{\mathbf{q}}{r^{2\nu - 1}}\|^2}{\frac{\alpha^4 \sigma_{\mathbf{A}}^4}{r^{4\nu-1}}(m-0.2)\|\mathbf{q}\|^2}  = \frac{r}{m-0.2},
\end{align}
which linearly increases with adaptation rank $r$ (similar to \Cref{eq:linearsnr} for Gaussian initialization). Choosing $\alpha=1$, $\nu=1$ yields to the following distribution:

$$
\mathcal{N}\big(\sigma_{\mathbf{A}}^2~\mathbf{q}/r, ~\sigma_{\mathbf{A}}^4 (\|\mathbf{q}\|_2^2 ~ \mathbb{I}_m + \mathbf{D})/r^3\big),
$$

with $\sigma_{\mathbf{A}}^2 = 2/m$. A scatter plot is shown in \Cref{fig:scatterplot_uniform} for an example $\mathbf{q} = (10, 10)$. It is noteworthy that, due to the 2D constraint of the figure, the condition $r\ll m (m=2)$ is not satisfied in this figure. Hence, the scatter plots are not completely round-shaped like a Gaussian distribution.

\newpage
\clearpage
\section{Effect of freezing and reusing $\mathbf{A}_\ell^0$ across time steps}\label{app:meanshift}

We derived \Cref{lem:noise_var} under the assumption of independence between $\mathbf{q}$ and $\mathbf{A}_\ell^{0}$. However, in the paper we proposed to resample $\mathbf{A}_\ell^{0}$ for clients at the beginning of each round, i.e., after every $T$ local steps. Therefore, in \cref{eq:adapterupdate} and \cref{frankffa+}, the gradient $\nabla_{\mathbf{W}_\ell^t} \mathcal{L}^t$ weakly depends on $\mathbf{A}_\ell^{0}$ for $1 \leq t \leq T-1$, as $\mathbf{A}_\ell^{0}$ is resampled at step $t=0$ and frozen after that. 

We now show that the true distribution of $\nabla_{\mathbf{W}_{\ell}^t} \mathcal{L}^t (s\mathbf{A}_\ell^0)^\top (s\mathbf{A}_\ell^0)$ and the distribution returned by \Cref{lem:noise_var}, which assumes independence, exhibit a bounded mean and covariance shift, where the bound decreases linearly with $T$ (and decreases with $r$). 

Let us assume $s=1/r^\nu$, i.e., $\alpha=1$. Also, consistent with \Cref{lem:noise_var}, let us assume that $\mathbf{A}_\ell^0 \in \mathbb{R}^{r \times m}\sim \mathcal{N}(0, \sigma_{\mathbf{A}}^2\mathbb{I}_{r \times m})$, with $\sigma_{\mathbf{A}}^2 = 1/r^\mu$. Let us denote the scaled matrix $s\mathbf{A}_\ell^0$ with $\tilde{ \mathbf{A}}$. Therefore, we have:

$$[\tilde{\mathbf{A}}^{\top} \tilde{\mathbf{A}}]_{i,j} =  \tilde{\mathbf{A}}_{:,i}^{\top} \tilde{\mathbf{A}}_{:,j} = \sum_{l=1}^r \tilde{a}_{l,i} \tilde{a}_{l,j} = r^{-2 \nu} \sum_{l=1}^r a_{l,i} a_{l,j}.
$$

Therefore, we have:

$$
    \mathbb E \big[ [\tilde{\mathbf{A}}^{\top} \tilde{\mathbf{A}}]_{i,j}\big] = \mathbb E \big[ \sum_{l=1}^r \tilde{a}_{l,i} \tilde{a}_{l,j}\big] = \mathbb E \big[r^{-2 \nu} \sum_{l=1}^r a_{l,i} a_{l,j}\big] = r^{-2 \nu} \sum_{l=1}^r \mathbb E \big[a_{l,i} a_{l,j}\big] \nonumber 
$$
$$
    = r^{-2 \nu} \cdot r \cdot r^{-\mu} \cdot \delta_{i,j} = r^{1-2 \nu - \mu} \delta_{i,j}.
$$
Therefore, we have:

$$
    \mathbb E \big[ \tilde{\mathbf{A}}^{\top} \tilde{\mathbf{A}}\big] = r^{1-2 \nu - \mu} \mathbb I_m.
$$

For simplicity of notations, let us denote the batch gradient $\nabla_{\mathbf{W}_{\ell}^t} \mathcal{L}^t$ at time step $t$ with $\mathbf g^t$, where $1 \leq t \leq T-1$. The true mean of the perturbed batch gradient at time step $t$ can be shown as the total expectation $\mathbb E [\mathbf g^t \tilde{\mathbf{A}}^{\top} \tilde{\mathbf{A}}]$ (over $\mathbf g^t$ and $\tilde{\mathbf{A}}$). In contrast, under the assumption of independence between $\tilde{\mathbf{A}}$ and $\mathbf g^t$, which resembles \Cref{lem:noise_var}, the mean of the perturbed batch gradient at time step $t$ is 

$$
\mathbb E [\mathbf g^t \tilde{\mathbf{A}}^{\top} \tilde{\mathbf{A}}] = \mathbb E [\mathbf g^t ] \cdot \mathbb E [ \tilde{\mathbf{A}}^{\top} \tilde{\mathbf{A}}] = r^{1-2 \nu - \mu} \mathbb E [\mathbf g^t ].    
$$

Therefore, the mean shift between the true distribution and the distribution returned by \Cref{lem:noise_var} under independence assumption can be found as follows:

$$
    \Delta_\mu^t = \mathbb E [\mathbf g^t \tilde{\mathbf{A}}^{\top} \tilde{\mathbf{A}}] - \big(r^{1-2 \nu - \mu} \mathbb E [\mathbf g^t]\big) = \mathbb E \bigg[(\mathbf g^t - \mathbf g^0) (\tilde{\mathbf{A}}^{\top} \tilde{\mathbf{A}} - r^{1-2 \nu - \mu} \mathbb I_m)\bigg],
$$

where in the last equality, we used the fact that $\mathbf g^0$ and $\tilde{\mathbf{A}}$ are independent. Therefore, we have:

\begin{align}\label{eq:delta_mu}
    \|\Delta_\mu^t\|_F &= \bigg\|\mathbb E \bigg[(\mathbf g^t - \mathbf g^0) (\tilde{\mathbf{A}}^{\top} \tilde{\mathbf{A}} - r^{1-2 \nu - \mu} \mathbb I_m)\bigg]\bigg\|_F \leq \mathbb E \bigg[\bigg\|(\mathbf g^t - \mathbf g^0) (\tilde{\mathbf{A}}^{\top} \tilde{\mathbf{A}} - r^{1-2 \nu - \mu} \mathbb I_m)\bigg\|_F\bigg]\nonumber \\
    & \leq \mathbb E \bigg[\big\|(\mathbf g^t - \mathbf g^0)\big\|_F \big\|  (\tilde{\mathbf{A}}^{\top} \tilde{\mathbf{A}} - r^{1-2 \nu - \mu} \mathbb I_m)\big\|_2\bigg].
\end{align}

Assuming \texttt{SGD}, we have:

$$
    \mathbf{W}_\ell^t = \mathbf{W}_\ell^0 - \eta \sum_{\tau=0}^{t-1} \mathbf g^\tau (s\mathbf{A}_\ell^0)^\top (s\mathbf{A}_\ell^0) = \mathbf{W}_\ell^0 - \eta \sum_{\tau=0}^{t-1} \mathbf g^\tau \tilde{\mathbf{A}}^{\top} \tilde{\mathbf{A}},
$$

where $1 \leq t \leq T-1$. Using $\beta$-smoothness of the loss function, we have:

$$
    \|\mathbf g^t - \mathbf g^0\|_F \leq \beta \|\mathbf{W}_\ell^t - \mathbf{W}_\ell^0\|_F = \beta \eta \bigg\|\sum_{\tau=0}^{t-1} \mathbf g^\tau \tilde{\mathbf{A}}^{\top} \tilde{\mathbf{A}}\bigg\|_F \leq \beta \eta \sum_{\tau=0}^{t-1}\big\| \mathbf g^\tau \tilde{\mathbf{A}}^{\top} \tilde{\mathbf{A}}\big\|_F 
    $$
    $$
    \leq \beta \eta \sum_{\tau=0}^{t-1} \big\| \mathbf g^\tau\big\|_F \big\| \tilde{\mathbf{A}}^{\top} \tilde{\mathbf{A}}\big\|_2 = \beta \eta \cdot \bigg(\sum_{\tau=0}^{t-1} \big\| \mathbf g^\tau\big\|_F\bigg) \cdot \big\| \tilde{\mathbf{A}}^{\top} \tilde{\mathbf{A}}\big\|_2.
$$    

Now, note that for random variables $a$, $b$ and $c$, if $a \le c$ and $b \ge 0$ almost surely, then pointwise $ab \le cb$, which implies $\mathbb{E}[ab] \le \mathbb{E}[cb]$ by the monotonicity of expected value.
Therefore, by plugging the above bound into the right side of \Cref{eq:delta_mu}, we get:

\begin{align}\label{eq:norm_mu}
    \|\Delta_\mu^t\|_F \leq \beta \eta \cdot \mathbb E \bigg[\bigg(\sum_{\tau=0}^{t-1} \big\| \mathbf g^\tau\big\|_F\bigg) \cdot \big\| \tilde{\mathbf{A}}^{\top} \tilde{\mathbf{A}}\big\|_2 \cdot \big\|  (\tilde{\mathbf{A}}^{\top} \tilde{\mathbf{A}} - r^{1-2 \nu - \mu} \mathbb I_m)\big\|_2\bigg].
\end{align}

We can now use the $L_0$-Lipschitz continuity of the loss function to derive an upper bound for the term in parentheses:

$$
    \|\Delta_\mu^t\|_F \leq \beta \eta t L_0 \cdot \mathbb E \bigg[\big\| \tilde{\mathbf{A}}^{\top} \tilde{\mathbf{A}}\big\|_2 \cdot \big\|  (\tilde{\mathbf{A}}^{\top} \tilde{\mathbf{A}} - r^{1-2 \nu - \mu} \mathbb I_m)\big\|_2\bigg] 
    $$

    $$
    \leq \beta \eta T L_0 \cdot \mathbb E \bigg[\big\| \tilde{\mathbf{A}}^{\top} \tilde{\mathbf{A}}\big\|_2 \cdot \big\|  (\tilde{\mathbf{A}}^{\top} \tilde{\mathbf{A}} - r^{1-2 \nu - \mu} \mathbb I_m)\big\|_2\bigg],
$$

where in the last line, we simply used the fact that $1\leq t\leq T-1$. As observed, the above upper-bound on the mean shift decreases linearly as $T$ (number of clients' local steps during each round or the \emph{resampling period}) and $\eta$ decrease. Also, it can be shown that the expectation term decreases as rank $r$ increases. Therefore, the mean shift decreases as $r$ increases.

A similar bound can be derived on the covariance shift norm $\|\Delta_\Sigma^t\|_F$, which decreases linearly as $T$ and $\eta$ decrease.

The above results show that freezing and reusing the \algname{LoRA} matrix $\mathbf{A}_\ell^0$ results in a bounded shift in the mean and covariance derived in \Cref{lem:noise_var}. More frequent resampling of the matrix reduces this shift. This validates the distribution approximation mentioned at the end of  \Cref{sec:perturbation_analysis}. 

We can now get a more useful insight about \cref{eq:norm_mu}. Remember that $\mathbb E [\tilde{\mathbf{A}}^{\top} \tilde{\mathbf{A}}] = r^{1-2 \nu - \mu} \mathbb I_m$. Therefore, the norm product term $\big\| \tilde{\mathbf{A}}^{\top} \tilde{\mathbf{A}}\big\|_2 \cdot \big\|  (\tilde{\mathbf{A}}^{\top} \tilde{\mathbf{A}} - r^{1-2 \nu - \mu} \mathbb I_m)\big\|_2$ is almost surely bounded and has a high-probability bound $\zeta_r$, which decreases with $r$. Conditioning on the event that $\big\| \tilde{\mathbf{A}}^{\top} \tilde{\mathbf{A}}\big\|_2 \cdot \big\|  (\tilde{\mathbf{A}}^{\top} \tilde{\mathbf{A}} - r^{1-2 \nu - \mu} \mathbb I_m)\big\|_2 \leq \zeta_r$ holds, \cref{eq:norm_mu} can be rewritten as:

\begin{align}\label{eq:norm_mu_gamma}
    \|\Delta_\mu^t\|_F \leq \beta \eta \zeta_r \mathbb E \bigg[\bigg(\sum_{\tau=0}^{t-1} \big\| \mathbf g^\tau\big\|_F \bigg)\bigg] = \beta \eta\zeta_r \sum_{\tau=0}^{t-1} \mathbb E \big[\big\| \mathbf g^\tau\big\|_F \big] &\leq \gamma_r \sum_{\tau=0}^{t-1} \mathbb E \big[\big\| \mathbf g ^\tau\big\|_F \big],
\end{align}

where constants $\zeta_r$ and $\gamma_r = \beta \eta \zeta_r $ depend on rank $r$. \textbf{Note that for any $r$, $\gamma_r \propto \eta$}. We will use this point in our convergence proof in \Cref{app:conv_proof}. The above inequality bounds the mean shift norm at step $t$ ($1\leq t \leq T-1$) with the sum of the expected batch gradient norms at the previous steps. Note that batch gradient norms generally decrease as training continues. Therefore, as \algname{SPRA-LoRA} training continues, the upper-bound and consequently the mean shift decrease from one round to the next round. \textbf{Therefore, the per-$T$-steps resampling method of \algname{SPRA-LoRA} becomes more similar to the per-step resampling gradually}. This observation provides an intuitive insight about convergence of \algname{SPRA-LoRA}.

\section{Proof of \Cref{thm:franklora}}\label{app:conv_proof}

\franklora*

\begin{proof}

We first explain the notations and assumptions used in the proof in the following two subsections.

\subsection{Notations used in the proof}\label{sec:notations}
We use the following notations for simplicity:

\begin{align}\label{eq:gradnotation}
\mathbf g_i^{e,t}=\nabla_{\mathbf{W}_i^{e,t}}\mathcal{L}_i^{e,t}(\mathbf{W}_i^{e,t}) = \frac{1}{b_i}\sum_{x \in \mathcal{B}_i^{e,t}}\nabla_{\mathbf{W}_i^{e,t}}l(x; \mathbf{W}_i^{e,t}),    
\end{align}
i.e., batch gradient of client $i$ at adapter $\mathbf{W}_i^{e,t}$. We also use $\tilde{\mathbf g}_i^{e,t}$ to denote the perturbed batch gradient
\begin{align*}
    \tilde{\mathbf g}_i^{e,t} = \mathbf g_i^{e,t} (s_i\mathbf{A}_i^{e,0})^\top (s_i\mathbf{A}_i^{e,0}).
\end{align*}

We also use the following simplified notations: 

\begin{align}\label{eq:lossnotation}
 &f_i^{e,t} = f_i(\mathbf{W}_i^{e,t})=\frac{1}{N_i}\sum_{x\in \mathcal{D}_i}l(x;\mathbf{W}_i^{e,t}), \nonumber \\
 &\nabla f_i^{e,t}= \nabla_{\mathbf{W}_i^{e,t}} f_i(\mathbf{W}_i^{e,t}),
\end{align}
i.e., the objective function of client $i$ and its \emph{full} gradient at the adapter $\mathbf{W}_i^{e,t}$.  Finally, for matrices $\mathbf{A}, \mathbf{B} \in \mathbb R ^{n \times m}$, define the Frobenius inner product and norm by:

\begin{align*}
    \langle \mathbf{A}, \mathbf{B} \rangle_F = Tr(\mathbf{A}, \mathbf{B}), ~~~ \|\mathbf{A}\|_F  = \sqrt{\langle \mathbf{A}, \mathbf{A} \rangle_F}.
\end{align*}

\subsection{Assumptions used in the proof}
In this section, we formalize our assumptions used for the convergence analysis of \algname{SPRA-LoRA}:
\begin{assumption}[\textbf{Lipschitz continuity, $\beta$-smoothness, bounded gradient heterogeneity, and bounded gradient variance}]\label{assump:lipschitz_smooth_bounded}
\end{assumption}

\begin{itemize}
    \item Functions $\{f_i\}_{i=1}^n$ are $L_0$-Lipschitz continuous and $\beta$-smooth: $\forall ~ \mathbf{W},\mathbf{W}',\\
    |f_i(\mathbf{W}) - f_i(\mathbf{W}')| \leq L_{0}\|\mathbf{W} - \mathbf{W}'\|_F$ and $\|\nabla f_i(\mathbf{W}) - \nabla f_i(\mathbf{W}')\|_F \leq \beta\|\mathbf{W}-\mathbf{W}'\|_F$.
    \item For every $i$ and $j, f_i-f_j$ is $\sigma_{\textit{het}}$-Lipschitz continuous: $\|\nabla f_i(\mathbf{W})-\nabla f_j(\mathbf{W})\|_F\leq\sigma_{\textit{het}}.$ In particular $\normF{\nabla f_i(\mathbf W)-\nabla f(\mathbf W)}\le \sigma_{het}$.
    \item The stochastic batch gradient $\mathbf g_i^{e,t}$ is an unbiased estimate of $\nabla f_i^{e,t}$ with a bounded variance: \\
    $ \mathbb{E}_{\mathcal{B}_i^{e,t}} [\mathbf g_i^{e,t}] = \nabla f_i^{e,t}, ~~ \mathop{\mathbb{E}_{\mathcal{B}_i^{e,t}}}\big[\|\mathbf g_i^{e,t} - \nabla f_i^{e,t}\|_F^2\big] \leq \sigma_{i, g}^2$.
\end{itemize}

\subsection{Proof of the Theorem}

For the proof, we use the result of \Cref{lem:noise_var} and the findings in \Cref{app:meanshift}. In \algname{SPRA-LoRA} with $\nu=0.5$, i.e., rank-stabilized \algname{LoRA}, each client samples the frozen \algname{LoRA} components $\mathbf{A}_i^{e,0}$ from $\mathcal{N}(0, \frac{1}{r_i^2}\mathbb{I}_{r_i \times n})$.

The adapter update of client $i$ made at the $(t+1)$-th local \texttt{SGD} step ($t \in \{0, \ldots, T-1\}$) in round $e$ is obtained from the multiplication of the update $s_i \Delta \mathbf{B}_i^{e,t}$ and the resampled $\mathbf{A}_i^{e,0}$ of the round:

$$
    \Delta \mathbf{W}_i^{e,t} = s_i\Delta \mathbf{B}_i^{e,t} \mathbf{A}_i^{e,0} = -\eta_{r_i} (\nabla_{\mathbf{B}_i^{e,t}}\mathcal{L}_i^{e,t})(s_i\mathbf{A}_i^{e,0}) = -\eta_{r_i} (\nabla_{\mathbf{W}_i^{e,t}}\mathcal{L}_i^{e,t})(s_i\mathbf{A}_i^{e,0})^\top (s_i\mathbf{A}_i^{e,0})
$$
$$
~~~~~~~~~~~~~~~~~~~~~~~~~~~~~~ ~~~~~~~~~~~~~~~~~~~~~~~~~~~~~~~~~~~~~~~~~~~= -\eta_{r_i} \mathbf g_i^{e,t}(s_i\mathbf{A}_i^{e,0})^\top (s_i\mathbf{A}_i^{e,0})
$$
$$
~~~~~~~~~~~~~~~~~~~~~~~~~~~~~~ ~~~~~~~~~~~~= -\eta_{r_i} \widetilde{\mathbf g}_i^{e,t},
$$

where in the second line, we used the batch gradient notation introduced in \Cref{eq:gradnotation}. This is equivalent to a \texttt{SGD} step with step size $\eta_{r_i}$ and perturbed batch gradient $\widetilde{\mathbf g}_i^{e,t} = \mathbf g_i^{e,t} (s_i\mathbf{A}_i^{e,0})^\top (s_i\mathbf{A}_i^{e,0})$ w.r.t the adapter $\mathbf{W}_i^{e,t}$. Now, remember from \Cref{app:meanshift} that, due to the dependence between $\mathbf g_i^{e,t}$ ($t \geq 1$) and $\mathbf{A}_i^{e,0}$, $\mathbb E[\widetilde{\mathbf g}_i^{e,t}]$ experiences a bounded shift from the mean $\mathbb E[\mathbf g_i^{e,t}]/r_i^2 = \nabla f_i^{e,t}/r_i^2$ returned by \Cref{lem:noise_var} (Remember, we have assumed rank-stabilized \algname{LoRA} with $\nu=0.5$ and $\mu=2$). Therefore, based on the derivations from the same \Cref{app:meanshift}, $\mathbb E[\widetilde{\mathbf g}_i^{e,t}]$ can be written as follows based on a bounded mean shift $\mathbf \Delta_{\mu_i}^{e,t}$:

$$
    \mathbb E[\widetilde{\mathbf g}_i^{e,t}] = \frac{\nabla f_i^{e,t}}{r_i^2} + \mathbf \Delta_{i}^{e,t},
$$

where the constant mean shift $\mathbf \Delta_{i}^{e,t}$ satisfies $\|\mathbf \Delta_{i}^{e,t}\|_F \leq \gamma_{r_i} \sum_{\tau=0}^{t-1} \big\| \mathbf{g}_i^{e,\tau}\big\|_F$. Based on \cref{eq:norm_mu_gamma}, constant $\gamma_{r_i}$ depends on the rank $r_i$ of client $i$ and is proportional to the step size $\eta_{r_i} = \eta_c r_i^2$, and hence proportional to $\eta_c$ as well. Assuming a bounded variance around this shifted mean, we can rewrite the perturbed batch gradient $\widetilde{\mathbf g}_i^{e,t}$ as follows:

$$
    \widetilde{\mathbf g}_i^{e,t} = \frac{\nabla f_i^{e,t}}{r_i^2} + \mathbf \Delta_{i}^{e,t} + \mathbf \Sigma_i^{e,t},
$$

where $\mathbf \Sigma_i^{e,t} \in \mathbb R ^{n \times m}$ is random with zero mean and has a bounded variance $\mathbb E[\|\mathbf \Sigma_i^{e,t}\|_F^2] \leq \sigma_{i, \tilde{g}}^2$, and the random matrices $\{\mathbf \Sigma_i^{e,t}\}_{i,t}$ are mutually independent across clients $i$ and steps $t$ within a round. 

Throughout, $\mathcal F^e:=\mathcal F^{e,0}$ denotes the $\sigma$-algebra of all randomness up to (not including)
round $e$, so that $\W^e$ is $\mathcal F^e$-measurable. At each local step, client $i$ performs the following update:

$$
    \mathbf W_i^{e,t+1}
    =
    \mathbf W_i^{e,t}
    -
    \eta_c r_i^2
    \widetilde{\mathbf g}_i^{e,t}.
$$
After $T$ local steps, the server averages the local models:
$$
    \mathbf W^{e+1}
    =
    \frac{1}{n}\sum_{i=1}^{n}
    \mathbf W_i^{e,T}.
$$

We first rewrite the local update. Since
$$
    \widetilde{\mathbf g}_i^{e,t}
    =
    \frac{1}{r_i^2}
    \nabla f_i^{e,t}
    +
    \mathbf \Delta_{i}^{e,t}
    +
    \mathbf \Sigma_i^{e,t},
$$
the update
$$
    \mathbf W_i^{e,t+1}
    =
    \mathbf W_i^{e,t}
    -
    \eta_c r_i^2
    \widetilde{\mathbf g}_i^{e,t}
$$
can be written as
\begin{equation}\label{eq:perturbed_gd}
\boxed{
    \mathbf W_i^{e,t+1}
    =
    \mathbf W_i^{e,t}
    -
    \eta_c \nabla f_i^{e,t}
    -
    \eta_c r_i^2\mathbf \Delta_{i}^{e,t}
    -
    \eta_c r_i^2\mathbf \Sigma_i^{e,t}.
}
\end{equation}

\paragraph{Step 1: Round update.} Averaging the equality ebove over $i$ and telescoping over $t=0,\dots,T-1$ (using $\W_i^{e,0}=\W^e$),

\begin{align}\label{eq:4}
\W^{e+1}=\frac1n\sum_{i=1}^n \W_i^{e,T}
=\W^e-\eta_c\sum_{t=0}^{T-1}\Big(\bar g^{e,t}+\bar b^{e,t}+\bar{\mathbf\Sigma}^{e,t}\Big),
\end{align}

where
$$
\bar g^{e,t}:=\frac1n\sum_{i=1}^n \nabla f_i^{e,t},\qquad
\bar b^{e,t}:=\frac1n\sum_{i=1}^n r_i^2\mathbf\Delta_{i}^{e,t},\qquad
\bar{\mathbf\Sigma}^{e,t}:=\frac1n\sum_{i=1}^n r_i^2\mathbf\Sigma_i^{e,t}.
$$
Define
$$
\mathbf E^e:=\sum_{t=0}^{T-1}\big(\bar g^{e,t}-\nabla f(\W^e)\big),\qquad
\mathbf B^e:=\sum_{t=0}^{T-1}\bar b^{e,t},\qquad
\mathbf N^e:=\sum_{t=0}^{T-1}\bar{\mathbf\Sigma}^{e,t}.
$$
Then \cref{eq:4} reads
\begin{align}\label{eq:global_step}
\boxed{
\W^{e+1}-\W^e=-\eta_c T\nabla f(\W^e)-\eta_c\big(\mathbf E^e+\mathbf B^e+\mathbf N^e\big).}
\end{align}

\paragraph{Step 2: Bounding client drift.}
Telescoping \cref{eq:perturbed_gd} for a single client from step $0$ to $t-1$,
$$
\W_i^{e,t}-\W^e=-\eta_c\Big(\underbrace{\sum_{s=0}^{t-1}\nabla f_i^{e,s}}_{A_t}
+r_i^2\underbrace{\sum_{s=0}^{t-1}\mathbf\Delta_{i}^{e,s}}_{B_t}
+r_i^2\underbrace{\sum_{s=0}^{t-1}\mathbf\Sigma_i^{e,s}}_{C_t}\Big).
$$

\textbf{\emph{Bound on $A_t$:}} By our assumptions,
$\normF{\nabla f_i^{e,s}}\le \normF{\nabla f(\W_i^{e,s})}+\sigma_{het}\le L_0+\sigma_{het}$ a.s., hence
$\normF{A_t}\le t(L_0+\sigma_{het})$ a.s.

\textbf{\emph{Bound on $B_t$:}} By our bounded variance assumption,
$$
\mathbb E\big[\normF{g_i^{e,s}}\big] \le \sqrt{\mathbb E\big[\normF{g_i^{e,s}}^2}\big]
=\sqrt{\normF{\nabla f_i^{e,s}}^2+\mathbb E \big[\normF{g_i^{e,s}-\nabla f_i^{e,s}}^2\big]}
\le \sqrt{L_0^2+\sigma_{i,g}^2}\le L_0+\sigma_{i,g}.
$$
By the above inequality and the bound we derived on $\normF{\mathbf\Delta_{i}^{e,s}}$, we have
$\normF{\mathbf\Delta_{i}^{e,s}}\le\gamma_{r_i}\sum_{\tau<s}\mathbb E \big[\normF{g_i^{e,\tau}}\big]\le \gamma_{r_i}(L_0+\sigma_{i,g})s$ a.s.,
so
\begin{align}\label{eq:boundondeltanorm}
    \normF{B_t}\le \sum_{s=0}^{t-1}\gamma_{r_i}(L_0+\sigma_{i,g})\,s=\gamma_{r_i}(L_0+\sigma_{i,g})\frac{t(t-1)}{2}\quad\text{a.s.}
\end{align}

\textbf{\emph{Bound on $C_t$:}} By assumption of $\mathbf \Sigma_i^{e,s}$ having a zero-mean with bounded variance $\sigma_{i,\tilde g}^2$ and their independence across $s$, the cross terms vanish in expectation, so
$$
\mathbb E \big[\normF{C_t}^2\big]=\sum_{s=0}^{t-1}\mathbb E \big[\normF{\mathbf \Sigma_i^{e,s}}^2\big]\le t\,\sigma_{i,\tilde g}^2.
$$

Combining the bounds on $A_t$, $B_t$ and $C_t$ using $\normF{x+y+z}^2\le 3(\normF x^2+\normF y^2+\normF z^2)$ together with $t\le T$,
$$
\mathbb E \bigg[\normF{\W_i^{e,t}-\W^e}^2\bigg]
\le 3\eta_c^2\Big[T^2(L_0+\sigma_{het})^2
+\frac{r_i^4\gamma_{r_i}^2(L_0+\sigma_{i,g})^2T^2(T-1)^2}{4}
+r_i^4T\sigma_{i,\tilde g}^2\Big]=:\hat D_i^2.
$$

\paragraph{Step 3: Bounding $\mathbf E^e$.}
By the $\beta$-smoothness assumption and Jensen's inequality,
\begin{align}\label{eq:norm_diff}
\normF{\bar g^{e,t}-\nabla f(\W^e)}^2
&=\Big\|\frac1n\sum_i\big(\nabla f_i(\W_i^{e,t})-\nabla f_i(\W^e)\big)\Big\|_\F^2
\nonumber \\ &\le \frac1n\sum_i\normF{\nabla f_i(\W_i^{e,t})-\nabla f_i(\W^e)}^2 
\nonumber \\
&\le \frac{\beta^2}{n}\sum_i\normF{\W_i^{e,t}-\W^e}^2 .
\end{align}

Taking expectation of \cref{eq:norm_diff} and using the previous upper bound $\hat{D}_i^2$:
$$
\mathbb E \big[\normF{\bar g^{e,t}-\nabla f(\W^e)}^2\big]\le \beta^2 \hat{\bar D}^2,\qquad \hat{\bar D}^2:=\frac1n\sum_i\hat D_i^2 . 
$$
Since $\mathbf E^e$ is a sum of $T$ such terms, by Cauchy--Schwarz,
$$
\mathbb E\big[\normF{\mathbf E^e}^2\big]\le T\sum_{t=0}^{T-1}\mathbb E \bigg[\normF{\bar g^{e,t}-\nabla f(\W^e)}^2\bigg]\le T^2\beta^2\hat{\bar D}^2=:\epsilon_1^2.
$$

\paragraph{Step 4: Bounding $\mathbf B^e$.}
By the bound derived before \cref{eq:boundondeltanorm},
we have $\normF{\mathbf\Delta_{i}^{e,t}}\le\gamma_{r_i}(L_0+\sigma_{i,g})\,t$ a.s. Hence, by the triangle inequality,
$$
\normF{\mathbf B^e}\le\sum_{t=0}^{T-1}\frac1n\sum_i r_i^2\gamma_{r_i}(L_0+\sigma_{i,g})\,t
=\Gamma\sum_{t=0}^{T-1}t=\Gamma\,\frac{T(T-1)}{2}=:\epsilon_2\qquad\text{a.s.,}
$$
where $\Gamma = \frac{1}{n}\sum_{i} r_i^2 \gamma_{r_i}(L_0 + \sigma_{i,g})$.
\paragraph{Step 5: Bounding $\mathbf N^e$.}
By our assumptions on $\mathbf \Sigma_i^{e,t}$, $\mathbb E[\bar{\mathbf\Sigma}^{e,t}\mid\mathcal F^{e,t}]=0$, so $\mathbb E[\mathbf N^e\mid\mathcal F^e]=0$.
By the assumption of independence of $\mathbf\Sigma_i^{e,t}$ across $i$ and $t$), cross terms vanish and
$$
\mathbb E \big[\normF{\mathbf N^e}^2\big]=\sum_{t=0}^{T-1}\mathbb E\big[\normF{\bar{\mathbf\Sigma}^{e,t}}^2\big]
=\sum_{t=0}^{T-1}\frac{1}{n^2}\sum_i r_i^4\,\mathbb E\big[\normF{\mathbf\Sigma_i^{e,t}}^2\big]
\le T\cdot\frac1{n^2}\sum_i r_i^4\sigma_{i,\tilde g}^2=TS^2=:\epsilon_3^2,
$$

where $S^2:=\frac{1}{n^2}\sum_{i=1}^n r_i^4\sigma_{i,\tilde g}^2$.

\paragraph{Step 6: One-round descent.}
Using the smoothness inequality with $\mathbf \W^e$ and $\W^{e+1}$, and \cref{eq:global_step}:
$$
f(\W^{e+1})\le f(\W^e)-\eta_c T\ip{\nabla f(\W^e)}{\nabla f(\W^e)}-\eta_c\ip{\nabla f(\W^e)}{\mathbf E^e+\mathbf B^e+\mathbf N^e}
$$
$$
+\frac{\beta}{2}\eta_c^2\normF{T\nabla f(\W^e)+\mathbf E^e+\mathbf B^e+\mathbf N^e}^2 .
$$

Take $\mathbb E[\cdot\mid\mathcal F^e]$. For the linear term, since $\mathbb E[\mathbf N^e\mid\mathcal F^e]=0$,
$$
-\eta_c\,\mathbb E\big[\ip{\nabla f(\W^e)}{\mathbf E^e+\mathbf B^e+\mathbf N^e}\,\big|\,\mathcal F^e\big]
=-\eta_c\ip{\nabla f(\W^e)}{\mathbb E[\mathbf E^e\mid\mathcal F^e]+\mathbf B^e}
$$
$$
~~~~~~~~~~~~~~~~~~~~~~~~~~~~~~~~~~~~~~~~~~~~~~~~~~~~~~~~~~\le \eta_c L_0\big(\mathbb E[\normF{\mathbf E^e}\mid\mathcal F^e]+\epsilon_2\big),
$$
using Lipschitz continuity assumption, Cauchy--Schwarz, and the bound derived in Step 4. By Jensen and the bound derived in Step 3, $\mathbb E[\normF{\mathbf E^e}\mid\mathcal F^e]\le\epsilon_1$
(taking full expectation later). For the quadratic term, using $\normF{a+b+c+d}^2\le 4(\normF a^2+\normF b^2+\normF c^2+\normF d^2)$,
$$
\mathbb E\big[\normF{T\nabla f(\W^e)+\mathbf E^e+\mathbf B^e+\mathbf N^e}^2\,\big|\,\mathcal F^e\big]
\le 4T^2\normF{\nabla f(\W^e)}^2+4\big(\epsilon_1^2+\epsilon_2^2+\epsilon_3^2\big).
$$
Combining and taking full expectation,
$$
\mathbb E \big[f(\W^{e+1})\big]\le \mathbb E \big[f(\W^e)\big]-\eta_c T\,\mathbb E \big[\normF{\nabla f(\W^e)}^2\big]
+\eta_c L_0(\epsilon_1+\epsilon_2)
$$
$$
~~~~~~~~~~~~~~~~~~~~~~+2\beta\eta_c^2T^2\,\mathbb E \big[\normF{\nabla f(\W^e)}^2\big]
+2\beta\eta_c^2(\epsilon_1^2+\epsilon_2^2+\epsilon_3^2).
$$
Rearranging the above, we get
$$
\eta_c T\big(1-2\beta\eta_c T\big)\,\mathbb E \big[\normF{\nabla f(\W^e)}^2\big]
\le \mathbb E \big[f(\W^e)\big]-\mathbb E \big[f(\W^{e+1})\big]+\eta_cL_0(\epsilon_1+\epsilon_2)
$$
$$
~~~~~~~~~~~~~~~~~~~~~~~~~+2\beta\eta_c^2(\epsilon_1^2+\epsilon_2^2+\epsilon_3^2).
$$

\paragraph{Step 7: Telescoping.}
Since $\eta_c \le \tfrac1{4\beta T}$, we have $1-2\beta\eta_c T\ge \tfrac12$, so from above,
$$
\frac{\eta_c T}{2}\,\mathbb E\big[\normF{\nabla f(\W^e)}^2\big]
\le \mathbb E \big[f(\W^e)\big]-\mathbb E \big[f(\W^{e+1})\big]+\eta_c L_0(\epsilon_1+\epsilon_2)+2\beta\eta_c^2(\epsilon_1^2+\epsilon_2^2+\epsilon_3^2).
$$
Summing over $e=0,\dots,E-1$, the right-hand side telescopes:
$$
\frac{\eta_c T}{2}\sum_{e=0}^{E-1}\mathbb E\big[\normF{\nabla f(\W^e)}^2\big]
\le f(\W^0)-\mathbb E \big[f(\W^{E})\big]+E\eta_c L_0(\epsilon_1+\epsilon_2)+2E\beta\eta_c^2(\epsilon_1^2+\epsilon_2^2+\epsilon_3^2).
$$
Using $\mathbb E \big[f(\W^E)\big]\ge f^\star$ and dividing by $\eta_c TE/2$,
$$
\frac1E\sum_{e=0}^{E-1}\mathbb E \big[\normF{\nabla f(\W^e)}^2\big]
\le \frac{2\big(f(\W^0)-f^\star\big)}{\eta_c TE}+\frac{2L_0(\epsilon_1+\epsilon_2)}{T}+\frac{4\beta\eta_c(\epsilon_1^2+\epsilon_2^2+\epsilon_3^2)}{T}.
$$

We observe that $\epsilon_1=O(\eta_c)$ (each $\hat D_i=O(\eta_c)$). Also, substituting $\epsilon_2=\tfrac{T(T-1)}2\Gamma$ so that $\tfrac{2L_0\epsilon_2}{T}=L_0(T-1)\Gamma$, we observe that  $\epsilon_2=O(\eta_c)$ ($\Gamma \propto \sum_i r_i^2 \gamma_{r_i}$, where $\gamma_{r_i}\propto \eta_{r_i}$ (from \cref{eq:norm_mu_gamma}), and $\eta_{r_i}=\eta_c r_i^2$) and $\epsilon_3^2=O(1)$ independent of $\eta_c$. Therefore,

$$
\boxed{
\min_e \mathbb E \big[\normF{\nabla f(\W^e)}^2\big]
\le \frac{2\big(f(\W^0)-f^\star\big)}{\eta_c TE}+ \Psi}
$$
where $\Psi = \mathcal{O}(\eta_c)$. 

\paragraph{Corollary (rate).}
Choosing
$\eta_c=\min\big(\tfrac1{4\beta T},\tfrac1{\sqrt{E}}\big)$ makes
$\tfrac{2(f(\W^0)-f^\star)}{\eta_c TE}=\mathcal{O}(1/\sqrt{E})$, and
$\Psi=\mathcal{O}(\eta_c)=\mathcal{O}(1/\sqrt{E})$. Hence
$$
\min_e \mathbb E \big[\normF{\nabla f(\W^e)}^2\big]=\mathcal{O}\!\left(\frac1{\sqrt{E}}\right),
$$
proving the bound reduces to
the classical $\mathcal{O}(1/\sqrt{E})$ non-convex FedAvg convergence rate.

\end{proof}


\end{document}